\documentclass[preprint,review,12pt]{elsarticle}

\usepackage{amssymb}
\usepackage{amsmath}

\usepackage{url}

\usepackage[version=4]{mhchem}
\usepackage[version=4]{mhchem}
\usepackage{multirow}
\usepackage[table,xcdraw]{xcolor}

\graphicspath{{./figures/}}

\begin{document}

\begin{frontmatter}

\title{Systematic Review and Meta-analysis of AI-driven MRI Motion Artifact Detection and Correction} 

 \author[label1]{Mojtaba Safari}
 \author[label1]{Zach Eidex}
 \author[label1]{Richard L.J. Qiu}
 \author[label1]{Matthew Goette}
 \author[label2]{Tonghe Wang}
 \author[label1]{Xiaofeng Yang\corref{cor1}}

 \cortext[cor1]{Author to whom correspondence should be addressed}
 \ead{xiaofeng.yang@emory.edu}
 \affiliation[label1]{organization={Department of Radiation Oncology and Winship Cancer Institute},
             addressline={Emory University},
             city={Atlanta},
             postcode={30322},
             state={GA},
             country={USA}}
 \affiliation[label2]{organization={Department of Medical Physics},
 	addressline={Memorial Sloan Kettering Cancer Center},
 	city={New York},
 	postcode={10065},
 	state={NY},
 	country={USA}}

\begin{abstract}
\textbf{Background:} To systematically review and perform a meta-analysis of artificial intelligence (AI)-driven methods for detecting and correcting magnetic resonance imaging (MRI) motion artifacts, assessing current developments, effectiveness, challenges, and future research directions. \textbf{Methods:} A comprehensive systematic review and meta-analysis were conducted, focusing on deep learning (DL) approaches, particularly generative models, for the detection and correction of MRI motion artifacts. Quantitative data were extracted regarding utilized datasets, DL architectures, and performance metrics. \textbf{Results:} DL, particularly generative models, shows promise for reducing motion artifacts and improving image quality; however, limited generalizability, reliance on paired training data, and risk of visual distortions remain key challenges that motivate standardized datasets and reporting. \textbf{Conclusions:} AI-driven methods, particularly DL generative models, show significant potential for improving MRI image quality by effectively addressing motion artifacts. However, critical challenges must be addressed, including the need for comprehensive public datasets, standardized reporting protocols for artifact levels, and more advanced, adaptable DL techniques to reduce reliance on extensive paired datasets. Addressing these aspects could substantially enhance MRI diagnostic accuracy, reduce healthcare costs, and improve patient care outcomes.
\end{abstract}

\begin{highlights}
	\item Conducted the first meta-analysis of AI-driven MRI motion artifact detection and correction methods
	\item Systematically reviewed deep learning models used for MRI artifact correction
	\item Identified current dataset trends, key challenges, and existing research gaps
	\item Proposed directions to develop robust, generalizable, and effective AI models
\end{highlights}


\end{frontmatter}

\subsubsection*{\textbf{Abbreviations}}
\begin{itemize}
	
	\item \ce{B0}: Magnetic field strength in MRI
	\item CLAIM: Checklist for Artificial Intelligence in Medical Imaging
	\item CNN: Convolutional Neural Network
	\item CycleGAN: Cycle-Consistent Generative Adversarial Network
	\item cGAN: Conditional Generative Adversarial Network
	\item DDPM: Denoising Diffusion Probabilistic Model
	\item DNN: Deep Neural Network
	\item DL: Deep Learning
	\item DVF: Deformation Vector Field
	\item DWI: Diffusion Weighted Imaging
	\item FLAIR: Fluid-Attenuated Inversion Recovery
	\item GAN: Generative Adversarial Network
	\item MRI: Magnetic Resonance Imaging
	\item MoCo: Motion Correction
	\item MoDe: Motion Detection
	\item MoM: Multiple of the mean
	\item MSE: Mean Squared Error
	\item MS-SSIM: Multi-scale Structural Similarity Index
	\item NMSE: Normalized Mean Squared Error
	\item PE: Phase Encoding
	\item PSNR: Peak Signal-to-Noise Ratio
	\item SSIM: Structural Similarity Index
	\item \ce{T1}c: Postcontrast \ce{T1}-weighted
	\item \ce{T1}w: \ce{T1}-weighted

\end{itemize}

\section{Introduction}

Magnetic resonance imaging (MRI) is a non-invasive medical imaging technique that provides high-resolution anatomical and functional information without using ionizing radiation. However, the acquisition of high-quality MR images often requires long scan times, which increases the likelihood of image degradation due to both voluntary and involuntary patient motion. Such motion can alter the static magnetic field (\ce{B0})~\cite{motyka2024predicting}, induce susceptibility artifacts~\cite{safari2023patient}, affect spin history leading to signal loss~\cite{wood1985mr}, and cause inconsistencies in \textit{k}-space sampling that violate Nyquist criteria~\cite{zaitsev2015motion}. Motion artifacts are among the most prevalent sources of image degradation in MRI~\cite{sreekumari2019deep}. These artifacts can also compromise the performance of post-processing tasks, including target tracking in MR-guided radiation therapy~\cite{sui2024intra}, image segmentation~\cite{wangnicholas2022simulated, kemenczky2022effect}, and machine learning-based classification~\cite{hanson2024examining}.

Mitigating motion artifacts often necessitates repeating scans, which increases healthcare costs and contributes to patient discomfort. It is estimated that 15–20\% of neuroimaging exams require repeat acquisitions, potentially incurring additional annual costs exceeding \$300,000 per scanner~\cite{beljaards2024ai, slipsager2020quantifying, andre2015toward}. Therefore, the development of effective motion detection and correction strategies is essential to ensure diagnostic accuracy and improve healthcare efficiency.

Approaches to motion mitigation are typically classified into two broad categories: prospective and retrospective correction. Prospective motion correction methods attempt to compensate for motion during image acquisition. These include external optical tracking systems with reflective markers~\cite{stucht2015highest}, physiologic gating~\cite{scott2009motion}, and newer strategies using active NMR field probes~\cite{haeberlin2015real, aranovitch2018prospective, vionnet2021simultaneous}. Sequence-embedded navigators, such as PROMO and vNavs~\cite{white2010promo, tisdall2012volumetric}, and navigator-free methods using dynamic image reconstruction~\cite{zhu2020iterative} also fall within this category. Hybrid systems combining sensors and MR-based feedback have demonstrated increased robustness, particularly in challenging clinical scenarios~\cite{gumus2015comparison, van2019toward, vaculvciakova2022combining}.

Despite their utility, prospective methods face technical and logistical limitations. They often require hardware modifications, rigid coupling of sensors to the anatomy, or increased sequence complexity. These constraints can reduce their applicability in routine clinical settings and limit effectiveness for fast or non-rigid motion~\cite{zaitsev2015motion, slipsager2022comparison}.

In contrast, retrospective motion correction methods operate on data acquired during routine scans, without requiring additional hardware. Common techniques include rigid or non-rigid image registration, slice-to-volume reconstruction, and model-based reconstructions that jointly estimate both motion and image content~\cite{zaitsev2015motion, godenschweger2016motion}. These approaches remain essential in clinical practice because they can handle residual artifacts that persist despite prospective correction and are more adaptable to a range of motion patterns.

Recent developments in deep learning (DL), particularly those adapted from computer vision, have shown great promise in enhancing both prospective and retrospective motion correction. For prospective applications, convolutional neural networks (CNNs) have been used to estimate motion from image navigators or \textit{k}-space data with sub-second latency, enabling real-time feedback for acquisition control~\cite{9103624, shao2022real, neves2023real}. In retrospective settings, DL models can be trained to detect the presence and severity of motion artifacts, and to reconstruct motion-reduced or motion-free images using supervised, unsupervised, or unpaired learning strategies~\cite{duffy2018retrospective, kustner2019retrospective}. Unlike conventional iterative algorithms, DL-based motion correction models can learn direct mappings between corrupted and clean images, often yielding improved perceptual quality and reduced reconstruction time. These models are particularly powerful when integrated with generative architectures such as GANs, cGANs, CycleGANs, and diffusion models, which can capture complex image priors and correct non-linear distortions.

Previous reviews have primarily offered narrative syntheses of DL approaches for mitigating motion artifacts in MRI, typically organizing methods by task or application. For instance, one review explored network training strategies for motion correction and introduced a simulation tool, but provided only a qualitative overview rather than a quantitative synthesis~\cite{lee2020deep}. Another focused on rigid motion correction, summarizing model families and architectural choices, though without conducting a study-level meta-analysis~\cite{chang2023deep}. A further review examined retrospective learning-based correction across acquisition sequences and reconstruction stages, but again did not include statistical aggregation of findings across studies~\cite{spieker2023deep}. Similarly, a separate work addressed DL for brain MRI motion correction, emphasizing algorithmic advances and illustrative examples rather than pooled quantitative evidence~\cite{zhang2024motion}. In contrast, this study integrates a systematic review with meta-analysis, extracting study level variables and quantifying temporal trends in datasets, designs, and image quality metrics, thereby extending prior narrative syntheses.

Given the growing diversity of DL-based motion detection (MoDe) and correction (MoCo) methods, a systematic review and meta-analysis are needed to summarize recent developments, identify common patterns, and evaluate the effectiveness of existing approaches. This study presents the first comprehensive meta-analysis of AI driven MRI motion artifact detection and correction methods, with an emphasis on generative models. Section~\ref{sec:deep_learning_01} provides a review of DL model types, followed by motion simulation methods in Section~\ref{sec:motion_simulation_methods_01}, and an analysis of MoDe and MoCo approaches in Section~\ref{sec:motionCorrection_motionDetection_models_v01}. We then present the results of our meta-analysis in Section~\ref{sec:metaAnalysis_results_01} and discuss future research directions in Section~\ref{sec:disscussion_futureDirection}. Our key contributions are:

\begin{itemize}
	\item A unified framework summarizing AI-driven MoDe and MoCo strategies (Figure~\ref{fig:overalMoDeMoCoModels_01}).
	\item The first meta-analysis quantifying the performance of DL-based motion artifact detection and correction models.
	\begin{itemize}
		\item Assessment of image quality metrics (e.g., PSNR, SSIM), and comparison across studies.
		\item Temporal trends in dataset usage (public vs. institutional), model components, and evaluation metrics.
		\item Evaluation of hyperparameters such as learning rate, loss functions, and implementation frameworks.
	\end{itemize}
\end{itemize}

\section{Deep learning}\label{sec:deep_learning_01}

DL algorithms, a subset of machine learning, are particularly effective at modeling complex and nonlinear relationships, especially in computer vision and medical imaging tasks~\cite{lecun2015deep}. In medical imaging, DL models have been successfully applied to a wide range of applications, including anatomical segmentation~\cite{ronneberger2015u, 7785132, safari2024information}, image registration~\cite{8633930}, image enhancement~\cite{8736838, 7947200}, super-resolution~\cite{pham2019multiscale, rudie2022clinical, Safari_2025}, modality synthesis~\cite{chartsias2017adversarial, 8062235, eidex2024high}, and disease classification~\cite{litjens2017survey, esteva2017dermatologist, safari2022shuffle}.

The trainable parameters of a network are optimized to generate outputs that closely match the target outputs. In the context of motion artifact correction, the network parameters are finely adjusted to transform input motion-corrupted images into motion-free images. On the other hand, the MoDe network parameters are optimized to enhance the models in predicting the presence of motion artifacts and their severity level. This refinement process is known as training, during which the network adjusts its parameters based on the difference between the prediction and target ground   truth, such as motion-free images for MoCo and motion severity levels for MoDe. The loss function, which measures the discrepancy between actual and desired outputs, guides the network in updating its parameters to model optimal DL-based MoDe and MoCo processes effectively. 

Among the various types of DL models, deep generative models have the potential to revolutionize the field by enabling advanced data synthesis and representation learning. The ability of generative models to synthesize realistic and diverse motion-corrupted and motion-free data, as well as to capture complex distributions of motion patterns, will be crucial for advancing the robustness and generalizability of future motion correction and detection methods.

\subsection{Deep generative models for motion artifact correction}\label{subsec:deepGeneativeModels_for_moco}

Generative models are a class of DL methods designed to learn the underlying distribution of data so they can generate new samples that resemble the original dataset~\cite{murphy2023gan, goodfellow2020generative}. Unlike discriminative models, which learn decision boundaries to classify or distinguish between data points, generative models focus on capturing data structure and variability. Key families include variational autoencoders~\cite{kingma2013auto}, generative adversarial networks (GANs)~\cite{goodfellow2020generative}, and diffusion models~\cite{ho2020denoising}, each of which employs different mechanisms to approximate complex data distributions. In medical imaging, these models are especially relevant because they can synthesize realistic examples, augment limited datasets, and improve motion correction and detection by modeling non-linear motion-related distortions~\cite{yi2019generative}.

\subsubsection{Generative adversarial network}

GANs~\cite{goodfellow2020generative} have revolutionized the field of medical image processing, particularly in tasks such as image synthesis~\cite{lei2019mri}, segmentation~\cite{dong2019automatic}, denoising~\cite{ran2019denoising}, and harmonization~\cite{abbasi2024deep}. Beyond these specific applications, early surveys and foundational works have highlighted the versatility of GANs in medical imaging, including their use for cross-modality translation, data augmentation, and domain adaptation~\cite{yi2019generative, chartsias2017adversarial}.

GANs consist of two networks, including a generator and a discriminator, that are trained simultaneously in an adversarial framework to generate realistic data from input noise. The generator receives an input $z \sim q(z)$ and transforms it to $z^*$ using a network $G_\kappa$ with parameter $\kappa$. Simultaneously, the discriminator $D_\vartheta$ with parameters $\vartheta$ distinguish generated realistic image $z^*$ from real image $x \sim p^*(x)$. The discriminator objective is to minimize the Jensen-Shannon divergence between $q_\kappa$ (the likelihood function of the generated data $z^*$) and $p^*$ (the distribution of real data $x$)~\cite{murphy2023gan}, which leads to the following loss function:

\begin{equation}
	\underset{\kappa}{\min}\,\,\underset{\vartheta}{\max}\,\,\,\mathbb{E}_{x \sim p^*(x)} \left[\log D_\vartheta (x)\right] + \mathbb{E}_{z \sim q(z)} \left[1 - \log D_\vartheta (G_\kappa (z))\right]
	\label{eq:discriminator_loss_v01}
\end{equation}

GANs have been applied directly to remove brain motion artifacts~\cite{hewlett2024deep, wu2023image, yoshida2022motion}, and to define an adversarial regularizer to enhance the model performance~\cite{zhao20203d}.

However, GANs face limitations in controlling the traits of generated outputs and are prone to mode collapse, where the model produces a limited variety of outputs. Conditional GANs (cGANs) address these issues by introducing conditional inputs to guide the generation process and allow for specific control over the characteristics of the outputs. This approach not only provides greater control but also helps mitigate mode collapse by encouraging the generation of more diverse and varied samples based on the given conditions~\cite{mirza2014conditional}. Thus, several MoCo models have leveraged cGANs to reduce or remove motion artifacts~\cite{johnson2019conditional, usui2023evaluation,bao2022retrospective, ghodrati2021retrospective}. cGANs also have some limitations, namely the need for paired training data, which CycleGANs can help address. While cGANs require paired examples to learn the mapping from input to output (e.g., translating a motion-corrupted image to its motion-free counterpart), CycleGANs can learn this mapping using unpaired data. CycleGANs achieve this by introducing a cycle consistency loss, ensuring that an image translated from domain A to domain B and back to domain A remains unchanged, thus enabling effective domain translation without the need for paired datasets~\cite{zhu2017unpaired}. Eliminating the image pair requirement makes the CycleGAN training framework attractive for unsupervised MoCo models~\cite{oh2021unpaired, wu2023unsupervised, safari2024unsupervised, liu2021learning}.

\subsubsection{Denoising diffusion probabilistic model}
\begin{figure*}[tbh!]
	\includegraphics[width=\textwidth, draft=false]{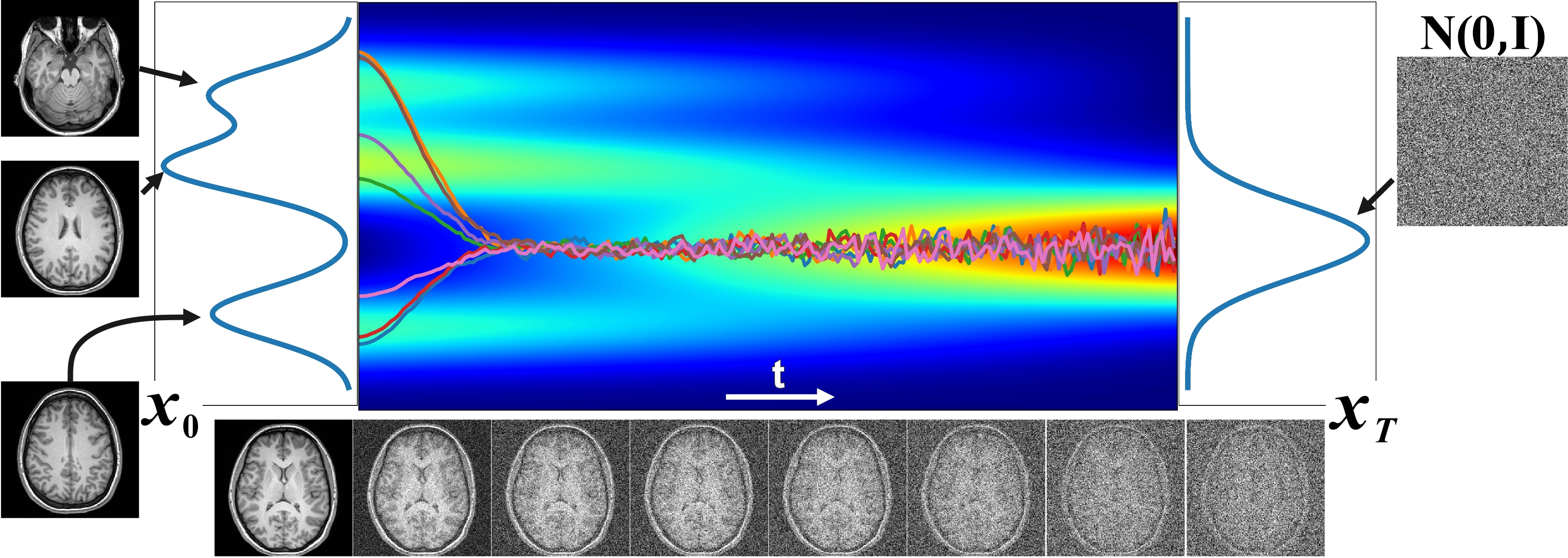}
	\centering
	\caption{Trajectory plot illustrating the evolution of samples from a three-mode Gaussian mixture distribution as they transition to a standard normal distribution $\mathcal{N}(\mathbf{0}, \mathbf{I})$. Here we simulated the Gaussian mixture as  $\sum_{i=1}^{3} \omega_i \mathcal{N}(\mu_i, \sigma_i)$ with $\mu = [-2, 2, 4]^T$ and $\sigma = [0.5, 0.5, 0.5]^T$ and forward diffusion step was performed over $T=1000$ steps.}
	\label{fig:ddpm_01}
\end{figure*}

The denoising diffusion probabilistic model (DDPM) is a generative model aimed at approximating complex intractable distributions with simple and tractable distributions, such as the normal Gaussian distribution~\cite{sohl2015deep}. The DDPM consists of two processes: a forward process and a reverse process. The forward process injects controlled Gaussian noise into the input image  over a large number of steps, denoted as $T$, until the image is transformed into normal Gaussian noise. This process heavily utilizes the first-order Markov process, which markedly reduces the computational time. On the other hand, the reverse process trains a deep neural network (DNN) to recover input image from normal Gaussian noise over $T$ steps as shown in Figure~\ref{fig:ddpm_01} by minimizing the following loss function:

\begin{equation}
	\underset{\kappa}{\arg \min} \parallel \epsilon_0 - \hat{\epsilon}_\kappa (x_t, t) \parallel_2^2
	\label{eq:ddpm_loss_01}
\end{equation}
where the neural network $\hat{\epsilon}$ with parameter $\kappa$ learns to predict the source noise $\epsilon_0 \sim \mathcal{N}(\epsilon\vert \mathbf{0}, \mathbf{I})$ that determines noisy data $x_t$ in step $t$ from noise-free image $x_0$.

Although the DDPM models were initially developed to generate images from normal Gaussian noise, they have also been used to remove motion artifacts from anatomical brain images~\cite{levac2024accelerated, safari2024mri}.

\subsubsection{Risks of hallucinations in generative models}
A critical challenge in applying generative models to medical imaging is the risk of hallucinations, where networks synthesize artificial features that appear realistic but do not correspond to the underlying anatomy. Such hallucinations can arise when the generator overfits to distribution-matching objectives or when training datasets are limited or biased. Although visually plausible, these artifacts may introduce false structures or remove clinically relevant features, which poses a significant risk of misdiagnosis in clinical practice~\cite{jung2024image, antun2020instabilities}.

Concrete examples of hallucinations have been demonstrated in prior work. Kazeminia et al.~\cite{kazeminia2020gans} illustrated spurious details in GAN-generated medical images, while Cohen et al.~\cite{cohen2018distribution} showed that distribution matching losses in image translation can hallucinate features, and Bhadra et al.~\cite{9424044} analyzed hallucinations in tomographic image reconstruction. Reviews of GAN applications in medical imaging have also highlighted hallucination risks as a major limitation for clinical deployment~\cite{yi2019generative}.

To mitigate these risks, strategies such as uncertainty quantification~\cite{kendall2017uncertainties}, physics-informed training objectives~\cite{ahmadi2025physics}, and adversarial regularization~\cite{https://doi.org/10.1002/mp.17675} have been proposed. However, further research is needed to develop standardized evaluation protocols that can reliably identify and quantify hallucinations in generated images. Addressing this limitation is essential to ensure that generative models can be safely integrated into clinical workflows.

\subsection{Supervised and unsupervised training frameworks}

Training MoCo and MoDe models can be broadly categorized into two approaches: supervised and unsupervised. The supervised approach relies on the availability of paired datasets where each motion-corrupted image has a corresponding motion-free counterpart. This enables MoCo models to learn the transformation from the input (motion-corrupted images) to the target (motion-free images). Similarly, in the context of MoDe, paired datasets are essential for accurately predicting the presence and severity of motion artifacts.

However, acquiring large, high-quality paired datasets is both challenging and costly, presenting a significant obstacle to the widespread implementation of supervised frameworks. To address this limitation, alternative frameworks that reduce the dependency on paired data have been developed. These alternatives can be divided into two categories: unpaired and unsupervised methods. Unpaired methods utilize motion-free datasets from the same MRI sequence to remove motion artifacts from similar but unpaired MRI sequences~\cite{oh2021unpaired, wu2023unsupervised,liu2021learning}. Despite their potential, these methods face practical challenges, such as the difficulty of acquiring motion-free images from different patients, which can raise clinical and privacy concerns, particularly regarding data sharing outside of hospital systems.

To address these limitations, unsupervised methods have been developed that eliminate the need for unpaired datasets. These methods leverage transfer learning techniques~\cite{chen2021ground} and auxiliary information from other MRI sequences~\cite{safari2024unsupervised} to achieve effective motion correction and detection without the requirements of the previous approaches. These advancements offer a promising direction for reducing the reliance on large-scale datasets and enhancing the feasibility of AI-driven motion artifact correction and detection in clinical settings.

\section{Motion simulation}\label{sec:motion_simulation_methods_01}
The supervised MoCo and MoDe models require a large dataset containing paired motion-free and motion-corrupted images, which is time-consuming and expensive to acquire~\cite{Dabrowski2022}. Thus, generating \textit{in-silico} motion-corrupted images can avoid the paired data requirement. In addition, motion simulation can serve as an augmentation technique to improve the generalization of DL models in tasks such as segmentation~\cite{shaw2020k, perez2021torchio}.

Patient motion induces phase shift in \textit{k}-space data, which results in discrete ghost artifacts along the phase encode direction when the motion is periodic in nature, respiratory and cardiac motion, and diffuse image noise when the motion is aperiodic, like peristalsis. Thus, realistic motion-corrupted images can be generated by randomly manipulating \textit{k}-space lines along the phase encoding (PE) and slab encoding directions, which are markedly slower to acquire than the lines along a readout direction~\cite{lee2020deep}. This section covers the simulation of abrupt and coherent motion artifacts that are apparent for brain, cardiac, and abdominal MR images, respectively. The motion simulation techniques are illustrated in Figure~\ref{fig:motion_simulation_01}.

\subsection{Head movement}

Rigid head movement is modeled using rotation and translation matrices. Translation movement is simulated by randomly shifting the \textit{k}-space lines along the PE direction $k_y$ as follows:

\begin{equation}
	\centering
	{Y}_{\text{distorted}}(k_x, k_y)=\left\{\begin{matrix}
		Y(k_x, k_y) e^{-j \phi (k_y)}  & k_y \in \mathcal{D} \\
		Y(k_x, k_y) &  \text{otherwise,}\\
	\end{matrix}\right.	
	\label{eq:translation_motion_artifact_01}
\end{equation}
where $Y$ and $ {Y}_{\text{distorted}} $ are motion-free and motion-corrupted \textit{k}-space images. The distortion phase $\phi(k_y)$ controls the distortion level, $\mathcal{D}$ is the randomly selected \textit{k}-space PE lines, and $k_y$ is the distorted line along PE, as shown by a dashed white line in Figure~\ref{fig:motion_simulation_01}A. The translations $T_{x,y,z}$ are randomly sampled to simulate the distortion phase $\phi (k_y)$ given below:

\begin{equation}
	\centering
	\phi (k_y)=\left\{\begin{matrix}
		k_y  \Delta & \vert k_y\vert > k_0 \\
		0 &  \text{otherwise,}\\
	\end{matrix}\right.	
	\label{eq:abrupt_translation_01}
\end{equation}
where $\Delta$ controls the motion artifact level and $k_0$ is a delay time of the phase error due to centric \textit{k}-space filling. The term $k_0$ refers to a threshold or specific point in \textit{k}-space, particularly in the PE direction, where the phase errors start to become significant. Before this point, the phase errors are either minimal or do not contribute substantially to the artifacts~\cite{tamada2020motion}.

Furthermore, abrupt rotational movements are generally simulated in both \textit{k}-space and the image domain. The rotation $R_{\theta_i}$ values are randomly sampled to rotate the images, and their corresponding \textit{k}-space data are sub-sampled using random masks $M_i$. Finally, the sub-sampled \textit{k}-space data are added to the \textit{k}-space data of motion-free images (red dashed line in Figure~\ref{fig:motion_simulation_01}A) to generate motion-corrupted data. This process is shown in Figure~\ref{fig:motion_simulation_01}A (lower panel), where the center of \textit{k}-space data was excluded from motion simulation.

\subsection{Abdominal and cardiovascular movement: coherent motion}

\begin{figure*}[tbh!]
	\includegraphics[width=\textwidth, draft=false]{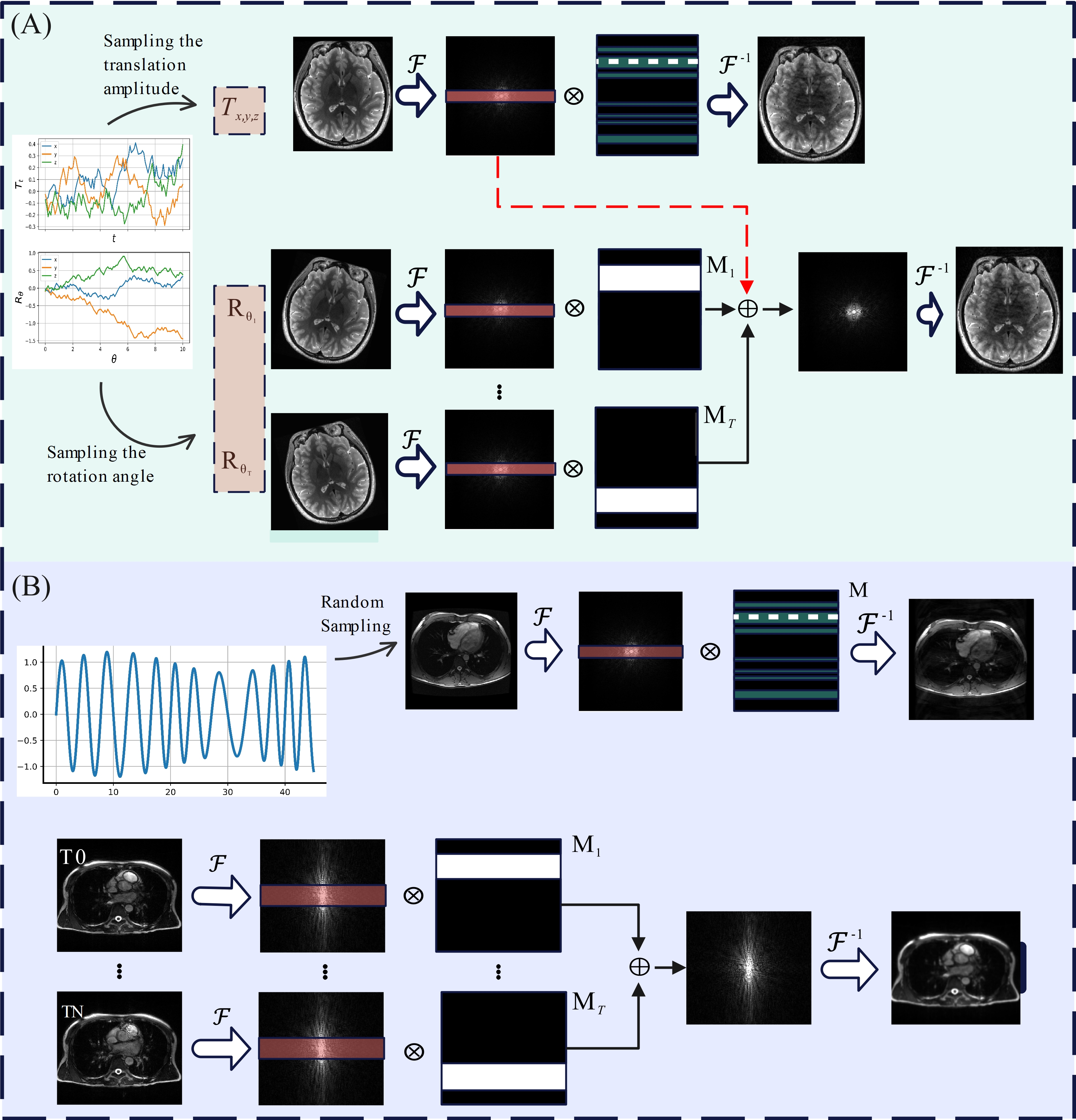}
	\centering
	\caption{Motion simulation methods are illustrated. (A) The abrupt brain movement involves rigid motion and rotation, simulated by randomly sampling translation $T_{x,y,z}$ and rotation $R_{\theta_i}$. (B) The abdominal and cardiovascular coherent movement simulation introduces blurring to the images. Motion simulation methods for static and dynamic MR images are shown. The center of \textit{k}-space data, shown by red slabs, is excluded from modifications.}
	\label{fig:motion_simulation_01}
\end{figure*}

Coherent motion typically occurs due to patients' involuntary movement, which can introduce significant blurring and ghosting artifacts. A random distortion phase $\phi(k_y)$ can be sampled from a quasi-sinusoidal as shown in Figure~\ref{fig:motion_simulation_01}B to simulate motion artifacts for static MR images. The coherent distortion phase is as follows:

\begin{equation}
	\centering
	\phi (k_y)=\left\{\begin{matrix}
		k_y  \Delta \sin(\alpha k_y + \beta) & \vert k_y\vert > k_0 \\
		0 &  \text{otherwise,}\\
	\end{matrix}\right.	
	\label{eq:coherent_motion_01}
\end{equation}
where $\alpha$ and $\beta$ are frequency and phase shift constants and $\Delta$ is a random distortion level.

To simulate coherent motion artifacts in dynamic MR images, random masks $M_i$ are used to sub-sample the images acquired at different times. The sub-sampled \textit{k}-space lines are then combined to generate motion-corrupted images. This method also excludes the center of the \textit{k}-space data to preserve the low-frequency content of the images.

\section{Motion Correction and Detection Models}\label{sec:motionCorrection_motionDetection_models_v01}
This section outlines the MoDe and MoCo methods. MoCo techniques are broadly classified into four categories: image-based, estimation-based, model-based, and other techniques where MoCo models serve as auxiliary tasks. MoDe methods are broadly categorized into two groups: models that predict the presence and severity of motion artifacts, and models designed to select the most appropriate downstream method for specific tasks, such as image reconstruction. These categories are illustrated in Figure~\ref{fig:overalMoDeMoCoModels_01}.

\subsection{Image-based MoCo}
\begin{figure*}[h!]
	\includegraphics[width=\textwidth, draft=false]{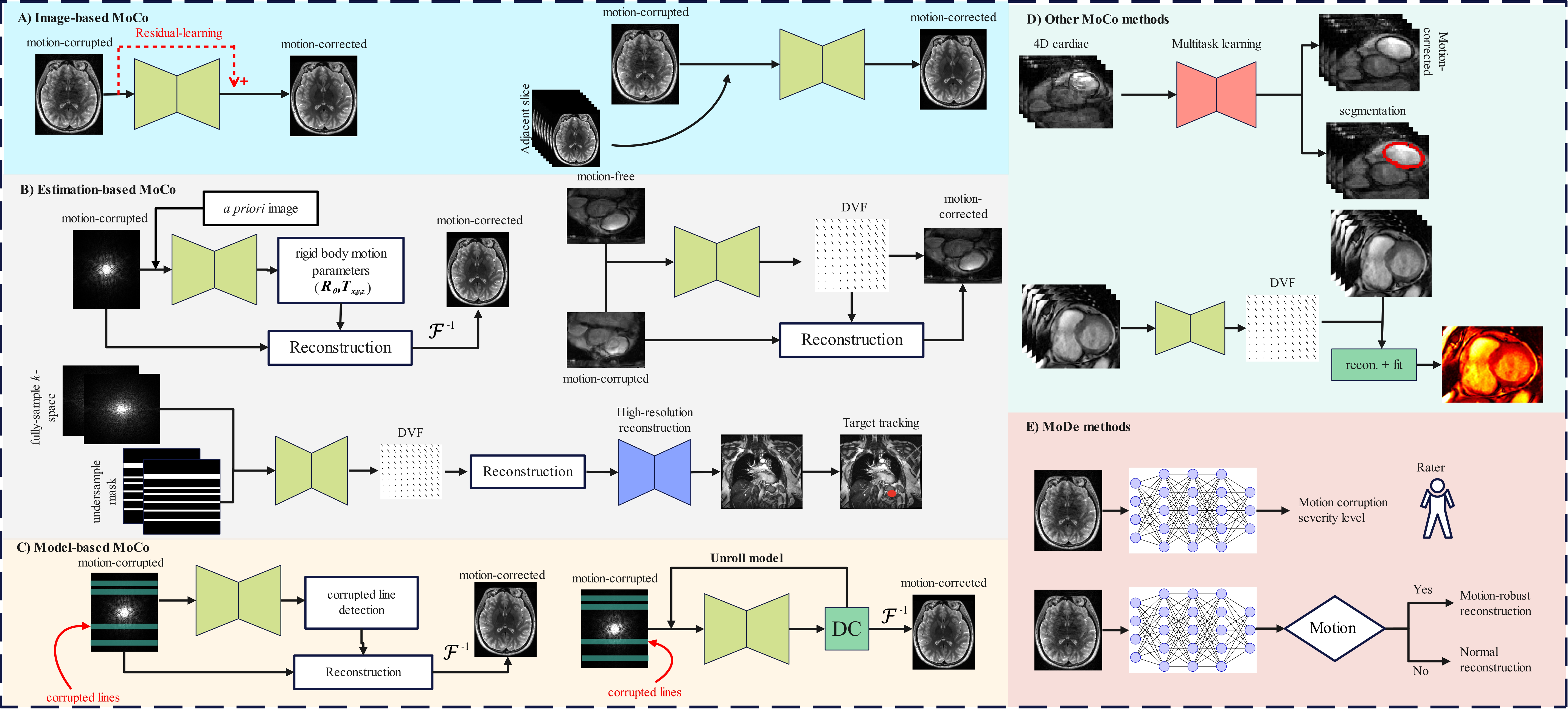}
	\centering
	\caption{Illustration of the overall DL-based MoCo and MoDe models. (A) left depicts an image-based training method that uses motion-corrupted and motion-free images, with an alternative approach focusing on reconstructing residual errors to generate motion-free images (red dashed line). (A) right shows an enhancement to this method using auxiliary images (data) to enhance the previous method. (B) presents motion estimation-based models, where a DNN in the left panel estimates the rigid motion parameters and another DNN in the right panel reconstructs the deformation vector field (DVF). This approach also includes the application of an DNN for DVF reconstruction from under-sampled \textit{k}-space, crucial for real-time target tracking in image-guided radiation therapy.(C) illustrates model-based methods trained on \textit{k}-space data. The left panel uses an DNN to estimate the amount of rigid motion, which is then used to reconstruct motion-free images iteratively. The right panel illustrates a method that unrolls the model into two modules, a denoiser DNN and data consistency module for motion artifact correction. The estimated motion parameters are used to correct the motion artifacts. (D) illustrates other types of motion-correction methods, including multi-task learning and quantitative MRI. (E) represent motion detection method, which are used either to detect motion artifacts or to select appropriate downstream tasks, such as motion-robust image reconstruction.}
	\label{fig:overalMoDeMoCoModels_01}
\end{figure*}

An image-based correction technique models the DL-based MoCo directly in the image domain, as illustrated in Figure~\ref{fig:overalMoDeMoCoModels_01}A. The motion artifact model is represented by:

\begin{equation}
	\mathbf{y} = \mathcal{F}^{-1} \mathbf{\mathcal{A}} \mathcal{F} \mathbf{x},
	\label{eq:motion_simulation_model_01}
\end{equation}
where $x \in \mathbb{R}^{N_x \times N_y \times N_z}$ is a motion-free image with $N_i$ for $i = \{x, y, z\} $ voxels, $\mathcal{F}$ is the Fourier transform, $\mathcal{A}$ is a motion-corruption model in \textit{k}-space, $\mathcal{F}^{-1}$ is the inverse Fourier transform, and $y\in \mathbb{R}^{N_x \times N_y \times N_z}$ is the motion-corrupted images. We assumed the images were real-valued  for simplicity. Image-based correction techniques address this inverse problem directly, mapping from motion-corrupted images ($y$) to motion-free images ($x$).

One advantage of this approach is its capacity to incorporate new advancements seamlessly and models from various DL domains. For instance, the U-net model was initially designed for segmentation tasks but was easily repurposed to produce motion-free images with minimal adjustments. Similarly, generative models such as GANs~\cite{goodfellow2020generative} and DDPM~\cite{ho2020denoising} were initially developed to generate images from noise and have since proven useful in mitigating motion artifacts~\cite{kustner2019retrospective, wu2023unsupervised, bao2022retrospective, ghodrati2021retrospective, reddy2024gan, eichhorn2023deep, safari2024mri, weng2024convolutional,pawar2020clinical, al2022stacked, pawar2022suppressing, safari2025mrimotioncorrectionefficient}.

However, a potential limitation of this approach is its reliance on large volumes of data, as well as the risk of introducing false visual perceptions, particularly in cases with severe motion artifacts. To address these challenges, image-based MoCo techniques have been developed that utilize auxiliary images or perform corrections across multiple adjacent image slices. This additional information guides the training process, as depicted in Figure~\ref{fig:overalMoDeMoCoModels_01}A (right panel)~\cite{al2023knowledge,zhang2024anti,lyu2021cine,lee2021mc2}.

\subsubsection{Residual learning}

Our literature review identified three studies that aimed to predict the difference between motion-corrupted and motion-free images (residual errors)~\cite{liu2020motion,tamada2020motion, pirkl2022learning}. The concept of residual learning suggests that networks produce improved motion-free images by predicting the residual errors with smaller variations than motion-free images. Furthermore, this method may bypass the constraint of preserving the soft-tissue contrast of motion-free images, potentially leading to the generation of sharper images. This approach is shown with a red dashed line in Figure~\ref{fig:overalMoDeMoCoModels_01}A (left panel).

\subsection{Estimation-based MoCo}
Estimation-based approaches, which involve estimating motion parameters using a DNN, can be categorized into two main groups: rigid-motion artifact estimation and deformable motion-artifact estimation methods. These methods can also serve as auxiliary tasks for reconstructing high-resolution images from down-sampled, motion-corrupted MRI data, thereby accelerating the imaging process - an essential factor for target tracking during radiation therapy. The general mathematical formulation for rigid-body motion artifact modeling is as follows~\cite{https://doi.org/10.1002/mrm.20656}:

\begin{equation}
	\mathbf{k}^n_{\text{dist}} = \sum_{j=1}^{J} \underbrace{\mathbf{M}^j \mathcal{F} \mathbf{C}^n \mathbf{T}^j_{x,y,z} \mathbf{R}^j_{\theta}}_{\mathbf{E}_{\theta^j}} \mathbf{x},
	\label{eq:rigidMoCo_01}
\end{equation}
where $ \mathbf{k}^n_{\text{dist}} \in \mathbb{C}^{N_x \times N_y \times N_z} $ motion distorted \textit{k}-space for a given coil $n$, $j$ represents the number of time steps that rigid distortion occurs, $ \mathbf{M}^j $ is a binary mask to select the corresponding \textit{k}-space line in step $j$, $\mathbf{C}^n$ is a sensitivity map of coil $n$, $\mathbf{T}^j_{x,y,z}$ is a translation matrix, and $\mathbf{R}^j_{\theta}$ is a rotation matrix. In rigid-body motion artifacts, such as those encountered in brain imaging, a DNN is trained to estimate rotation and translation parameters at each time step $\mathbf{E}_{\theta^j}$, with $\theta^j \in \mathbb{R}^{\text{DOF}}$. In 3D motion, the degree of freedom (DOF) is six (three translations and three rotations), while in the case of 2D in-plane rigid-body motion, only three parameters are needed (two translations and one in-plane rotation).

These parameters are then used in an iterative correction process where the initial motion estimates are used to correct the \textit{k}-space data to generate a preliminary image, which may still contain residual motion artifacts. The process iterates by refining the motion estimates based on the partially corrected image, reapplying these refined corrections to the \textit{k}-space data, and reconstructing the image again. This loop continues until the motion estimates converge, resulting in a final high-resolution image with minimal motion artifacts. The iterative approach ensures that even complex and non-linear motion is corrected accurately, leading to a clear and reliable final MR image~\cite{hossbach2023deep, dabrowski_SISMIK_2024, haskell2019network}. 

Considering that multiple MRI sequences are acquired subsequently in an single imaging session, this technique can be extended to use auxiliary input images--possibly from different sequences--to guide the training process of estimating $\theta$ and correct for rigid motion artifacts~\cite{rizzuti2022joint, rizzuti2024towards}. 

It is noteworthy that the mentioned method is particularly effective for correcting rigid motion artifact. To extend this method to the regions with non-rigid deformations voxel-wise deformation vector fields (DVFs) must be estimated using a DNN, as shown in Figure~\ref{fig:overalMoDeMoCoModels_01}B (right panel). The predicted DVF is then used to correct the non-rigid motion artifacts~\cite{morales2019implementation, gonzales2021moconet}. Furthermore, this technique can be combined with the acceleration algorithms to compensate for involuntary abdominal motions~\cite{terpstra2020deep,terpstra2021real,shao2022real} and super-resolution~\cite{zhi2023coarse, chen2024motion}.

\subsection{Model-based MoCo}

Given the limitations of image-based MoCo models, including the requirement for large datasets and the potential for hallucinations, model-based MoCo models incorporate a data acquisition model to remove motion artifacts. This approach can be categorized into two techniques corrupted \textit{k}-space line and methods that unroll the training process (see Figure~\ref{fig:overalMoDeMoCoModels_01}B right panel). 

In the former, a DNN is employed to detect \textit{k}-space lines affected by motion. The prediction results are then integrated into an iterative reconstruction process to reduce motion artifacts in the final images~\cite{eichhorn2023physics, yasaka2024iterative}. An alternative approach leverages a combination of DL models and \textit{k}-space analysis to identify motion-affected lines. This method filters the motion-corrupted images using a convolutional neural network, compares the filtered \textit{k}-space data with the original motion-corrupted \textit{k}-space, and reconstructs the final image using only the unaffected \textit{k}-space lines with compressed sensing, effectively mitigating motion artifacts~\cite{cui2023motion, safari2025physicsinformeddeeplearningmodel}.

The latter technique splits the models into a denoising network and data consistency layer. The denoiser reconstructs a first estimation of motion-free images to solve the data fidelity by a gradient descent method in an iterative fashion, as follows~\cite{wang2024deep}:

\begin{equation}
	\mathbf{x}_{t + 1} = \mathbf{x}_{t} - 2 \lambda \mathcal{A}^H (\mathbf{y} - \mathcal{A} \mathbf{x}_t) 
	\label{eq:dataFidelityReconstrunction_01}
\end{equation}
where $\mathcal{A} = \mathcal{F} \mathbf{C}$ is the encoding matrix, $\lambda$ is the learning rate to balance between the current estimation and the updated step size, $\mathcal{A}^H$ is Hermitian transform of $\mathcal{A}$, and $\mathbf{x}_{t}$ is the image at iteration $t$, with $\mathbf{x}_{0}$ being the output from the denoiser~\cite{wang2024deep, eichhorn2024physics}. 

\subsection{Other MoCo methods}

MoCo models can be modified to perform several tasks simultaneously since multi-task DL models tend to generalize better, and auxiliary tasks can enhance the model's performance on the primary task due to potential correlations between the tasks~\cite{zhen2016multi}. This adaptability allows for the integration of MoCo techniques with other DL-based approaches, such as segmentation~\cite{sui2024intra,pei2020anatomy, oksuz2020deep} or quantitative MRI reconstruction~\cite{xu2022learning,li2022motion}, resulting in multi-task models that achieve comprehensive results while maintaining high performance across each task.

\subsection{Image-based motion detection}

Image-based motion detection techniques focus on predicting motion artifacts directly using the image domain. This approach offers the advantage of incorporating advances from other domains of DL without requiring significant changes. For example, models like 3D-CNN~\cite{tran2015learning}, VGG16~\cite{simonyan2014very}, EfficientNet~\cite{tan2019efficientnet}, AlexNet~\cite{krizhevsky2012imagenet}, and ResNet~\cite{he2016deep} have been adapted effectively to classify natural images. These models were also used to predict the presence and severity of motion artifacts in cardiothoracic and brain regions~\cite{vakli2023automatic, oksuz2019automatic}.

Furthermore, MRI reconstruction techniques have been developed to mitigate the impact of patient movement during imaging~\cite{usman2013motion, arshad2024motion}. Inspired by this, DL-based methods have been proposed to assess the severity of motion artifacts and to adapt the reconstruction process accordingly. This approach involves predicting the level of motion artifact severity from undersampled MR images, which can then be used to accelerate acquisition~\cite{safari2024fast} and to determine whether a standard or a motion-robust reconstruction method should be employed~\cite{beljaards2024ai}. 

In line with this, a DL-based technique using a hierarchical convolutional neural network has been developed to reduce residual motion effects in diffusion MRI data~\cite{gong2021deep}. This method selectively rejects only the most severely motion-corrupted data, while retaining the remaining data for accurate diffusion parameter estimation, thus enhancing the robustness of MRI reconstructions even in the presence of significant motion artifacts.

\section{Meta-analysis method}\label{sec:metaAnalysis_01}
\subsection{Data collection}

We collected literature across PubMed to evaluate quantitatively the trend of DL-based MoCo and MoDe models. The search utilized the keywords: ``MRI'' or ``magnetic resonance imaging'' and ``motion'' and ``deep learning.'' We used the keyword ``motion'' in lieu of ``motion correction'' or ''motion reduction'' to broaden the search range. The first search was carried out in May 2024 but we have included new published papers after that date. We exported the references of the matched studies as a (PubMed) .txt file and imported them into Zotero Desktop.  

\paragraph{\textit{1. Initial filtering:}}Initially, 449 studies were selected that match the keywords. After removing 91 duplicates and irrelevant studies, 358 non-duplicated studies remained.

\paragraph{\textit{2. Title and Abstract screening:}} We reviewed the titles to determine the relevance of each study and screened out those that did not contain the keywords ``MRI,'' ``motion,'' ``deep learning,'' and ``reduction''. One reviewer with previous experience in DL and MRI motion correction examined the abstract section to determine the relevant studies. In case the Abstract section was generic, the Introduction section was examined. This process left us with 121 studies, which advanced to the next data collection stage.

\begin{figure}[tbh!]
	\includegraphics[width=0.9\textwidth, draft=false]{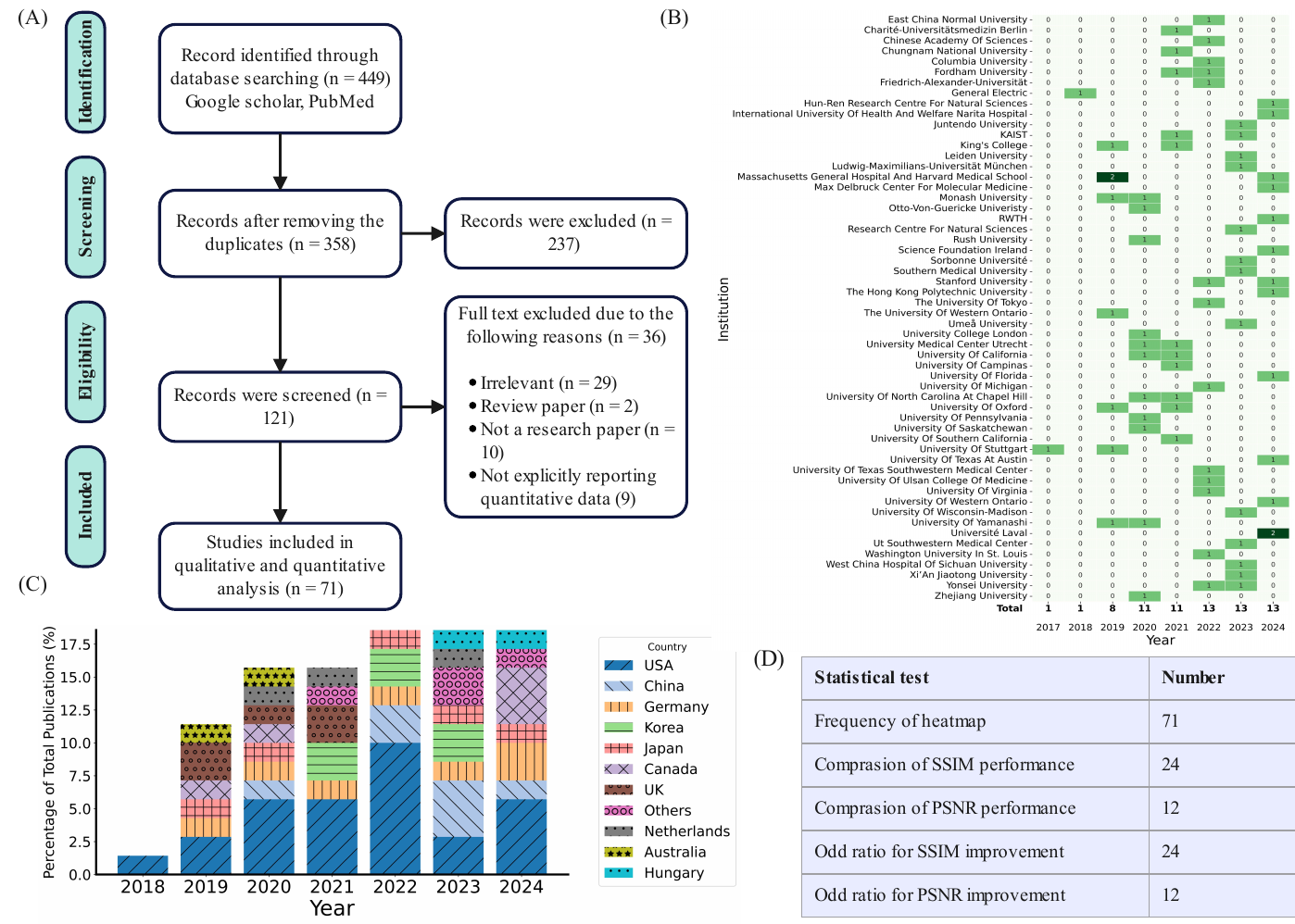}
	\centering
	\caption{Information of the reviewed studies. (A) The preferred reporting items for systematic reviews and meta-analyses (PRISMA) flowchart of this meta-analysis study~\cite{liberati2009prisma}. (B) Number of publications by research institute across years, illustrating the major contributing institutions and highlighting leading research hubs in DL-based MoCo and MoDe. (C) Number of publications by country, showing the global distribution and growth of research activity, which provides context on regional contributions and opportunities for collaboration. (D) Number of studies included for each type of statistical analysis performed in this review.}
	\label{fig:data_collection_01}
\end{figure}

\paragraph{\textit{3. Full-text screening}}We read the full text and inspected the results. Finally, 71 studies were included. The entire process is illustrated in Figure~\ref{fig:data_collection_01}A. 

\paragraph{\textit{4. Data collection:}}We modified the criteria given in the checklist for artificial intelligence in medical imaging (CLAIM) criteria~\cite{mongan2020checklist} (see Table~\ref{tab:comparative_reviews_table01}) to extract data from the studies. The modified table includes data critical for MoCo and MoDe studies, such as motion simulation methods, motion artifact levels, and quantitative metrics, which are not necessarily required for general-purpose medical image analysis.

\begin{table*}[tbh!]
	\caption{Our criteria to collecting data for each study by modifying the CLAIM criteria are summarized.}
	\label{tab:comparative_reviews_table01}
	\resizebox{\textwidth}{!}{%
		\begin{tabular}{p{2cm} p{3cm} p{14cm}}
			
			\hline

			\hline
			Category                & Item                           & Explanation                \\ \hline

			\hline

			& Dataset                        & What dataset(s) were used and how was collected \\
			
			& \cellcolor[HTML]{EFEFEF}Region & \cellcolor[HTML]{EFEFEF}Which region of the body was studied \\                                  
			& Ground truth & What were the ground truth, e.g. \textit{in-silico} or \textit{in-vivo}.\\
			
			\multirow{0}{*}{Data}    		    & \cellcolor[HTML]{EFEFEF} MRI sequence & \cellcolor[HTML]{EFEFEF}What MRI sequence(s) was included in the study (\ce{T1}, \ce{T2}, etc.)         \\ 
			
			& Motion simulation & What was the motion-simulation method and how the corresponding values were selected \\
			
			&\cellcolor[HTML]{EFEFEF} Partition& \cellcolor[HTML]{EFEFEF}  What was the partition strategy, percentage of training and testing partition data\\
			& Availability & Whether data are publicly available or not \\ 
			&\cellcolor[HTML]{EFEFEF} Augmentation&\cellcolor[HTML]{EFEFEF} What augmentation method was used, e.g. rotation, flipping, etc. \\
			\hline
			& Training  & What was the training method, e.g. image-based MoCo, estimation-based MoCo, model-based MoCo, and MoDe  \\
			\multirow{0}{*}{Model}  			& \cellcolor[HTML]{EFEFEF}Library & \cellcolor[HTML]{EFEFEF}What library was used to implement a proposed DL model, e.g. PyTorch and TensorFlow\\ 
			& Input domain & What was the input's domain, \textit{k}-space or image space \\
			& \cellcolor[HTML]{EFEFEF}Open source & \cellcolor[HTML]{EFEFEF}Whether the original implementation is completely available, and not partially\\
			&  Loss & What was the loss function and how many were used \\
			& \cellcolor[HTML]{EFEFEF}Optimizer & \cellcolor[HTML]{EFEFEF}What was the optimizer and the corresponding learning rate \\ 
			\hline
			
			& Metric                        & How many quantitative metrics were reported, and what were they, e.g. PSNR, SSIM, and NMSE\\
			& \cellcolor[HTML]{EFEFEF}Comparison & \cellcolor[HTML]{EFEFEF}What other models were the proposed model compared with\\                                  
			& Motion levels & How many motion artifact levels were the model evaluated for, e.g., minor, moderate, and heavy\\
			\multirow{-5}{*}{\begin{tabular}[c]{@{}l@{}}Evaluation\\ method\end{tabular}}  
			& \cellcolor[HTML]{EFEFEF} Testing mode & \cellcolor[HTML]{EFEFEF}What was the testing dataset, e.g. \textit{in-vivo} and \textit{in-silico} \\ 
			& External dataset & Whether an external dataset was used to evaluate the proposed model's generalization \\
			&\cellcolor[HTML]{EFEFEF} Diagnostic&\cellcolor[HTML]{EFEFEF}Whether a clinician evaluated the diagnostic accuracy of the MoCo models \\
			
			\hline

			& Novelty     & What could be the potential novelty of the proposed method\\
			\multirow{-2}{*}{\begin{tabular}[c]{@{}l@{}}Discussion\end{tabular}}  					
			& \cellcolor[HTML]{EFEFEF}Limitation & \cellcolor[HTML]{EFEFEF}What are the main limitation of the study\\

			\hline
			
		\end{tabular}%
		
	}
	
\end{table*}

\subsection{Data analysis}\label{sec:data_analysis}

\paragraph{\textit{1. Statistical rationale and analysis plan:}}
The study was the unit of analysis. Each patient or volunteer was counted once; image slices were not treated as independent samples. When a study reported development, evaluation, and testing sets, we combined evaluation with testing for summaries because both reflect out of sample performance. We distinguished 2D and 3D acquisitions due to differences in acquisition time and spatial resolution.

Reporting of dispersion was inconsistent across studies. Because many papers did not provide standard deviations or confidence intervals for PSNR and SSIM, we used two complementary strategies. First, for descriptive cross study comparison we transformed each metric to a dimensionless quantity, the multiple of the mean (MoM), defined as the study value divided by the corpus mean for that metric. MoM places heterogeneous scales on a common footing without assuming equal variances. We compared MoM distributions across publication years with nonparametric tests. Second, when a subset of studies reported dispersion, we ran sensitivity checks with meta regression on raw values and reached the same qualitative conclusions.

Temporal trend analyses served two goals. For continuous counts or proportions over time we report Pearson $r$ and Spearman $\rho$ to capture linear and monotonic associations. For year wise comparisons of sample sizes and MoM we used the Kruskal Wallis test because group sizes are unequal and normality is not guaranteed. To evaluate whether image quality improved over time, we fit ordinary least squares meta regression models with publication year as the predictor and PSNR or SSIM as the response, and we report heteroskedasticity robust standard errors.

\paragraph{\textit{2. Development trend:}}
We assessed temporal trends in publication volume and other continuous summaries using Pearson correlation coefficients and also report Spearman correlations to capture monotonic relations that may not be linear. Year wise comparisons of numerical variables, such as training and testing sample sizes, were performed with the Kruskal Wallis test because group sizes are unequal and normality is not guaranteed. All analyses used \texttt{scipy.stats 1.13.1}~\cite{virtanen2020scipy} and \texttt{statsmodels.api 0.15.0} (\url{https://github.com/statsmodels/statsmodels/}) in Python 3.10.14.

In our analysis, each patient or volunteer was treated as an individual sample rather than considering image slices. When a study divided data into training, evaluation, and testing sets, we incorporated the evaluation dataset into the testing dataset size. Additionally, we distinguished between 3D MRI and 2D MRI images, such as \ce{T1}-weighted (\ce{T1}w) versus 3D \ce{T1}w, due to differences in acquisition time and spatial resolution.

\paragraph{\textit{3. Performance:}}We divided each performance metric by its average across all studies to obtain the multiple of the mean (MoM), which reflects the relative improvement of models over time. We reported the average MoM values when models were tested under different motion artifact levels. The Kruskal-Wallis test was used to calculate \textit{p}-values for MoM across years.

\section{Meta-analysis results}\label{sec:metaAnalysis_results_01}

Out of the 71 reviewed studies that met our meta-criteria shown in Figure~\ref{fig:data_collection_01}A and \ref{fig:data_collection_01}D, we observed increasing number of publications between 2018 and 2024, with the USA contributing the most to the research studies (see Figure~\ref{fig:data_analysis_01}C). Massachusetts General Hospital and Harvard Medical School was identified as the institution with the highest number of publications as shown in Figure~\ref{fig:data_collection_01}B.

\subsection{Dataset characteristic}

Training and evaluation of MoCo and MoDe models require ground truth data. For MoDe, ground truth typically consists of quality scores (e.g., between zero and five) assigned by radiologists, because simulated datasets might not capture the full diversity of real motion artifacts and scanner-dependent variations. In contrast, MoCo datasets can often be generated by simulation methods (see Section~\ref{sec:motion_simulation_methods_01}), which eliminates the necessity for paired motion-free and motion-corrupted acquisitions.

\begin{figure}[tbh!]
	\centering
	\includegraphics[width=\textwidth, draft=false]{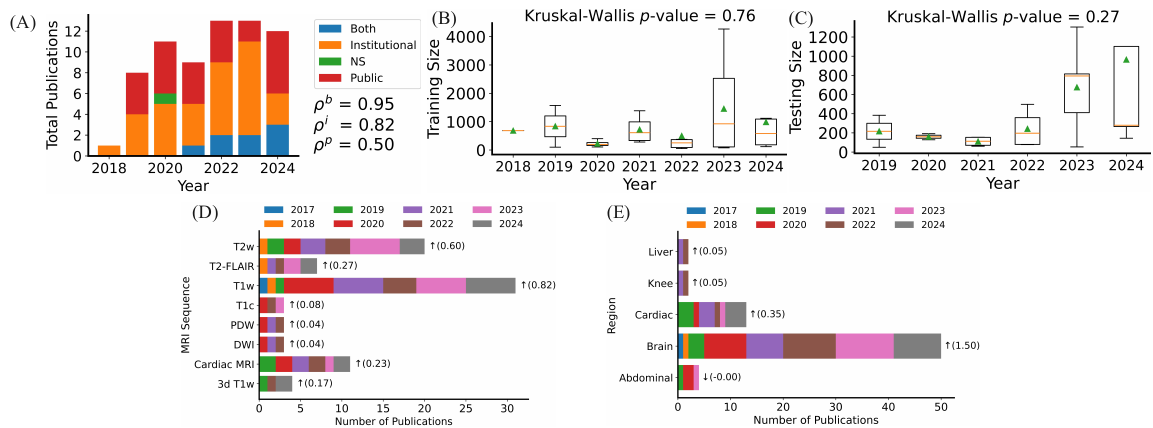}
	
	\caption{Dataset information from the reviewed studies. (A) The total number of publications using institutional, public, and both types of datasets over time is illustrated. The Pearson correlation between the number of publication uses public, private, and both with time are $\rho ^p$, $\rho ^i$, and $\rho ^b$, respectively. Abbreviation: NS: Not specified. (B) and (C) Training and testing sample size are illustrated over time. The Kruskal-Wallis \textit{p}-values were reported for each box plot. (D) MRI sequences used by the studies grouped by year of publication are illustrated. Abbreviations: FLAIR: fluid-attenuated inversion recovery, T1c: postcontrast \ce{T1}-weighted, PD: proton density, DWI: diffusion-weighted imaging. (E) Anatomical region of the images used by the studies. The arrow shows the usage trends over time and the numbers inside parenthesis represents the slope value obtained from linear regression represents the rate of change in the number of publications per unit change in the year.}
	\label{fig:data_analysis_01}
\end{figure}

In terms of dataset usage, 49.3\% of the studies utilized institutional datasets exclusively, 37.3\% used public datasets exclusively, and 11.9\% used both institutional and public datasets. The most frequently utilized public datasets were fastMRI~\cite{zbontar2018fastmri} and movement-related artifacts (MR-ART)~\cite{narai2022movement}, followed by the Human Connectome Project (HCP), UK Biobank~\cite{bycroft2018uk}, and IXI ({https://brain-development.org/ixi-dataset/}) datasets. These datasets and the corresponding studies are listed in Table~\ref{tab:public_dataset_used_studies}. There is a noticeable increase in the use of public datasets in 2024, as depicted in Figure~\ref{fig:data_analysis_01}A. The Pearson correlation coefficients for the use of institutional, public, and combined datasets are 0.82, 0.50, and 0.95, respectively. These values show the strength of the linear relationship between time and the number of studies using each dataset type. The institutional datasets exhibit a robust positive correlation of 0.82, indicating a significant increase in the number of studies utilizing institutional datasets over time. Public datasets show a moderate positive correlation of 0.50, suggesting a steady but less pronounced growth in the number of studies using public datasets over time. In contrast, combined datasets have a robust positive correlation of 0.95, indicating a nearly perfect increase in the number of studies integrating institutional and public datasets over time. These correlations highlight that while the use of institutional datasets alone is growing robustly, there is an increasing trend towards integrating both institutional and public datasets, reflecting a shift towards more comprehensive and multifaceted data usage in research studies over the observed timeframe.

\begin{table}[tbhp!]
	\centering
	\caption{Public datasets used by the studies.}
	\label{tab:public_dataset_used_studies}
	\resizebox{0.85\textwidth}{!}{%
		\begin{tabular}{llll} 
			\hline
			\textbf{Dataset}          & \textbf{Sample Size}  & \textbf{Region}           & \textbf{Used by}   \\ \hline
			fastMRI~\cite{zbontar2018fastmri}                   & 8,400 scans                  & Brain, knee   &          \cite{levac2024accelerated, beljaards2024ai,hewlett2024deep,oh2021unpaired,simko2023improving,wang2023dual}          \\ 
			
			\rowcolor[HTML]{EFEFEF}MR ART~\cite{narai2022movement}                    & 148 subjects                 & Brain     &               \cite{beljaards2024ai,safari2024mri, safari2024unsupervised, olsson2024simulating,kemenczky2022effect,belton2024towards}       \\ 
			
			HCP~\cite{van2012human}                       & 1,200 subjects       &                   Brain                     &      \cite{zhao20203d, bao2022retrospective,oh2021unpaired,wu2023unsupervised,ettehadi2022automated}                 \\ 
			
			\rowcolor[HTML]{EFEFEF}UK Biobank~\cite{sudlow2015uk}                   & 100,000 subjects                       & \begin{tabular}[c]{@{}l@{}}Whole body\\  (focus on brain, heart)\end{tabular} &        \cite{vakli2023automatic,oksuz2019automatic, gonzales2021moconet,chen2024motion,kemenczky2022effect}              \\ 
			
			IXI                       & 600 subjects         &                  Brain                     &             \cite{mohebbian2021classifying,pawar2020clinical,cui2023motion,belton2024towards,chatterjee2020retrospective}         \\ 
			
			\rowcolor[HTML]{EFEFEF}ABIDE~\cite{di2014autism, di2017enhancing}                     & 2,226 subjects       &                  Brain                     &      \cite{zhao20203d, li2022addressing,duffy2018retrospective,zhao2021localized}       \\ 
			
			OSAIS~\cite{marcus2007open}                     & 565 subjects                  &                  Brain                      &           \cite{li2022addressing, vakli2023automatic,zhao2021localized}           \\ 
			
			\rowcolor[HTML]{EFEFEF}ACDC~\cite{bernard2018deep}                      & 150 patients         &                  Heart                     &       \cite{lyu2021cine,morales2019implementation,ettehadi2022automated}              \\
			
			ADNI~\cite{petersen2010alzheimer}                      & 819 subjects       &                   Brain                     &        \cite{shaw2020k,loizillon2024automatic}            \\ 
			
			\rowcolor[HTML]{EFEFEF}GLIS-ART~\cite{shusharina2021glioma}                  & 230                                  & Brain                       &         \cite{safari2024mri, safari2024unsupervised}           \\ 
			
			MSSEG~\cite{commowick2018objective}                     & 53 patients          &                   Brain (Multiple Sclerosis) &       \cite{loizillon2024automatic}            \\ 
			
			\rowcolor[HTML]{EFEFEF}MNI BITE~\cite{mercier2012online}             & 14 subjects                          & Brain                     &     \cite{loizillon2024automatic}            \\ 
			
			IBSR~\cite{cocosco5online}                      & 18 subjects          &                   Brain                     &       \cite{mohebbian2021classifying}           \\ 
			
			\rowcolor[HTML]{EFEFEF}Forstmann, et al.~\cite{forstmann2014multi}                     & 30 subjects                  &                   Brain                       &         \cite{johnson2019conditional}         \\ 
			
			1000BRAINS~\cite{caspers2014studying}                & 1,000 subjects       &                   Brain                     &    \cite{roecher2024motion}             \\ 
			\hline
		\end{tabular}
	}
\end{table}

On average, studies used 65\% and 35\% of data for training and evaluation. The Kruskal-Wallis tests indicate the train and evaluation cohort populations did not significantly change overtime with p-values of 0.57 and 0.76 , respectively (see Figure~\ref{fig:data_analysis_01}B and C). The big jumps in cohort population in 2023 and 2024 are due to the MoDe studies that utilized several large public datasets. To increase the data population, eight studies employed data augmentations including mainly random translation, flip, and rotation. One study used methods such as adding random noise and bias fields to further increase the data population~\cite{terpstra2020deep}. Patch-based model training was utilized in two studies to increase data samples~\cite{liu2020motion, haskell2019network}.

Among the MRI sequences, \ce{T1}w, \ce{T1}c, and CMR were the most frequently used, with 30, 30, and 10 studies, respectively, reporting their use (see Figure~\ref{fig:data_analysis_01}D). Linear regression analysis indicated an increasing trend in their usage, as shown by the arrows. Among the imaging regions, 50 studies utilized brain datasets, and 10 studies used cardiac datasets (see Figure~\ref{fig:data_analysis_01}E), with a higher usage tendency for brain and cardiac datasets, as calculated by linear regression.

\subsection{Design}

\begin{figure}[tbh!]
	\includegraphics[width=\textwidth]{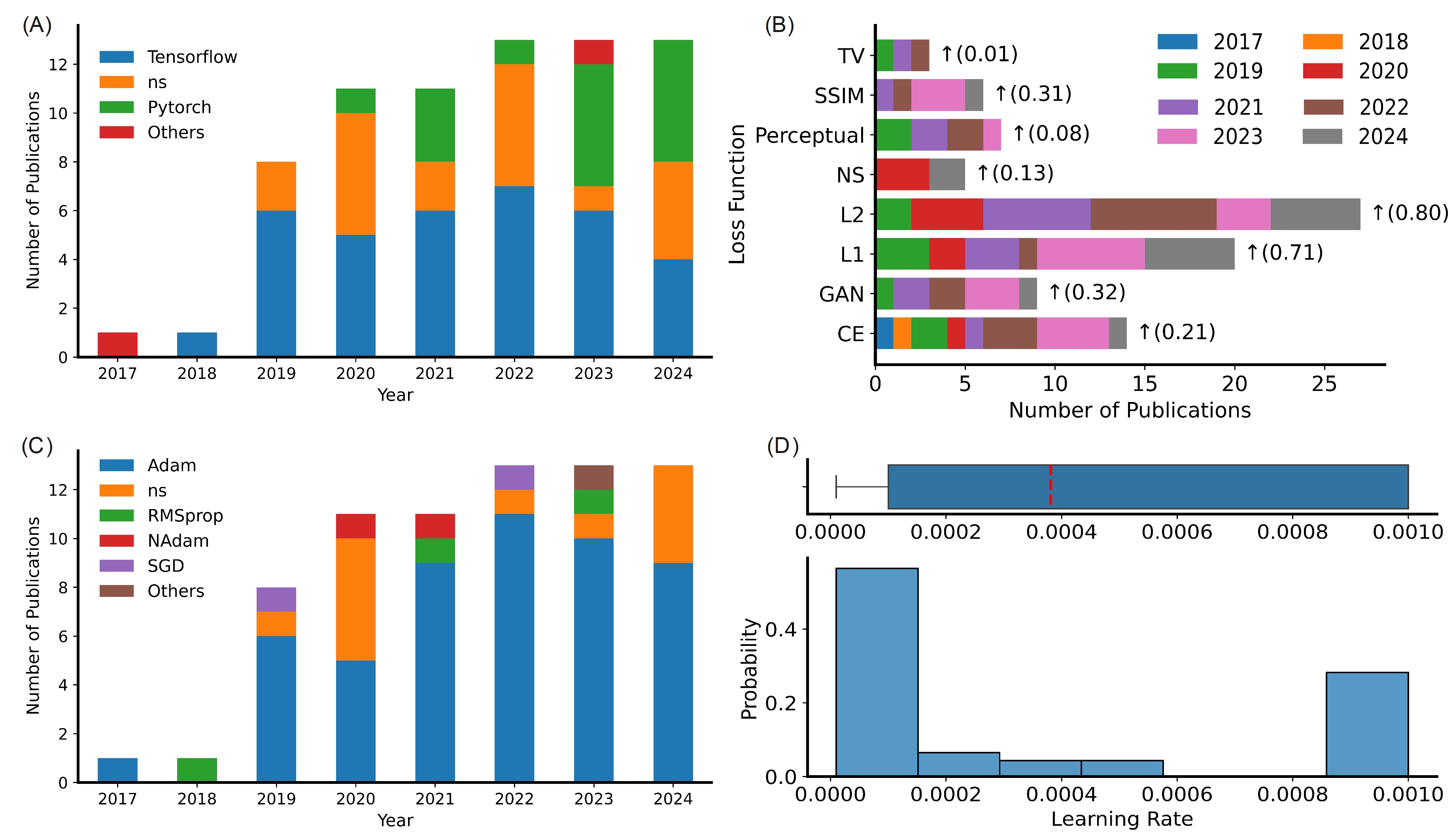}
	\centering
	\caption{Neural network design traits. (A)  DL libraries, (B) loss functions, and (C) Optimizers used by different studies. The arrow indicate the trend of the variable usage over time. The number inside the parenthesis represents the slop value calculated using logistic regression. (D) Distribution of the learning rate used by the studies over time and the red dashed line represents the mean value.}
	\label{fig:neural_network_design_01}
\end{figure}

\paragraph{\textit{1. Deep learning library:}}TensorFlow~\cite{tensorflow2015whitepaper} was the most commonly used library, employed in 35 studies. PyTorch~\cite{paszke2019pytorch} was the second most used, with 19 studies. MATLAB and Keras (without TensorFlow as the backend) were each used in only one study. Logistic regression analysis indicates a higher tendency toward the use of PyTorch compared to TensorFlow, as illustrated in Figure~\ref{fig:neural_network_design_01}A.

\paragraph{\textit{2. Loss function:}}The loss or cost functions used in the studies are depicted in Figure~\ref{fig:neural_network_design_01}B and summarized in Table~\ref{tab:lossFunctionSummarizehere}. L2 and L1 loss functions were the most frequently used, utilized by 26 and 25 studies, respectively~\cite{levac2024accelerated,zhao20203d, shaw2020k,beljaards2024ai}. Cross entropy, used in 14 studies, ranked second in usage. The L1 and L2 loss functions were predominantly used by the MoCo models, and cross-entropy was used by the MoDe models~\cite{gao2023lightweight, sreekumari2019deep,weaver2023automated,pawar2020clinical}.

While the L2 loss function is computationally inexpensive, it often results in overly smoothed images because it penalizes large errors more severely due to its squared term. In contrast, the L1 loss, which is also computationally efficient, was employed by a many studies and is know to produce sharper images compared to L2 loss. To improve the perceptual quality of reconstructed images, Perceptual loss has been utilized by minimizing the difference in image embeddings derived from pretrained networks such as VGG16~\cite{simonyan2014very}, though this approach is computationally expensive. The structural similarity index (SSIM)~\cite{wang2004image} loss function was also utilized as a less computationally intensive method to preserve perceptual quality~\cite{al2023knowledge}. Despite their advantages, both Perceptual and SSIM losses might be less sensitive to fine details, necessitating the use of additional L1 or L2 loss functions~\cite{lyu2021cine, weng2024convolutional}.

\paragraph{\textit{3. Optimizer:}} The optimizers used by the studies over time are illustrated in Figure~\ref{fig:neural_network_design_01}C. The Adam optimizer~\cite{kingma2014adam} was the most commonly used, employed in 50 studies. Linear regression analysis indicates a positive tendency in the increasing popularity of the Adam optimizer over time. Additionally, the histogram of learning rate values used by the studies is illustrated in Figure~\ref{fig:neural_network_design_01}D, with the red line indicating a mean learning rate value of $4\times 10^{-4}$. We reported the initial learning rate values when a study used learning rate schedulers to modify them. The purpose of Figure~\ref{fig:neural_network_design_01}D is not to compare learning rates across different architectures and optimizers, but rather to provide an overview of the distribution of initial learning rates reported in the literature, thereby reflecting common practices in hyperparameter selection. In contrast, SGD and AdamW optimizers were used less frequently.

\begin{table*}[tbh!]
	\centering
	\caption{Summary of Loss Functions Used in AI for Motion Artifact Detection and Correction.}
	\label{tab:lossFunctionSummarizehere}
	\resizebox{\textwidth}{!}{%
		\begin{tabular}{p{3.5cm}p{7cm}p{12cm}}
			\hline
			\textbf{Loss Function} & \textbf{Mathematical Formula} & \textbf{Pros and Cons} \\ \hline
			Total Variation (TV) & \begin{tabular}[c]{@{}l@{}}$\sum_{i,j} \sqrt{(x_{i+1,j} - x_{i,j})^2 + (x_{i,j+1} - x_{i,j})^2}$\\  \textit{where $x$ is the image, $i,j$ are pixel} \\ \textit{indices}\end{tabular}   & \textbf{Pros:} Effective in reducing noise and smoothing motion artifacts while preserving edges, making it suitable for motion correction tasks. \newline \textbf{Cons:} May oversmooth images, potentially eliminating fine motion details important for accurate detection and correction. \\

			\rowcolor[HTML]{EFEFEF}SSIM (Structural Similarity Index) &  \begin{tabular}[c]{@{}l@{}} $ \frac{(2\mu_x\mu_y + c_1)(2\sigma_{xy} + c_2)}{(\mu_x^2 + \mu_y^2 + c_1)(\sigma_x^2 + \sigma_y^2 + c_2)}$ \\ \textit{where $\mu$ is mean, $\sigma$ is variance,} \\ \textit{$c_1, c_2$ are constants} \end{tabular}    & \textbf{Pros:} Preserves structural and perceptual quality, enhancing the integrity of corrected images in motion compensation. \newline \textbf{Cons:} Computationally intensive and may be less sensitive to subtle motion artifacts, possibly overlooking minor corrections needed. \\ 
			
			Perceptual Loss &  \begin{tabular}[c]{@{}l@{}} $\sum_{i} \| \phi_i(x) - \phi_i(y) \|_2^2$ \\ \textit{where $\phi_i$ is the feature map of the $i$-th}\\\textit{layer of a pretrained network such as} \\ \textit{VGG16~\cite{simonyan2014very}} \end{tabular}  & \textbf{Pros:} Aligns corrected images with human perception by utilizing deep feature representations, improving motion correction quality. \newline \textbf{Cons:} Depends on pretrained models not specialized for motion correction, and increases computational load due to deep network processing. \\ 
			
			\rowcolor[HTML]{EFEFEF}L2 Loss &  \begin{tabular}[c]{@{}l@{}} $ \|x - y\|_2^2$  \\ \textit{where $x$ is the predicted image, $y$} \\ \textit{is thetarget image}\end{tabular}  & \textbf{Pros:} Simple to implement with smooth gradients, facilitating the training of models for motion correction. \newline \textbf{Cons:} Over-penalizes large errors from motion artifacts, potentially leading to blurred images and loss of important details. \\

			L1 Loss & \begin{tabular}[c]{@{}l@{}} $ \|x - y\|_1$ \\ \textit{where $x$ is the predicted image, $y$ is} \\ \textit{the target image} \end{tabular}  & \textbf{Pros:} More robust to outliers introduced by motion, better at preserving sharp edges during correction. \newline \textbf{Cons:} May produce less smooth results, leaving residual artifacts after motion correction. \\

			\rowcolor[HTML]{EFEFEF}GAN Loss & \begin{tabular}[c]{@{}l@{}} $ \mathbb{E}_{x}[\log D(x)] + \mathbb{E}_{z}[\log (1 - D(G(z)))]$ \\  \textit{where $G$ is the generator, $D$ is the}\\ \textit{discriminator, $x$ is real data, $z$ is noise} \end{tabular}  &  \textbf{Pros:} Generates highly realistic images by learning the distribution of motion-free data, effectively correcting complex motion artifacts. \newline \textbf{Cons:} Training can be unstable and prone to mode collapse, requiring careful tuning and increasing complexity in motion correction models. \\

			Cross Entropy Loss & \begin{tabular}[c]{@{}l@{}}  $ -\sum_{i} p(i) \log q(i)$  \\ \textit{where $p$ is the true distribution,} \\ \textit{$q$ is the predicted distribution} \end{tabular} &  \textbf{Pros:} Effective for motion detection when framed as a classification problem, distinguishing between motion and motion free regions. \newline \textbf{Cons:} Sensitive to class imbalance common in motion detection tasks, potentially leading to biased detection models. \\ \hline
		\end{tabular}
	}
\end{table*}

\subsection{Evaluation metrics and performance}

We also reported the evaluation metrics that were reported by the studies. Quantitative metrics were typically reported for \textit{in-silico} datasets where ground truth motion-free images are available. However, when ground truth data is not available, raters graded image quality for motion-corrected brain images~\cite{gonzales2021moconet} and cardiovascular images~\cite{kromrey2020reduction}. This method is subjective and is better used in conjunction with quantitative results~\cite{safari2024unsupervised}.

\begin{figure*}[tbh!]
	\includegraphics[width=\textwidth, draft=false]{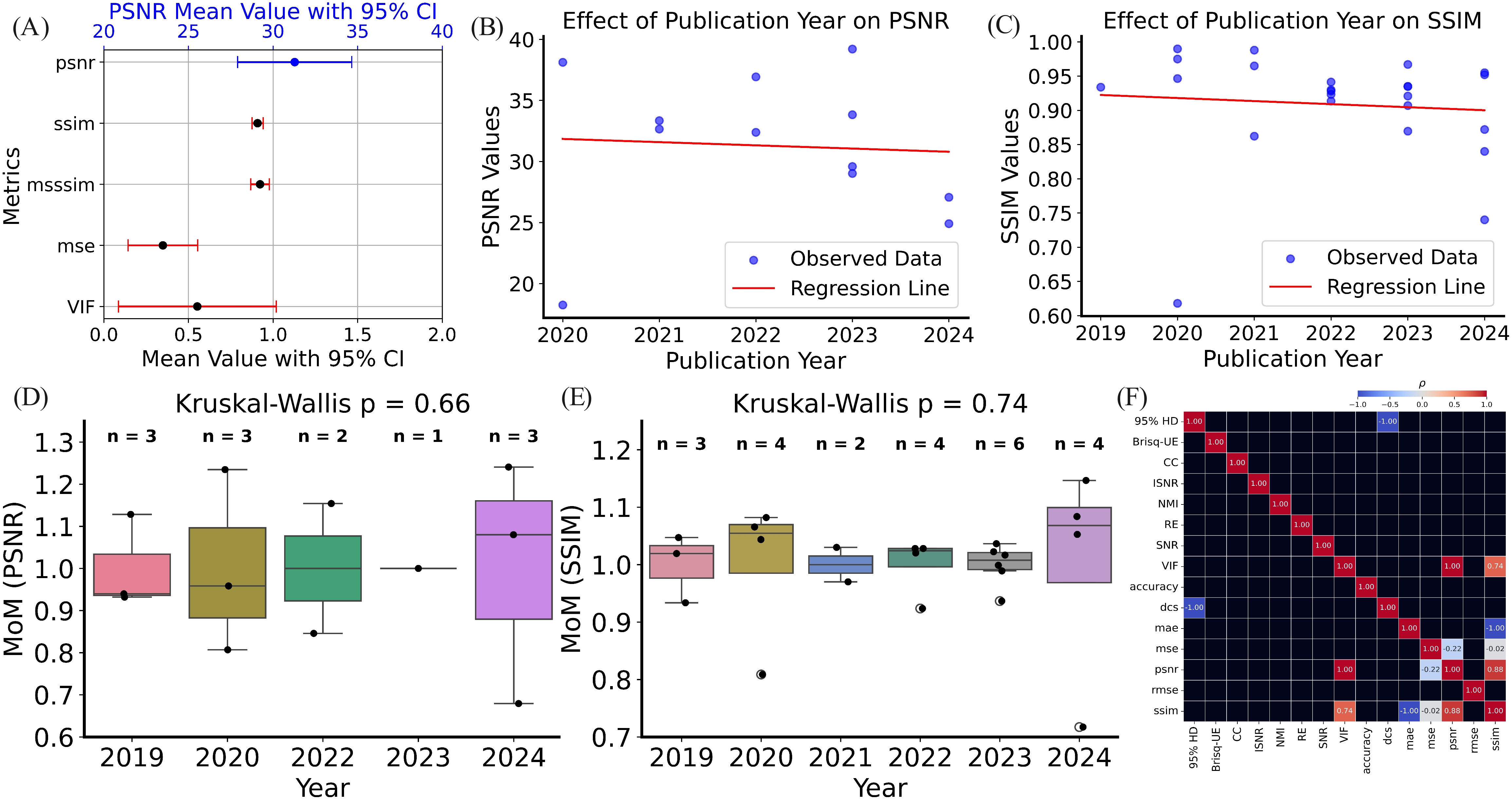}
	\centering

	\caption{Performance improvement relative to the average values of metrics. (A) Average values with 95\% CI are given for PSNR, SSIM, MS-SSIM, MSE, and VIF. These metrics were selected because they appeared more than three times across the reviewed studies. Due to the different value ranges, the PSNR axis is shown separately on the top in blue. (B) and (C) illustrate meta regression results for PSNR and SSIM to assess whether publication year accounts for variance in these metrics; we fit ordinary least squares meta regression with publication year as the predictor and PSNR or SSIM as the response and report heteroskedasticity robust standard errors (Section~\ref{sec:data_analysis}). (D) MoM of PSNR over time is illustrated. The numbers above the box plots indicate the number of studies published in that year. (E) MoM of SSIM over time. (F) Pearson correlation ($\rho$) heatmap between pairs of metrics. Black regions indicate where $\rho$ values are not available due to small sample size. MoM is defined as the study value divided by the corpus wide mean of that metric.}
	\label{fig:odd_ratio_v01}
\end{figure*}

Figure~\ref{fig:odd_ratio_v01}A presents a comparison of the mean values and 95\% confidence intervals for various image quality metrics, including PSNR, SSIM, multi-scale structural similarity index (MS-SSIM), mean squared error (MSE), and visual information fidelity (VIF). The metrics are plotted on the y-axis, with their mean values with associated confidence intervals displayed on the x-axis: downward SSIM, MS-SSIM, MSE, and VIF metrics, and upward for the PSNR metric. PSNR is highlighted as the primary metric, with a mean value centered around 30 dB, which is consistent with the typical values reported in the literature. The narrow confidence intervals for PSNR indicate a relatively low variability among the reported values, whereas other metrics like MSE exhibit wider confidence intervals, suggesting greater variability.

The statistical analysis for the trend of PSNR values over publication years, as depicted in Figure~\ref{fig:odd_ratio_v01}B, reveals an \ce{R^2} value of $0.004$, indicating that only $0.4\%$ of the variance in PSNR values can be explained by the publication year. The F-statistic of $0.03831$, with an associated \textit{p}-value of $0.849$, confirms that the regression model is not statistically significant, suggesting that there is no significant linear relationship between publication year and PSNR values in the observed data. The results imply that, despite the slight downward trend observed in the scatter plot, publication year does not have a meaningful impact on PSNR values. The same trend is observed for SSIM. The statistical analysis for SSIM values over publication years (see Figure~\ref{fig:odd_ratio_v01}C) shows an \ce{R^2} value of $0.007$, suggesting that only 0.7\% of the variance in SSIM values is explained by the publication year. The F-statistic of $0.1520$ and the corresponding \textit{p}-value of $0.700$ also indicate that the regression model is not statistically significant. Therefore, the slight downward trend observed in the SSIM values over time is not statistically supported, and publication year does not significantly affect SSIM values.

Figure~\ref{fig:odd_ratio_v01}D presents a boxplot of PSNR MoM across different publication years. These MoMs are compared across years to assess whether the quality improvement, as measured by PSNR, has remained consistent. The Kruskal-Wallis test yielded a \textit{p}-value of 0.66, suggesting no significant difference in the PSNR MoMs across the years. This indicates that, despite some year-to-year variations, the overall trend in PSNR improvements has remained stable. Figure~\ref{fig:odd_ratio_v01}E shows a similar boxplot for SSIM MoM across the same publication years. The Kruskal-Wallis test resulted in a \textit{p}-value of $0.74$, indicating no significant differences in SSIM MoM over the years. Like PSNR, this suggests that SSIM metric values have not experienced significant fluctuations over time.

Figure~\ref{fig:odd_ratio_v01}F provides a heatmap representing the correlation matrix ($\rho$ values) between various image quality metrics. The matrix illustrates how strongly different metrics are correlated with each other. Strong positive correlations are indicated by darker red shades, while weaker correlations or negative correlations are in lighter shades. For example, PSNR shows a strong correlation with SSIM and other related metrics, while MSE, which is inversely related to PSNR, shows an expected negative correlation. This matrix is critical for understanding the interdependence of different image quality metrics and how they collectively influence the overall assessment of image quality (see Table~\ref{tab:metrics_summary}).

\begin{table*}[tbh!]
	\centering
	\caption{Summary of Image Quality Metrics for Motion Artifact Correction and Detection}
	\label{tab:metrics_summary}
	\resizebox{\textwidth}{!}{%
	\begin{tabular}{p{1cm}cp{14cm}}
		\hline
		\textbf{Metric} & \textbf{Formula} & \textbf{Comment} \\ \hline

		MAE & 
		$ \frac{1}{n}\sum_{i=1}^{n} |x_i - y_i|$ & 
		Measures the average absolute difference between the artifact-corrected image and the ground truth; simple to compute but may not capture perceptual differences introduced by motion artifacts. \\ 
		
		\rowcolor[HTML]{EFEFEF}MSE & 
		$ \frac{1}{n}\sum_{i=1}^{n} (x_i - y_i)^2$ & 
		Calculates the average of squared differences; sensitive to large errors typical in regions affected by motion; useful for quantifying overall correction performance but may exaggerate the impact of outliers. \\ 
		
		PSNR & 
		$10 \log_{10} \left(\frac{\text{MAX}^2}{\text{MSE}}\right)$ & 
		Assesses the peak error between the corrected image and ground truth; higher PSNR indicates better artifact suppression; however, it may not align with perceived visual quality in the presence of motion artifacts. \\ 
		
		\rowcolor[HTML]{EFEFEF}SSIM & 
		$ \frac{(2\mu_x\mu_y + C_1)(2\sigma_{xy} + C_2)}{(\mu_x^2 + \mu_y^2 + C_1)(\sigma_x^2 + \sigma_y^2 + C_2)}$ & 
		Evaluates structural similarity, focusing on luminance, contrast, and structure; more sensitive to motion-induced distortions; better correlates with human perception of image quality after artifact correction. \\ 
		
		MS-SSIM & 
		$ \prod_{j=1}^M \text{SSIM}_j(x, y)^{\alpha_j}$ & 
		Multiscale extension of SSIM; assesses image quality across multiple resolutions; effectively captures the correction of motion artifacts occurring at different spatial scales. \\ 
		
		\rowcolor[HTML]{EFEFEF}Dice Score & 
		$\frac{2|X \cap Y|}{|X| + |Y|}$ & 
		Used for evaluating segmentation accuracy in artifact detection; measures the overlap between detected motion artifact regions and the ground truth; sensitive to small artifacts and class imbalance in motion prevalence. \\ 
		
		\hline
	\end{tabular}
	
}
\end{table*}

\section{Discussion}\label{sec:disscussion_futureDirection}

DL algorithms have markedly transformed the landscape of MRI motion correction and detection, offering substantial improvements in correcting and detecting motion artifacts. We performed a systematic review and meta-analysis to highlight the motion detection and correction techniques, progress,  trends of data usage and DL models, limitations and challenges, future directions of DL-based motion detection (MoDe) and motion correction (MoCo) models.

\subsection{Advancements and impact}

One of the key advancements is the application of generative models like generative adversarial networks (GANs) and denoising diffusion probabilistic models (DDPMs) in this domain. These models have been adapted from their original tasks in image generation to handle effectively the unique challenges posed by MRI motion artifacts. GANs, for example, have been employed both for direct motion correction and as adversarial regularizers to enhance model performance, while DDPMs offer a novel way to reconstruct high-quality images from corrupted data. Additionally, the development of unsupervised and unpaired learning methods has reduced the dependency on large, paired datasets, which are often difficult and expensive to acquire.

The impact of these advancements is substantial, offering the potential for more accurate diagnostics, fewer repeated scans, and overall improvements in patient care. By enhancing image quality and diagnostic accuracy, DL-based MoCo and MoDe models not only improve clinical outcomes but also contribute to cost savings in healthcare settings. The ability to generalize these models across different MRI sequences and patient populations further underscores their robustness and applicability in real-world clinical scenarios. As these technologies continue to evolve, they are poised to become integral tools in the ongoing effort to improve the reliability and accessibility of MRI diagnostics.

\subsection{Challenges and future direction}

Although DL-based MoCo and MoDe techniques have achieved excellent results and outperformed traditional methods~\cite{zhang2024motion}, they require large datasets with raw \textit{k}-space data for training. While large-scale 2D raw \textit{k}-space datasets such as fastMRI~\cite{zbontar2018fastmri} are already available and have greatly advanced the field (e.g., in compressed sensing MRI~\cite{safari2024fast}), there remains a critical need for large 3D MRI datasets. Releasing such 3D datasets could similarly transform the development and benchmarking of DL-based MoCo and MoDe methods. Moreover, zero-shot and transfer learning methods can be explored to detect and remove motion artifacts. Zero-shot learning models can be designed to handle motion artifacts in MRI scans without being explicitly trained on every possible type of motion artifact. As such, zero-shot learning leverages semantic information or prior knowledge about motion artifacts to make accurate corrections and detections even for types of motion artifacts that the model has not encountered during training.

On the other hand, in the transfer learning technique, a model is first trained on a large dataset, such as a natural image dataset or a medical imaging dataset that captures related features. Once pre-trained, the model is fine-tuned on a smaller domain-specific dataset containing MRI scans with motion artifacts. During this fine-tuning phase, the model's parameters are adjusted to specialize in detecting and correcting the motion artifacts present in the MRI data. Transfer learning is particularly effective for MoCo and MoDe tasks because it allows the model to build on the knowledge gained from other tasks, which share underlying principles with motion detection and correction. This approach can markedly reduce the amount of paired data needed for training while improving the model's ability to generalize to new, unseen motion artifacts.

Another issue with DL techniques is their generalization to different MRI sequences and datasets. While we were unable to find studies specifically exploring the generalizability of these models to out-of-distribution datasets, it is hypothesized that their performance may decline when applied to such datasets. We further speculate that the use of vision-language models~\cite{zhang2024vision} could enhance generalization, as text prompts might provide additional context that can guide the motion detection and correction processes, potentially making these models more robust across varied datasets and MRI sequences.

A further challenge arises from the heterogeneity of reported architectures. Many studies employ hybrid models that combine elements such as convolutional backbones, Transformer blocks~\cite{10647626}, GAN frameworks, or diffusion modules. This blending of design paradigms reflects the current direction of the field, but it makes it difficult to categorize studies unambiguously or to disentangle the contribution of each component. As a result, comparative analysis across architectures is limited, underscoring the need for standardized benchmarks and reporting practices in future work.

Our systematic review and meta-analysis focused on how much DL models improved image quality in MoCo tasks and their performances in MoDe tasks. However, such a comparison is challenging as the majority of studies used motion-simulated \textit{in-silico} datasets, which may not be consistent in different studies due to the differences in the locations of motion-corrupted lines. Lines close to the \textit{k}-space center contribute to the blurring and ghosting, while the peripheral lines contribute to ringing artifacts~\cite{shaw2020k}. Standardized and reproducible methods are required to generate motion artifacts in different studies. One quick remedy could be the consistent reporting of quantitative metrics that can specify ghosting and blurring contents in motion-corrupted MR images such as PSNR and MS-SSIM metrics.

\section{Conclusion}

In this systematic review and meta-analysis, we explored the landscape of DL techniques for MoCo and MoDe in MRI. The advancements in DL have shown significant potential in addressing the challenges posed by motion artifacts in MRI. These models have demonstrated their potential to enhance image quality, reduce the need for repeated scans, and improve diagnostic accuracy, thereby contributing to better clinical outcomes and cost savings in healthcare.

However, the generalizability of these models across different MRI sequences and out-of-distribution datasets remains a concern. While the current research shows great promise, there is a clear need for further studies to explore the robustness of these models in varied clinical settings. Additionally, the integration of emerging techniques such as zero-shot learning and transfer learning could further enhance the ability of these models to generalize and perform effectively in diverse scenarios.

Overall, the continued evolution of DL-based MoCo and MoDe models is poised to play a crucial role in the future of MRI diagnostics, offering the potential to improve significantly the reliability and accessibility of high-quality medical imaging. As these technologies mature, their impact on patient care and healthcare systems will likely become even more profound.

\section*{Acknowledgments}
This research is supported in part by the National Institutes of Health under Award Numbers R56EB033332, R01EB032680, and R01CA272991.

\bibliographystyle{elsarticle-num-names} 
\bibliography{MoCo_review.bib}

\begin{thebibliography}{175}
\expandafter\ifx\csname natexlab\endcsname\relax\def\natexlab#1{#1}\fi
\providecommand{\url}[1]{\texttt{#1}}
\providecommand{\href}[2]{#2}
\providecommand{\path}[1]{#1}
\providecommand{\DOIprefix}{doi:}
\providecommand{\ArXivprefix}{arXiv:}
\providecommand{\URLprefix}{URL: }
\providecommand{\Pubmedprefix}{pmid:}
\providecommand{\doi}[1]{\href{http://dx.doi.org/#1}{\path{#1}}}
\providecommand{\Pubmed}[1]{\href{pmid:#1}{\path{#1}}}
\providecommand{\bibinfo}[2]{#2}
\ifx\xfnm\relax \def\xfnm[#1]{\unskip,\space#1}\fi
\bibitem[{Motyka et~al.(2024)Motyka, Weiser, Bachrata, Hingerl, Strasser,
  Hangel, Niess, Niess, Zaitsev, Robinson et~al.}]{motyka2024predicting}
\bibinfo{author}{S.~Motyka}, \bibinfo{author}{P.~Weiser},
  \bibinfo{author}{B.~Bachrata}, \bibinfo{author}{L.~Hingerl},
  \bibinfo{author}{B.~Strasser}, \bibinfo{author}{G.~Hangel},
  \bibinfo{author}{E.~Niess}, \bibinfo{author}{F.~Niess},
  \bibinfo{author}{M.~Zaitsev}, \bibinfo{author}{S.~D. Robinson}, et~al.,
\newblock \bibinfo{title}{Predicting dynamic, motion-related changes in b 0
  field in the brain at a 7t mri using a subject-specific fine-trained u-net},
\newblock \bibinfo{journal}{Magnetic resonance in medicine}
  \bibinfo{volume}{91} (\bibinfo{year}{2024}) \bibinfo{pages}{2044--2056}.
\bibitem[{Safari et~al.(2023)Safari, Fatemi, Afkham, and
  Archambault}]{safari2023patient}
\bibinfo{author}{M.~Safari}, \bibinfo{author}{A.~Fatemi},
  \bibinfo{author}{Y.~Afkham}, \bibinfo{author}{L.~Archambault},
\newblock \bibinfo{title}{Patient-specific geometrical distortion corrections
  of mri images improve dosimetric planning accuracy of vestibular schwannoma
  treated with gamma knife stereotactic radiosurgery},
\newblock \bibinfo{journal}{Journal of Applied Clinical Medical Physics}
  \bibinfo{volume}{24} (\bibinfo{year}{2023}) \bibinfo{pages}{e14072}.
\bibitem[{Wood and Henkelman(1985)}]{wood1985mr}
\bibinfo{author}{M.~L. Wood}, \bibinfo{author}{R.~M. Henkelman},
\newblock \bibinfo{title}{Mr image artifacts from periodic motion},
\newblock \bibinfo{journal}{Medical physics} \bibinfo{volume}{12}
  (\bibinfo{year}{1985}) \bibinfo{pages}{143--151}.
\bibitem[{Zaitsev et~al.(2015)Zaitsev, Maclaren, and
  Herbst}]{zaitsev2015motion}
\bibinfo{author}{M.~Zaitsev}, \bibinfo{author}{J.~Maclaren},
  \bibinfo{author}{M.~Herbst},
\newblock \bibinfo{title}{Motion artifacts in mri: A complex problem with many
  partial solutions},
\newblock \bibinfo{journal}{Journal of Magnetic Resonance Imaging}
  \bibinfo{volume}{42} (\bibinfo{year}{2015}) \bibinfo{pages}{887--901}.
\bibitem[{Sreekumari et~al.(2019)Sreekumari, Shanbhag, Yeo, Foo, Pilitsis,
  Polzin, Patil, Coblentz, Kapadia, Khinda et~al.}]{sreekumari2019deep}
\bibinfo{author}{A.~Sreekumari}, \bibinfo{author}{D.~Shanbhag},
  \bibinfo{author}{D.~Yeo}, \bibinfo{author}{T.~Foo},
  \bibinfo{author}{J.~Pilitsis}, \bibinfo{author}{J.~Polzin},
  \bibinfo{author}{U.~Patil}, \bibinfo{author}{A.~Coblentz},
  \bibinfo{author}{A.~Kapadia}, \bibinfo{author}{J.~Khinda}, et~al.,
\newblock \bibinfo{title}{A deep learning--based approach to reduce rescan and
  recall rates in clinical mri examinations},
\newblock \bibinfo{journal}{American Journal of Neuroradiology}
  \bibinfo{volume}{40} (\bibinfo{year}{2019}) \bibinfo{pages}{217--223}.
\bibitem[{Sui et~al.(2024)Sui, Palaniappan, Brenner, Paganelli, Kurz, Landry,
  and Riboldi}]{sui2024intra}
\bibinfo{author}{Z.~Sui}, \bibinfo{author}{P.~Palaniappan},
  \bibinfo{author}{J.~Brenner}, \bibinfo{author}{C.~Paganelli},
  \bibinfo{author}{C.~Kurz}, \bibinfo{author}{G.~Landry},
  \bibinfo{author}{M.~Riboldi},
\newblock \bibinfo{title}{Intra-frame motion deterioration effects and
  deep-learning-based compensation in mr-guided radiotherapy},
\newblock \bibinfo{journal}{Medical Physics} \bibinfo{volume}{51}
  (\bibinfo{year}{2024}) \bibinfo{pages}{1899--1917}.
\bibitem[{WangNicholas et~al.(2022)WangNicholas, NollDouglas, KimMichelle
  et~al.}]{wangnicholas2022simulated}
\bibinfo{author}{C.~WangNicholas}, \bibinfo{author}{C.~NollDouglas},
  \bibinfo{author}{M.~KimMichelle}, et~al.,
\newblock \bibinfo{title}{Simulated mri artifacts: Testing machine learning
  failure modes},
\newblock \bibinfo{journal}{BME frontiers}  (\bibinfo{year}{2022}).
\bibitem[{Kemenczky et~al.(2022)Kemenczky, Vakli, Somogyi, Homolya, Hermann,
  G{\'a}l, and Vidny{\'a}nszky}]{kemenczky2022effect}
\bibinfo{author}{P.~Kemenczky}, \bibinfo{author}{P.~Vakli},
  \bibinfo{author}{E.~Somogyi}, \bibinfo{author}{I.~Homolya},
  \bibinfo{author}{P.~Hermann}, \bibinfo{author}{V.~G{\'a}l},
  \bibinfo{author}{Z.~Vidny{\'a}nszky},
\newblock \bibinfo{title}{Effect of head motion-induced artefacts on the
  reliability of deep learning-based whole-brain segmentation},
\newblock \bibinfo{journal}{Scientific reports} \bibinfo{volume}{12}
  (\bibinfo{year}{2022}) \bibinfo{pages}{1618}.
\bibitem[{Hanson et~al.(2024)Hanson, Adkins, Bacas, and
  Zhou}]{hanson2024examining}
\bibinfo{author}{J.~L. Hanson}, \bibinfo{author}{D.~J. Adkins},
  \bibinfo{author}{E.~Bacas}, \bibinfo{author}{P.~Zhou},
\newblock \bibinfo{title}{Examining the reliability of brain age algorithms
  under varying degrees of participant motion},
\newblock \bibinfo{journal}{Brain Informatics} \bibinfo{volume}{11}
  (\bibinfo{year}{2024}) \bibinfo{pages}{9}.
\bibitem[{Beljaards et~al.(2024)Beljaards, Pezzotti, Rao, Doneva, van Osch, and
  Staring}]{beljaards2024ai}
\bibinfo{author}{L.~Beljaards}, \bibinfo{author}{N.~Pezzotti},
  \bibinfo{author}{C.~Rao}, \bibinfo{author}{M.~Doneva}, \bibinfo{author}{M.~J.
  van Osch}, \bibinfo{author}{M.~Staring},
\newblock \bibinfo{title}{Ai-based motion artifact severity estimation in
  undersampled mri allowing for selection of appropriate reconstruction
  models},
\newblock \bibinfo{journal}{Medical Physics} \bibinfo{volume}{51}
  (\bibinfo{year}{2024}) \bibinfo{pages}{3555--3565}.
\bibitem[{Slipsager et~al.(2020)Slipsager, Glimberg, S{\o}gaard, Paulsen,
  Johannesen, Martens, Seth, Marner, Henriksen, Olesen
  et~al.}]{slipsager2020quantifying}
\bibinfo{author}{J.~M. Slipsager}, \bibinfo{author}{S.~L. Glimberg},
  \bibinfo{author}{J.~S{\o}gaard}, \bibinfo{author}{R.~R. Paulsen},
  \bibinfo{author}{H.~H. Johannesen}, \bibinfo{author}{P.~C. Martens},
  \bibinfo{author}{A.~Seth}, \bibinfo{author}{L.~Marner},
  \bibinfo{author}{O.~M. Henriksen}, \bibinfo{author}{O.~V. Olesen}, et~al.,
\newblock \bibinfo{title}{Quantifying the financial savings of motion
  correction in brain mri: a model-based estimate of the costs arising from
  patient head motion and potential savings from implementation of motion
  correction},
\newblock \bibinfo{journal}{Journal of Magnetic Resonance Imaging}
  \bibinfo{volume}{52} (\bibinfo{year}{2020}) \bibinfo{pages}{731--738}.
\bibitem[{Andre et~al.(2015)Andre, Bresnahan, Mossa-Basha, Hoff, Smith, Anzai,
  and Cohen}]{andre2015toward}
\bibinfo{author}{J.~B. Andre}, \bibinfo{author}{B.~W. Bresnahan},
  \bibinfo{author}{M.~Mossa-Basha}, \bibinfo{author}{M.~N. Hoff},
  \bibinfo{author}{C.~P. Smith}, \bibinfo{author}{Y.~Anzai},
  \bibinfo{author}{W.~A. Cohen},
\newblock \bibinfo{title}{Toward quantifying the prevalence, severity, and cost
  associated with patient motion during clinical mr examinations},
\newblock \bibinfo{journal}{Journal of the American College of Radiology}
  \bibinfo{volume}{12} (\bibinfo{year}{2015}) \bibinfo{pages}{689--695}.
\bibitem[{Stucht et~al.(2015)Stucht, Danishad, Schulze, Godenschweger, Zaitsev,
  and Speck}]{stucht2015highest}
\bibinfo{author}{D.~Stucht}, \bibinfo{author}{K.~A. Danishad},
  \bibinfo{author}{P.~Schulze}, \bibinfo{author}{F.~Godenschweger},
  \bibinfo{author}{M.~Zaitsev}, \bibinfo{author}{O.~Speck},
\newblock \bibinfo{title}{Highest resolution in vivo human brain mri using
  prospective motion correction},
\newblock \bibinfo{journal}{PloS one} \bibinfo{volume}{10}
  (\bibinfo{year}{2015}) \bibinfo{pages}{e0133921}.
\bibitem[{Scott et~al.(2009)Scott, Keegan, and Firmin}]{scott2009motion}
\bibinfo{author}{A.~D. Scott}, \bibinfo{author}{J.~Keegan},
  \bibinfo{author}{D.~N. Firmin},
\newblock \bibinfo{title}{Motion in cardiovascular mr imaging},
\newblock \bibinfo{journal}{Radiology} \bibinfo{volume}{250}
  (\bibinfo{year}{2009}) \bibinfo{pages}{331--351}.
\bibitem[{Haeberlin et~al.(2015)Haeberlin, Kasper, Barmet, Brunner, Dietrich,
  Gross, Wilm, Kozerke, and Pruessmann}]{haeberlin2015real}
\bibinfo{author}{M.~Haeberlin}, \bibinfo{author}{L.~Kasper},
  \bibinfo{author}{C.~Barmet}, \bibinfo{author}{D.~O. Brunner},
  \bibinfo{author}{B.~E. Dietrich}, \bibinfo{author}{S.~Gross},
  \bibinfo{author}{B.~J. Wilm}, \bibinfo{author}{S.~Kozerke},
  \bibinfo{author}{K.~P. Pruessmann},
\newblock \bibinfo{title}{Real-time motion correction using gradient tones and
  head-mounted nmr field probes},
\newblock \bibinfo{journal}{Magnetic resonance in medicine}
  \bibinfo{volume}{74} (\bibinfo{year}{2015}) \bibinfo{pages}{647--660}.
\bibitem[{Aranovitch et~al.(2018)Aranovitch, Haeberlin, Gross, Dietrich, Wilm,
  Brunner, Schmid, Luechinger, and Pruessmann}]{aranovitch2018prospective}
\bibinfo{author}{A.~Aranovitch}, \bibinfo{author}{M.~Haeberlin},
  \bibinfo{author}{S.~Gross}, \bibinfo{author}{B.~E. Dietrich},
  \bibinfo{author}{B.~J. Wilm}, \bibinfo{author}{D.~O. Brunner},
  \bibinfo{author}{T.~Schmid}, \bibinfo{author}{R.~Luechinger},
  \bibinfo{author}{K.~P. Pruessmann},
\newblock \bibinfo{title}{Prospective motion correction with nmr markers using
  only native sequence elements},
\newblock \bibinfo{journal}{Magnetic resonance in medicine}
  \bibinfo{volume}{79} (\bibinfo{year}{2018}) \bibinfo{pages}{2046--2056}.
\bibitem[{Vionnet et~al.(2021)Vionnet, Aranovitch, Duerst, Haeberlin, Dietrich,
  Gross, and Pruessmann}]{vionnet2021simultaneous}
\bibinfo{author}{L.~Vionnet}, \bibinfo{author}{A.~Aranovitch},
  \bibinfo{author}{Y.~Duerst}, \bibinfo{author}{M.~Haeberlin},
  \bibinfo{author}{B.~E. Dietrich}, \bibinfo{author}{S.~Gross},
  \bibinfo{author}{K.~P. Pruessmann},
\newblock \bibinfo{title}{Simultaneous feedback control for joint field and
  motion correction in brain mri},
\newblock \bibinfo{journal}{Neuroimage} \bibinfo{volume}{226}
  (\bibinfo{year}{2021}) \bibinfo{pages}{117286}.
\bibitem[{White et~al.(2010)White, Roddey, Shankaranarayanan, Han, Rettmann,
  Santos, Kuperman, and Dale}]{white2010promo}
\bibinfo{author}{N.~White}, \bibinfo{author}{C.~Roddey},
  \bibinfo{author}{A.~Shankaranarayanan}, \bibinfo{author}{E.~Han},
  \bibinfo{author}{D.~Rettmann}, \bibinfo{author}{J.~Santos},
  \bibinfo{author}{J.~Kuperman}, \bibinfo{author}{A.~Dale},
\newblock \bibinfo{title}{Promo: real-time prospective motion correction in mri
  using image-based tracking},
\newblock \bibinfo{journal}{Magnetic Resonance in Medicine: An Official Journal
  of the International Society for Magnetic Resonance in Medicine}
  \bibinfo{volume}{63} (\bibinfo{year}{2010}) \bibinfo{pages}{91--105}.
\bibitem[{Tisdall et~al.(2012)Tisdall, Hess, Reuter, Meintjes, Fischl, and
  van~der Kouwe}]{tisdall2012volumetric}
\bibinfo{author}{M.~D. Tisdall}, \bibinfo{author}{A.~T. Hess},
  \bibinfo{author}{M.~Reuter}, \bibinfo{author}{E.~M. Meintjes},
  \bibinfo{author}{B.~Fischl}, \bibinfo{author}{A.~J. van~der Kouwe},
\newblock \bibinfo{title}{Volumetric navigators for prospective motion
  correction and selective reacquisition in neuroanatomical mri},
\newblock \bibinfo{journal}{Magnetic resonance in medicine}
  \bibinfo{volume}{68} (\bibinfo{year}{2012}) \bibinfo{pages}{389--399}.
\bibitem[{Zhu et~al.(2020)Zhu, Chan, Lustig, Johnson, and
  Larson}]{zhu2020iterative}
\bibinfo{author}{X.~Zhu}, \bibinfo{author}{M.~Chan},
  \bibinfo{author}{M.~Lustig}, \bibinfo{author}{K.~M. Johnson},
  \bibinfo{author}{P.~E. Larson},
\newblock \bibinfo{title}{Iterative motion-compensation reconstruction
  ultra-short te (imoco ute) for high-resolution free-breathing pulmonary mri},
\newblock \bibinfo{journal}{Magnetic resonance in medicine}
  \bibinfo{volume}{83} (\bibinfo{year}{2020}) \bibinfo{pages}{1208--1221}.
\bibitem[{Gumus et~al.(2015)Gumus, Keating, White, Andrews-Shigaki, Armstrong,
  Maclaren, Zaitsev, Dale, and Ernst}]{gumus2015comparison}
\bibinfo{author}{K.~Gumus}, \bibinfo{author}{B.~Keating},
  \bibinfo{author}{N.~White}, \bibinfo{author}{B.~Andrews-Shigaki},
  \bibinfo{author}{B.~Armstrong}, \bibinfo{author}{J.~Maclaren},
  \bibinfo{author}{M.~Zaitsev}, \bibinfo{author}{A.~Dale},
  \bibinfo{author}{T.~Ernst},
\newblock \bibinfo{title}{Comparison of optical and mr-based tracking},
\newblock \bibinfo{journal}{Magnetic resonance in medicine}
  \bibinfo{volume}{74} (\bibinfo{year}{2015}) \bibinfo{pages}{894--902}.
\bibitem[{van Niekerk et~al.(2019)van Niekerk, van~der Kouwe, and
  Meintjes}]{van2019toward}
\bibinfo{author}{A.~van Niekerk}, \bibinfo{author}{A.~van~der Kouwe},
  \bibinfo{author}{E.~Meintjes},
\newblock \bibinfo{title}{Toward “plug and play” prospective motion
  correction for mri by combining observations of the time varying gradient and
  static vector fields},
\newblock \bibinfo{journal}{Magnetic resonance in medicine}
  \bibinfo{volume}{82} (\bibinfo{year}{2019}) \bibinfo{pages}{1214--1228}.
\bibitem[{Vacul{\v{c}}iakov{\'a} et~al.(2022)Vacul{\v{c}}iakov{\'a}, Podranski,
  Edwards, Ocal, Veale, Fox, Haak, Ehses, Callaghan, Pine
  et~al.}]{vaculvciakova2022combining}
\bibinfo{author}{L.~Vacul{\v{c}}iakov{\'a}}, \bibinfo{author}{K.~Podranski},
  \bibinfo{author}{L.~J. Edwards}, \bibinfo{author}{D.~Ocal},
  \bibinfo{author}{T.~Veale}, \bibinfo{author}{N.~C. Fox},
  \bibinfo{author}{R.~Haak}, \bibinfo{author}{P.~Ehses}, \bibinfo{author}{M.~F.
  Callaghan}, \bibinfo{author}{K.~J. Pine}, et~al.,
\newblock \bibinfo{title}{Combining navigator and optical prospective motion
  correction for high-quality 500 $\mu$m resolution quantitative
  multi-parameter mapping at 7t},
\newblock \bibinfo{journal}{Magnetic Resonance in Medicine}
  \bibinfo{volume}{88} (\bibinfo{year}{2022}) \bibinfo{pages}{787--801}.
\bibitem[{Slipsager et~al.(2022)Slipsager, Glimberg, H{\o}jgaard, Paulsen,
  Wighton, Tisdall, Jaimes, Gagoski, Grant, van Der~Kouwe
  et~al.}]{slipsager2022comparison}
\bibinfo{author}{J.~M. Slipsager}, \bibinfo{author}{S.~L. Glimberg},
  \bibinfo{author}{L.~H{\o}jgaard}, \bibinfo{author}{R.~R. Paulsen},
  \bibinfo{author}{P.~Wighton}, \bibinfo{author}{M.~D. Tisdall},
  \bibinfo{author}{C.~Jaimes}, \bibinfo{author}{B.~A. Gagoski},
  \bibinfo{author}{P.~E. Grant}, \bibinfo{author}{A.~van Der~Kouwe}, et~al.,
\newblock \bibinfo{title}{Comparison of prospective and retrospective motion
  correction in 3d-encoded neuroanatomical mri},
\newblock \bibinfo{journal}{Magnetic resonance in medicine}
  \bibinfo{volume}{87} (\bibinfo{year}{2022}) \bibinfo{pages}{629--645}.
\bibitem[{Godenschweger et~al.(2016)Godenschweger, K{\"a}gebein, Stucht,
  Yarach, Sciarra, Yakupov, L{\"u}sebrink, Schulze, and
  Speck}]{godenschweger2016motion}
\bibinfo{author}{F.~Godenschweger}, \bibinfo{author}{U.~K{\"a}gebein},
  \bibinfo{author}{D.~Stucht}, \bibinfo{author}{U.~Yarach},
  \bibinfo{author}{A.~Sciarra}, \bibinfo{author}{R.~Yakupov},
  \bibinfo{author}{F.~L{\"u}sebrink}, \bibinfo{author}{P.~Schulze},
  \bibinfo{author}{O.~Speck},
\newblock \bibinfo{title}{Motion correction in mri of the brain},
\newblock \bibinfo{journal}{Physics in medicine \& biology}
  \bibinfo{volume}{61} (\bibinfo{year}{2016}) \bibinfo{pages}{R32}.
\bibitem[{Singh et~al.(2020)Singh, Salehi, and Gholipour}]{9103624}
\bibinfo{author}{A.~Singh}, \bibinfo{author}{S.~S.~M. Salehi},
  \bibinfo{author}{A.~Gholipour},
\newblock \bibinfo{title}{Deep predictive motion tracking in magnetic resonance
  imaging: Application to fetal imaging},
\newblock \bibinfo{journal}{IEEE Transactions on Medical Imaging}
  \bibinfo{volume}{39} (\bibinfo{year}{2020}) \bibinfo{pages}{3523--3534}.
  \DOIprefix\doi{10.1109/TMI.2020.2998600}.
\bibitem[{Shao et~al.(2022)Shao, Li, Dohopolski, Wang, Cai, Tan, Wang, and
  Zhang}]{shao2022real}
\bibinfo{author}{H.-C. Shao}, \bibinfo{author}{T.~Li}, \bibinfo{author}{M.~J.
  Dohopolski}, \bibinfo{author}{J.~Wang}, \bibinfo{author}{J.~Cai},
  \bibinfo{author}{J.~Tan}, \bibinfo{author}{K.~Wang},
  \bibinfo{author}{Y.~Zhang},
\newblock \bibinfo{title}{Real-time mri motion estimation through an
  unsupervised k-space-driven deformable registration network (ks-regnet)},
\newblock \bibinfo{journal}{Physics in Medicine \& Biology}
  \bibinfo{volume}{67} (\bibinfo{year}{2022}) \bibinfo{pages}{135012}.
\bibitem[{Neves~Silva et~al.(2023)Neves~Silva, Aviles~Verdera, Tomi-Tricot,
  Neji, Uus, Grigorescu, Wilkinson, Ozenne, Lewin, Story
  et~al.}]{neves2023real}
\bibinfo{author}{S.~Neves~Silva}, \bibinfo{author}{J.~Aviles~Verdera},
  \bibinfo{author}{R.~Tomi-Tricot}, \bibinfo{author}{R.~Neji},
  \bibinfo{author}{A.~Uus}, \bibinfo{author}{I.~Grigorescu},
  \bibinfo{author}{T.~Wilkinson}, \bibinfo{author}{V.~Ozenne},
  \bibinfo{author}{A.~Lewin}, \bibinfo{author}{L.~Story}, et~al.,
\newblock \bibinfo{title}{Real-time fetal brain tracking for functional fetal
  mri},
\newblock \bibinfo{journal}{Magnetic resonance in medicine}
  \bibinfo{volume}{90} (\bibinfo{year}{2023}) \bibinfo{pages}{2306--2320}.
\bibitem[{Duffy et~al.(2018)Duffy, Zhang, Tang, Zhao, Law, Toga, and
  Kim}]{duffy2018retrospective}
\bibinfo{author}{B.~A. Duffy}, \bibinfo{author}{W.~Zhang},
  \bibinfo{author}{H.~Tang}, \bibinfo{author}{L.~Zhao},
  \bibinfo{author}{M.~Law}, \bibinfo{author}{A.~W. Toga},
  \bibinfo{author}{H.~Kim},
\newblock \bibinfo{title}{Retrospective correction of motion artifact affected
  structural mri images using deep learning of simulated motion},
\newblock in: \bibinfo{booktitle}{Medical imaging with deep learning},
  \bibinfo{year}{2018}.
\bibitem[{K{\"u}stner et~al.(2019)K{\"u}stner, Armanious, Yang, Yang, Schick,
  and Gatidis}]{kustner2019retrospective}
\bibinfo{author}{T.~K{\"u}stner}, \bibinfo{author}{K.~Armanious},
  \bibinfo{author}{J.~Yang}, \bibinfo{author}{B.~Yang},
  \bibinfo{author}{F.~Schick}, \bibinfo{author}{S.~Gatidis},
\newblock \bibinfo{title}{Retrospective correction of motion-affected mr images
  using deep learning frameworks},
\newblock \bibinfo{journal}{Magnetic resonance in medicine}
  \bibinfo{volume}{82} (\bibinfo{year}{2019}) \bibinfo{pages}{1527--1540}.
\bibitem[{Lee et~al.(2020)Lee, Jung, Jung, and Kim}]{lee2020deep}
\bibinfo{author}{S.~Lee}, \bibinfo{author}{S.~Jung}, \bibinfo{author}{K.-J.
  Jung}, \bibinfo{author}{D.-H. Kim},
\newblock \bibinfo{title}{Deep learning in mr motion correction: a brief review
  and a new motion simulation tool (view2dmotion)},
\newblock \bibinfo{journal}{Investigative Magnetic Resonance Imaging}
  \bibinfo{volume}{24} (\bibinfo{year}{2020}) \bibinfo{pages}{196--206}.
\bibitem[{Chang et~al.(2023)Chang, Li, Saju, Mao, and Liu}]{chang2023deep}
\bibinfo{author}{Y.~Chang}, \bibinfo{author}{Z.~Li}, \bibinfo{author}{G.~Saju},
  \bibinfo{author}{H.~Mao}, \bibinfo{author}{T.~Liu},
\newblock \bibinfo{title}{Deep learning-based rigid motion correction for
  magnetic resonance imaging: a survey},
\newblock \bibinfo{journal}{Meta-Radiology} \bibinfo{volume}{1}
  (\bibinfo{year}{2023}) \bibinfo{pages}{100001}.
\bibitem[{Spieker et~al.(2023)Spieker, Eichhorn, Hammernik, Rueckert,
  Preibisch, Karampinos, and Schnabel}]{spieker2023deep}
\bibinfo{author}{V.~Spieker}, \bibinfo{author}{H.~Eichhorn},
  \bibinfo{author}{K.~Hammernik}, \bibinfo{author}{D.~Rueckert},
  \bibinfo{author}{C.~Preibisch}, \bibinfo{author}{D.~C. Karampinos},
  \bibinfo{author}{J.~A. Schnabel},
\newblock \bibinfo{title}{Deep learning for retrospective motion correction in
  mri: a comprehensive review},
\newblock \bibinfo{journal}{IEEE Transactions on Medical Imaging}
  \bibinfo{volume}{43} (\bibinfo{year}{2023}) \bibinfo{pages}{846--859}.
\bibitem[{Zhang et~al.(2024)Zhang, Wang, Rawson, Balan, Herskovits, Melhem,
  Chang, Wang, and Ernst}]{zhang2024motion}
\bibinfo{author}{L.~Zhang}, \bibinfo{author}{X.~Wang},
  \bibinfo{author}{M.~Rawson}, \bibinfo{author}{R.~Balan},
  \bibinfo{author}{E.~H. Herskovits}, \bibinfo{author}{E.~R. Melhem},
  \bibinfo{author}{L.~Chang}, \bibinfo{author}{Z.~Wang},
  \bibinfo{author}{T.~Ernst},
\newblock \bibinfo{title}{Motion correction for brain mri using deep learning
  and a novel hybrid loss function},
\newblock \bibinfo{journal}{Algorithms} \bibinfo{volume}{17}
  (\bibinfo{year}{2024}) \bibinfo{pages}{215}.
\bibitem[{LeCun et~al.(2015)LeCun, Bengio, and Hinton}]{lecun2015deep}
\bibinfo{author}{Y.~LeCun}, \bibinfo{author}{Y.~Bengio},
  \bibinfo{author}{G.~Hinton},
\newblock \bibinfo{title}{Deep learning},
\newblock \bibinfo{journal}{nature} \bibinfo{volume}{521}
  (\bibinfo{year}{2015}) \bibinfo{pages}{436--444}.
\bibitem[{Ronneberger et~al.(2015)Ronneberger, Fischer, and
  Brox}]{ronneberger2015u}
\bibinfo{author}{O.~Ronneberger}, \bibinfo{author}{P.~Fischer},
  \bibinfo{author}{T.~Brox},
\newblock \bibinfo{title}{U-net: Convolutional networks for biomedical image
  segmentation},
\newblock in: \bibinfo{booktitle}{International Conference on Medical image
  computing and computer-assisted intervention},
  \bibinfo{organization}{Springer}, \bibinfo{year}{2015}, pp.
  \bibinfo{pages}{234--241}.
\bibitem[{Milletari et~al.(2016)Milletari, Navab, and Ahmadi}]{7785132}
\bibinfo{author}{F.~Milletari}, \bibinfo{author}{N.~Navab},
  \bibinfo{author}{S.-A. Ahmadi},
\newblock \bibinfo{title}{V-net: Fully convolutional neural networks for
  volumetric medical image segmentation},
\newblock in: \bibinfo{booktitle}{2016 Fourth International Conference on 3D
  Vision (3DV)}, \bibinfo{year}{2016}, pp. \bibinfo{pages}{565--571}.
  \DOIprefix\doi{10.1109/3DV.2016.79}.
\bibitem[{Safari et~al.(2024)Safari, Yang, and Fatemi}]{safari2024information}
\bibinfo{author}{M.~Safari}, \bibinfo{author}{X.~Yang},
  \bibinfo{author}{A.~Fatemi},
\newblock \bibinfo{title}{Information maximized u-nets for vestibular
  schwannoma segmentation using mri with missing modality},
\newblock in: \bibinfo{booktitle}{Medical Imaging 2024: Clinical and Biomedical
  Imaging}, volume \bibinfo{volume}{12930}, \bibinfo{organization}{SPIE},
  \bibinfo{year}{2024}, pp. \bibinfo{pages}{370--374}.
\bibitem[{Balakrishnan et~al.(2019)Balakrishnan, Zhao, Sabuncu, Guttag, and
  Dalca}]{8633930}
\bibinfo{author}{G.~Balakrishnan}, \bibinfo{author}{A.~Zhao},
  \bibinfo{author}{M.~R. Sabuncu}, \bibinfo{author}{J.~Guttag},
  \bibinfo{author}{A.~V. Dalca},
\newblock \bibinfo{title}{Voxelmorph: A learning framework for deformable
  medical image registration},
\newblock \bibinfo{journal}{IEEE Transactions on Medical Imaging}
  \bibinfo{volume}{38} (\bibinfo{year}{2019}) \bibinfo{pages}{1788--1800}.
  \DOIprefix\doi{10.1109/TMI.2019.2897538}.
\bibitem[{You et~al.(2020)You, Li, Zhang, Zhang, Shan, Li, Ju, Zhao, Zhang,
  Cong, Vannier, Saha, Hoffman, and Wang}]{8736838}
\bibinfo{author}{C.~You}, \bibinfo{author}{G.~Li}, \bibinfo{author}{Y.~Zhang},
  \bibinfo{author}{X.~Zhang}, \bibinfo{author}{H.~Shan},
  \bibinfo{author}{M.~Li}, \bibinfo{author}{S.~Ju}, \bibinfo{author}{Z.~Zhao},
  \bibinfo{author}{Z.~Zhang}, \bibinfo{author}{W.~Cong}, \bibinfo{author}{M.~W.
  Vannier}, \bibinfo{author}{P.~K. Saha}, \bibinfo{author}{E.~A. Hoffman},
  \bibinfo{author}{G.~Wang},
\newblock \bibinfo{title}{Ct super-resolution gan constrained by the identical,
  residual, and cycle learning ensemble (gan-circle)},
\newblock \bibinfo{journal}{IEEE Transactions on Medical Imaging}
  \bibinfo{volume}{39} (\bibinfo{year}{2020}) \bibinfo{pages}{188--203}.
  \DOIprefix\doi{10.1109/TMI.2019.2922960}.
\bibitem[{Chen et~al.(2017)Chen, Zhang, Kalra, Lin, Chen, Liao, Zhou, and
  Wang}]{7947200}
\bibinfo{author}{H.~Chen}, \bibinfo{author}{Y.~Zhang}, \bibinfo{author}{M.~K.
  Kalra}, \bibinfo{author}{F.~Lin}, \bibinfo{author}{Y.~Chen},
  \bibinfo{author}{P.~Liao}, \bibinfo{author}{J.~Zhou},
  \bibinfo{author}{G.~Wang},
\newblock \bibinfo{title}{Low-dose ct with a residual encoder-decoder
  convolutional neural network},
\newblock \bibinfo{journal}{IEEE Transactions on Medical Imaging}
  \bibinfo{volume}{36} (\bibinfo{year}{2017}) \bibinfo{pages}{2524--2535}.
  \DOIprefix\doi{10.1109/TMI.2017.2715284}.
\bibitem[{Pham et~al.(2019)Pham, Tor-D{\'\i}ez, Meunier, Bednarek, Fablet,
  Passat, and Rousseau}]{pham2019multiscale}
\bibinfo{author}{C.-H. Pham}, \bibinfo{author}{C.~Tor-D{\'\i}ez},
  \bibinfo{author}{H.~Meunier}, \bibinfo{author}{N.~Bednarek},
  \bibinfo{author}{R.~Fablet}, \bibinfo{author}{N.~Passat},
  \bibinfo{author}{F.~Rousseau},
\newblock \bibinfo{title}{Multiscale brain mri super-resolution using deep 3d
  convolutional networks},
\newblock \bibinfo{journal}{Computerized Medical Imaging and Graphics}
  \bibinfo{volume}{77} (\bibinfo{year}{2019}) \bibinfo{pages}{101647}.
\bibitem[{Rudie et~al.(2022)Rudie, Gleason, Barkovich, Wilson,
  Shankaranarayanan, Zhang, Wang, Gong, Zaharchuk, and
  Villanueva-Meyer}]{rudie2022clinical}
\bibinfo{author}{J.~D. Rudie}, \bibinfo{author}{T.~Gleason},
  \bibinfo{author}{M.~J. Barkovich}, \bibinfo{author}{D.~M. Wilson},
  \bibinfo{author}{A.~Shankaranarayanan}, \bibinfo{author}{T.~Zhang},
  \bibinfo{author}{L.~Wang}, \bibinfo{author}{E.~Gong},
  \bibinfo{author}{G.~Zaharchuk}, \bibinfo{author}{J.~E. Villanueva-Meyer},
\newblock \bibinfo{title}{Clinical assessment of deep learning--based
  super-resolution for 3d volumetric brain mri},
\newblock \bibinfo{journal}{Radiology: Artificial Intelligence}
  \bibinfo{volume}{4} (\bibinfo{year}{2022}) \bibinfo{pages}{e210059}.
\bibitem[{Safari et~al.(2025)Safari, Wang, Eidex, Li, Qiu, Middlebrooks, Yu,
  and Yang}]{Safari_2025}
\bibinfo{author}{M.~Safari}, \bibinfo{author}{S.~Wang},
  \bibinfo{author}{Z.~Eidex}, \bibinfo{author}{Q.~Li},
  \bibinfo{author}{R.~L.~J. Qiu}, \bibinfo{author}{E.~H. Middlebrooks},
  \bibinfo{author}{D.~S. Yu}, \bibinfo{author}{X.~Yang},
\newblock \bibinfo{title}{Mri super-resolution reconstruction using efficient
  diffusion probabilistic model with residual shifting},
\newblock \bibinfo{journal}{Physics in Medicine \& Biology}
  \bibinfo{volume}{70} (\bibinfo{year}{2025}) \bibinfo{pages}{125008}.
  \URLprefix \url{https://dx.doi.org/10.1088/1361-6560/ade049}.
  \DOIprefix\doi{10.1088/1361-6560/ade049}.
\bibitem[{Chartsias et~al.(2017)Chartsias, Joyce, Dharmakumar, and
  Tsaftaris}]{chartsias2017adversarial}
\bibinfo{author}{A.~Chartsias}, \bibinfo{author}{T.~Joyce},
  \bibinfo{author}{R.~Dharmakumar}, \bibinfo{author}{S.~A. Tsaftaris},
\newblock \bibinfo{title}{Adversarial image synthesis for unpaired multi-modal
  cardiac data},
\newblock in: \bibinfo{booktitle}{International workshop on simulation and
  synthesis in medical imaging}, \bibinfo{organization}{Springer},
  \bibinfo{year}{2017}, pp. \bibinfo{pages}{3--13}.
\bibitem[{Nan et~al.(2017)Nan, Huang, and Guo}]{8062235}
\bibinfo{author}{Y.~Nan}, \bibinfo{author}{X.~Huang}, \bibinfo{author}{Y.~J.
  Guo},
\newblock \bibinfo{title}{Passive synthetic aperture radar imaging with
  piecewise constant doppler algorithm},
\newblock in: \bibinfo{booktitle}{2017 IEEE-APS Topical Conference on Antennas
  and Propagation in Wireless Communications (APWC)}, \bibinfo{year}{2017}, pp.
  \bibinfo{pages}{41--44}. \DOIprefix\doi{10.1109/APWC.2017.8062235}.
\bibitem[{Eidex et~al.(2024)Eidex, Wang, Safari, Elder, Wynne, Wang, Shu, Mao,
  and Yang}]{eidex2024high}
\bibinfo{author}{Z.~Eidex}, \bibinfo{author}{J.~Wang},
  \bibinfo{author}{M.~Safari}, \bibinfo{author}{E.~Elder},
  \bibinfo{author}{J.~Wynne}, \bibinfo{author}{T.~Wang}, \bibinfo{author}{H.-K.
  Shu}, \bibinfo{author}{H.~Mao}, \bibinfo{author}{X.~Yang},
\newblock \bibinfo{title}{High-resolution 3t to 7t adc map synthesis with a
  hybrid cnn-transformer model},
\newblock \bibinfo{journal}{Medical Physics} \bibinfo{volume}{51}
  (\bibinfo{year}{2024}) \bibinfo{pages}{4380--4388}.
\bibitem[{Litjens et~al.(2017)Litjens, Kooi, Bejnordi, Setio, Ciompi,
  Ghafoorian, Van Der~Laak, Van~Ginneken, and S{\'a}nchez}]{litjens2017survey}
\bibinfo{author}{G.~Litjens}, \bibinfo{author}{T.~Kooi}, \bibinfo{author}{B.~E.
  Bejnordi}, \bibinfo{author}{A.~A.~A. Setio}, \bibinfo{author}{F.~Ciompi},
  \bibinfo{author}{M.~Ghafoorian}, \bibinfo{author}{J.~A. Van Der~Laak},
  \bibinfo{author}{B.~Van~Ginneken}, \bibinfo{author}{C.~I. S{\'a}nchez},
\newblock \bibinfo{title}{A survey on deep learning in medical image analysis},
\newblock \bibinfo{journal}{Medical image analysis} \bibinfo{volume}{42}
  (\bibinfo{year}{2017}) \bibinfo{pages}{60--88}.
\bibitem[{Esteva et~al.(2017)Esteva, Kuprel, Novoa, Ko, Swetter, Blau, and
  Thrun}]{esteva2017dermatologist}
\bibinfo{author}{A.~Esteva}, \bibinfo{author}{B.~Kuprel},
  \bibinfo{author}{R.~A. Novoa}, \bibinfo{author}{J.~Ko},
  \bibinfo{author}{S.~M. Swetter}, \bibinfo{author}{H.~M. Blau},
  \bibinfo{author}{S.~Thrun},
\newblock \bibinfo{title}{Dermatologist-level classification of skin cancer
  with deep neural networks},
\newblock \bibinfo{journal}{nature} \bibinfo{volume}{542}
  (\bibinfo{year}{2017}) \bibinfo{pages}{115--118}.
\bibitem[{Safari et~al.(2022)Safari, Beiki, Ameri, Toudeshki, Fatemi, and
  Archambault}]{safari2022shuffle}
\bibinfo{author}{M.~Safari}, \bibinfo{author}{M.~Beiki},
  \bibinfo{author}{A.~Ameri}, \bibinfo{author}{S.~H. Toudeshki},
  \bibinfo{author}{A.~Fatemi}, \bibinfo{author}{L.~Archambault},
\newblock \bibinfo{title}{Shuffle-resnet: Deep learning for predicting lgg idh1
  mutation from multicenter anatomical mri sequences},
\newblock \bibinfo{journal}{Biomedical Physics \& Engineering Express}
  \bibinfo{volume}{8} (\bibinfo{year}{2022}) \bibinfo{pages}{065036}.
\bibitem[{Murphy(2023)}]{murphy2023gan}
\bibinfo{author}{K.~P. Murphy},
\newblock \bibinfo{title}{Generative adversarial networks},
\newblock in: \bibinfo{booktitle}{Probabilistic machine learning: Advanced
  topics}, \bibinfo{publisher}{MIT Press}, \bibinfo{year}{2023}, pp.
  \bibinfo{pages}{883--914}.
\bibitem[{Goodfellow et~al.(2020)Goodfellow, Pouget-Abadie, Mirza, Xu,
  Warde-Farley, Ozair, Courville, and Bengio}]{goodfellow2020generative}
\bibinfo{author}{I.~Goodfellow}, \bibinfo{author}{J.~Pouget-Abadie},
  \bibinfo{author}{M.~Mirza}, \bibinfo{author}{B.~Xu},
  \bibinfo{author}{D.~Warde-Farley}, \bibinfo{author}{S.~Ozair},
  \bibinfo{author}{A.~Courville}, \bibinfo{author}{Y.~Bengio},
\newblock \bibinfo{title}{Generative adversarial networks},
\newblock \bibinfo{journal}{Communications of the ACM} \bibinfo{volume}{63}
  (\bibinfo{year}{2020}) \bibinfo{pages}{139--144}.
\bibitem[{Kingma and Welling(2013)}]{kingma2013auto}
\bibinfo{author}{D.~P. Kingma}, \bibinfo{author}{M.~Welling},
\newblock \bibinfo{title}{Auto-encoding variational bayes},
\newblock \bibinfo{journal}{arXiv preprint arXiv:1312.6114}
  (\bibinfo{year}{2013}).
\bibitem[{Ho et~al.(2020)Ho, Jain, and Abbeel}]{ho2020denoising}
\bibinfo{author}{J.~Ho}, \bibinfo{author}{A.~Jain},
  \bibinfo{author}{P.~Abbeel},
\newblock \bibinfo{title}{Denoising diffusion probabilistic models},
\newblock \bibinfo{journal}{Advances in neural information processing systems}
  \bibinfo{volume}{33} (\bibinfo{year}{2020}) \bibinfo{pages}{6840--6851}.
\bibitem[{Yi et~al.(2019)Yi, Walia, and Babyn}]{yi2019generative}
\bibinfo{author}{X.~Yi}, \bibinfo{author}{E.~Walia},
  \bibinfo{author}{P.~Babyn},
\newblock \bibinfo{title}{Generative adversarial network in medical imaging: A
  review},
\newblock \bibinfo{journal}{Medical image analysis} \bibinfo{volume}{58}
  (\bibinfo{year}{2019}) \bibinfo{pages}{101552}.
\bibitem[{Lei et~al.(2019)Lei, Harms, Wang, Liu, Shu, Jani, Curran, Mao, Liu,
  and Yang}]{lei2019mri}
\bibinfo{author}{Y.~Lei}, \bibinfo{author}{J.~Harms},
  \bibinfo{author}{T.~Wang}, \bibinfo{author}{Y.~Liu}, \bibinfo{author}{H.-K.
  Shu}, \bibinfo{author}{A.~B. Jani}, \bibinfo{author}{W.~J. Curran},
  \bibinfo{author}{H.~Mao}, \bibinfo{author}{T.~Liu},
  \bibinfo{author}{X.~Yang},
\newblock \bibinfo{title}{Mri-only based synthetic ct generation using dense
  cycle consistent generative adversarial networks},
\newblock \bibinfo{journal}{Medical physics} \bibinfo{volume}{46}
  (\bibinfo{year}{2019}) \bibinfo{pages}{3565--3581}.
\bibitem[{Dong et~al.(2019)Dong, Lei, Wang, Thomas, Tang, Curran, Liu, and
  Yang}]{dong2019automatic}
\bibinfo{author}{X.~Dong}, \bibinfo{author}{Y.~Lei}, \bibinfo{author}{T.~Wang},
  \bibinfo{author}{M.~Thomas}, \bibinfo{author}{L.~Tang},
  \bibinfo{author}{W.~J. Curran}, \bibinfo{author}{T.~Liu},
  \bibinfo{author}{X.~Yang},
\newblock \bibinfo{title}{Automatic multiorgan segmentation in thorax ct images
  using u-net-gan},
\newblock \bibinfo{journal}{Medical physics} \bibinfo{volume}{46}
  (\bibinfo{year}{2019}) \bibinfo{pages}{2157--2168}.
\bibitem[{Ran et~al.(2019)Ran, Hu, Chen, Chen, Sun, Zhou, and
  Zhang}]{ran2019denoising}
\bibinfo{author}{M.~Ran}, \bibinfo{author}{J.~Hu}, \bibinfo{author}{Y.~Chen},
  \bibinfo{author}{H.~Chen}, \bibinfo{author}{H.~Sun},
  \bibinfo{author}{J.~Zhou}, \bibinfo{author}{Y.~Zhang},
\newblock \bibinfo{title}{Denoising of 3d magnetic resonance images using a
  residual encoder--decoder wasserstein generative adversarial network},
\newblock \bibinfo{journal}{Medical image analysis} \bibinfo{volume}{55}
  (\bibinfo{year}{2019}) \bibinfo{pages}{165--180}.
\bibitem[{Abbasi et~al.(2024)Abbasi, Lan, Choupan, Sheikh-Bahaei, Pandey, and
  Varghese}]{abbasi2024deep}
\bibinfo{author}{S.~Abbasi}, \bibinfo{author}{H.~Lan},
  \bibinfo{author}{J.~Choupan}, \bibinfo{author}{N.~Sheikh-Bahaei},
  \bibinfo{author}{G.~Pandey}, \bibinfo{author}{B.~Varghese},
\newblock \bibinfo{title}{Deep learning for the harmonization of structural mri
  scans: a survey},
\newblock \bibinfo{journal}{BioMedical Engineering OnLine} \bibinfo{volume}{23}
  (\bibinfo{year}{2024}) \bibinfo{pages}{1--42}.
\bibitem[{Hewlett et~al.(2024)Hewlett, Petrov, Johnson, and
  Drangova}]{hewlett2024deep}
\bibinfo{author}{M.~Hewlett}, \bibinfo{author}{I.~Petrov},
  \bibinfo{author}{P.~M. Johnson}, \bibinfo{author}{M.~Drangova},
\newblock \bibinfo{title}{Deep-learning-based motion correction using
  multichannel mri data: a study using simulated artifacts in the fastmri
  dataset},
\newblock \bibinfo{journal}{NMR in Biomedicine}  (\bibinfo{year}{2024})
  \bibinfo{pages}{e5179}.
\bibitem[{Wu et~al.(2023)Wu, Liu, White, and Deng}]{wu2023image}
\bibinfo{author}{Y.~Wu}, \bibinfo{author}{J.~Liu}, \bibinfo{author}{G.~M.
  White}, \bibinfo{author}{J.~Deng},
\newblock \bibinfo{title}{Image-based motion artifact reduction on liver
  dynamic contrast enhanced mri},
\newblock \bibinfo{journal}{Physica Medica} \bibinfo{volume}{105}
  (\bibinfo{year}{2023}) \bibinfo{pages}{102509}.
\bibitem[{Yoshida et~al.(2022)Yoshida, Kageyama, Akai, Yasaka, Sugawara, Okada,
  and Kunimatsu}]{yoshida2022motion}
\bibinfo{author}{N.~Yoshida}, \bibinfo{author}{H.~Kageyama},
  \bibinfo{author}{H.~Akai}, \bibinfo{author}{K.~Yasaka},
  \bibinfo{author}{H.~Sugawara}, \bibinfo{author}{Y.~Okada},
  \bibinfo{author}{A.~Kunimatsu},
\newblock \bibinfo{title}{Motion correction in mr image for analysis of vsrad
  using generative adversarial network},
\newblock \bibinfo{journal}{PloS one} \bibinfo{volume}{17}
  (\bibinfo{year}{2022}) \bibinfo{pages}{e0274576}.
\bibitem[{Zhao et~al.(2020)Zhao, Li, Jiao, Du, and Fan}]{zhao20203d}
\bibinfo{author}{C.~Zhao}, \bibinfo{author}{H.~Li}, \bibinfo{author}{Z.~Jiao},
  \bibinfo{author}{T.~Du}, \bibinfo{author}{Y.~Fan},
\newblock \bibinfo{title}{A 3d convolutional encapsulated long short-term
  memory (3dconv-lstm) model for denoising fmri data},
\newblock in: \bibinfo{booktitle}{Medical Image Computing and Computer Assisted
  Intervention--MICCAI 2020: 23rd International Conference, Lima, Peru, October
  4--8, 2020, Proceedings, Part VII 23}, \bibinfo{organization}{Springer},
  \bibinfo{year}{2020}, pp. \bibinfo{pages}{479--488}.
\bibitem[{Mirza and Osindero(2014)}]{mirza2014conditional}
\bibinfo{author}{M.~Mirza}, \bibinfo{author}{S.~Osindero},
\newblock \bibinfo{title}{Conditional generative adversarial nets},
\newblock \bibinfo{journal}{arXiv preprint arXiv:1411.1784}
  (\bibinfo{year}{2014}).
\bibitem[{Johnson and Drangova(2019)}]{johnson2019conditional}
\bibinfo{author}{P.~M. Johnson}, \bibinfo{author}{M.~Drangova},
\newblock \bibinfo{title}{Conditional generative adversarial network for 3d
  rigid-body motion correction in mri},
\newblock \bibinfo{journal}{Magnetic resonance in medicine}
  \bibinfo{volume}{82} (\bibinfo{year}{2019}) \bibinfo{pages}{901--910}.
\bibitem[{Usui et~al.(2023)Usui, Muro, Shibukawa, Goto, Ogawa, Sakano, Kyogoku,
  and Daida}]{usui2023evaluation}
\bibinfo{author}{K.~Usui}, \bibinfo{author}{I.~Muro},
  \bibinfo{author}{S.~Shibukawa}, \bibinfo{author}{M.~Goto},
  \bibinfo{author}{K.~Ogawa}, \bibinfo{author}{Y.~Sakano},
  \bibinfo{author}{S.~Kyogoku}, \bibinfo{author}{H.~Daida},
\newblock \bibinfo{title}{Evaluation of motion artefact reduction depending on
  the artefacts’ directions in head mri using conditional generative
  adversarial networks},
\newblock \bibinfo{journal}{Scientific Reports} \bibinfo{volume}{13}
  (\bibinfo{year}{2023}) \bibinfo{pages}{8526}.
\bibitem[{Bao et~al.(2022)Bao, Chen, Bai, Li, Liu, Li, Zhang, Wang, and
  Liu}]{bao2022retrospective}
\bibinfo{author}{Q.~Bao}, \bibinfo{author}{Y.~Chen}, \bibinfo{author}{C.~Bai},
  \bibinfo{author}{P.~Li}, \bibinfo{author}{K.~Liu}, \bibinfo{author}{Z.~Li},
  \bibinfo{author}{Z.~Zhang}, \bibinfo{author}{J.~Wang},
  \bibinfo{author}{C.~Liu},
\newblock \bibinfo{title}{Retrospective motion correction for
  preclinical/clinical magnetic resonance imaging based on a conditional
  generative adversarial network with entropy loss},
\newblock \bibinfo{journal}{NMR in Biomedicine} \bibinfo{volume}{35}
  (\bibinfo{year}{2022}) \bibinfo{pages}{e4809}.
\bibitem[{Ghodrati et~al.(2021)Ghodrati, Bydder, Ali, Gao, Prosper, Nguyen, and
  Hu}]{ghodrati2021retrospective}
\bibinfo{author}{V.~Ghodrati}, \bibinfo{author}{M.~Bydder},
  \bibinfo{author}{F.~Ali}, \bibinfo{author}{C.~Gao},
  \bibinfo{author}{A.~Prosper}, \bibinfo{author}{K.-L. Nguyen},
  \bibinfo{author}{P.~Hu},
\newblock \bibinfo{title}{Retrospective respiratory motion correction in
  cardiac cine mri reconstruction using adversarial autoencoder and
  unsupervised learning},
\newblock \bibinfo{journal}{NMR in Biomedicine} \bibinfo{volume}{34}
  (\bibinfo{year}{2021}) \bibinfo{pages}{e4433}.
\bibitem[{Zhu et~al.(2017)Zhu, Park, Isola, and Efros}]{zhu2017unpaired}
\bibinfo{author}{J.-Y. Zhu}, \bibinfo{author}{T.~Park},
  \bibinfo{author}{P.~Isola}, \bibinfo{author}{A.~A. Efros},
\newblock \bibinfo{title}{Unpaired image-to-image translation using
  cycle-consistent adversarial networks},
\newblock in: \bibinfo{booktitle}{Proceedings of the IEEE international
  conference on computer vision}, \bibinfo{year}{2017}, pp.
  \bibinfo{pages}{2223--2232}.
\bibitem[{Oh et~al.(2021)Oh, Lee, and Ye}]{oh2021unpaired}
\bibinfo{author}{G.~Oh}, \bibinfo{author}{J.~E. Lee}, \bibinfo{author}{J.~C.
  Ye},
\newblock \bibinfo{title}{Unpaired mr motion artifact deep learning using
  outlier-rejecting bootstrap aggregation},
\newblock \bibinfo{journal}{IEEE Transactions on Medical Imaging}
  \bibinfo{volume}{40} (\bibinfo{year}{2021}) \bibinfo{pages}{3125--3139}.
\bibitem[{Wu et~al.(2023)Wu, Li, Zhang, Lai, Feng, and
  Huang}]{wu2023unsupervised}
\bibinfo{author}{B.~Wu}, \bibinfo{author}{C.~Li}, \bibinfo{author}{J.~Zhang},
  \bibinfo{author}{H.~Lai}, \bibinfo{author}{Q.~Feng},
  \bibinfo{author}{M.~Huang},
\newblock \bibinfo{title}{Unsupervised dual-domain disentangled network for
  removal of rigid motion artifacts in mri},
\newblock \bibinfo{journal}{Computers in Biology and Medicine}
  \bibinfo{volume}{165} (\bibinfo{year}{2023}) \bibinfo{pages}{107373}.
\bibitem[{Safari et~al.(2024)Safari, Yang, Chang, Qiu, Fatemi, and
  Archambault}]{safari2024unsupervised}
\bibinfo{author}{M.~Safari}, \bibinfo{author}{X.~Yang}, \bibinfo{author}{C.-W.
  Chang}, \bibinfo{author}{R.~L. Qiu}, \bibinfo{author}{A.~Fatemi},
  \bibinfo{author}{L.~Archambault},
\newblock \bibinfo{title}{Unsupervised mri motion artifact disentanglement:
  introducing maudgan},
\newblock \bibinfo{journal}{Physics in Medicine \& Biology}
  \bibinfo{volume}{69} (\bibinfo{year}{2024}) \bibinfo{pages}{115057}.
\bibitem[{Liu et~al.(2021)Liu, Thung, Qu, Lin, Shen, and Yap}]{liu2021learning}
\bibinfo{author}{S.~Liu}, \bibinfo{author}{K.-H. Thung},
  \bibinfo{author}{L.~Qu}, \bibinfo{author}{W.~Lin}, \bibinfo{author}{D.~Shen},
  \bibinfo{author}{P.-T. Yap},
\newblock \bibinfo{title}{Learning mri artefact removal with unpaired data},
\newblock \bibinfo{journal}{Nature Machine Intelligence} \bibinfo{volume}{3}
  (\bibinfo{year}{2021}) \bibinfo{pages}{60--67}.
\bibitem[{Sohl-Dickstein et~al.(2015)Sohl-Dickstein, Weiss, Maheswaranathan,
  and Ganguli}]{sohl2015deep}
\bibinfo{author}{J.~Sohl-Dickstein}, \bibinfo{author}{E.~Weiss},
  \bibinfo{author}{N.~Maheswaranathan}, \bibinfo{author}{S.~Ganguli},
\newblock \bibinfo{title}{Deep unsupervised learning using nonequilibrium
  thermodynamics},
\newblock in: \bibinfo{booktitle}{International conference on machine
  learning}, \bibinfo{organization}{PMLR}, \bibinfo{year}{2015}, pp.
  \bibinfo{pages}{2256--2265}.
\bibitem[{Levac et~al.(2024)Levac, Kumar, Jalal, and
  Tamir}]{levac2024accelerated}
\bibinfo{author}{B.~Levac}, \bibinfo{author}{S.~Kumar},
  \bibinfo{author}{A.~Jalal}, \bibinfo{author}{J.~I. Tamir},
\newblock \bibinfo{title}{Accelerated motion correction with deep generative
  diffusion models},
\newblock \bibinfo{journal}{Magnetic Resonance in Medicine}
  \bibinfo{volume}{92} (\bibinfo{year}{2024}) \bibinfo{pages}{853--868}.
\bibitem[{Safari et~al.(2024)Safari, Yang, Fatemi, and
  Archambault}]{safari2024mri}
\bibinfo{author}{M.~Safari}, \bibinfo{author}{X.~Yang},
  \bibinfo{author}{A.~Fatemi}, \bibinfo{author}{L.~Archambault},
\newblock \bibinfo{title}{Mri motion artifact reduction using a conditional
  diffusion probabilistic model (mar-cdpm)},
\newblock \bibinfo{journal}{Medical Physics} \bibinfo{volume}{51}
  (\bibinfo{year}{2024}) \bibinfo{pages}{2598--2610}.
\bibitem[{Jung et~al.(2024)Jung, Kim, Park, and Kim}]{jung2024image}
\bibinfo{author}{H.~K. Jung}, \bibinfo{author}{K.~Kim}, \bibinfo{author}{J.~E.
  Park}, \bibinfo{author}{N.~Kim},
\newblock \bibinfo{title}{Image-based generative artificial intelligence in
  radiology: comprehensive updates},
\newblock \bibinfo{journal}{Korean Journal of Radiology} \bibinfo{volume}{25}
  (\bibinfo{year}{2024}) \bibinfo{pages}{959}.
\bibitem[{Antun et~al.(2020)Antun, Renna, Poon, Adcock, and
  Hansen}]{antun2020instabilities}
\bibinfo{author}{V.~Antun}, \bibinfo{author}{F.~Renna},
  \bibinfo{author}{C.~Poon}, \bibinfo{author}{B.~Adcock},
  \bibinfo{author}{A.~C. Hansen},
\newblock \bibinfo{title}{On instabilities of deep learning in image
  reconstruction and the potential costs of ai},
\newblock \bibinfo{journal}{Proceedings of the National Academy of Sciences}
  \bibinfo{volume}{117} (\bibinfo{year}{2020}) \bibinfo{pages}{30088--30095}.
\bibitem[{Kazeminia et~al.(2020)Kazeminia, Baur, Kuijper, Van~Ginneken, Navab,
  Albarqouni, and Mukhopadhyay}]{kazeminia2020gans}
\bibinfo{author}{S.~Kazeminia}, \bibinfo{author}{C.~Baur},
  \bibinfo{author}{A.~Kuijper}, \bibinfo{author}{B.~Van~Ginneken},
  \bibinfo{author}{N.~Navab}, \bibinfo{author}{S.~Albarqouni},
  \bibinfo{author}{A.~Mukhopadhyay},
\newblock \bibinfo{title}{Gans for medical image analysis},
\newblock \bibinfo{journal}{Artificial intelligence in medicine}
  \bibinfo{volume}{109} (\bibinfo{year}{2020}) \bibinfo{pages}{101938}.
\bibitem[{Cohen et~al.(2018)Cohen, Luck, and Honari}]{cohen2018distribution}
\bibinfo{author}{J.~P. Cohen}, \bibinfo{author}{M.~Luck},
  \bibinfo{author}{S.~Honari},
\newblock \bibinfo{title}{Distribution matching losses can hallucinate features
  in medical image translation},
\newblock in: \bibinfo{booktitle}{International conference on medical image
  computing and computer-assisted intervention},
  \bibinfo{organization}{Springer}, \bibinfo{year}{2018}, pp.
  \bibinfo{pages}{529--536}.
\bibitem[{Bhadra et~al.(2021)Bhadra, Kelkar, Brooks, and Anastasio}]{9424044}
\bibinfo{author}{S.~Bhadra}, \bibinfo{author}{V.~A. Kelkar},
  \bibinfo{author}{F.~J. Brooks}, \bibinfo{author}{M.~A. Anastasio},
\newblock \bibinfo{title}{On hallucinations in tomographic image
  reconstruction},
\newblock \bibinfo{journal}{IEEE Transactions on Medical Imaging}
  \bibinfo{volume}{40} (\bibinfo{year}{2021}) \bibinfo{pages}{3249--3260}.
  \DOIprefix\doi{10.1109/TMI.2021.3077857}.
\bibitem[{Kendall and Gal(2017)}]{kendall2017uncertainties}
\bibinfo{author}{A.~Kendall}, \bibinfo{author}{Y.~Gal},
\newblock \bibinfo{title}{What uncertainties do we need in bayesian deep
  learning for computer vision?},
\newblock \bibinfo{journal}{Advances in neural information processing systems}
  \bibinfo{volume}{30} (\bibinfo{year}{2017}).
\bibitem[{Ahmadi et~al.(2025)Ahmadi, Biswas, Lin, Vrionis, Hashemi, and
  Tang}]{ahmadi2025physics}
\bibinfo{author}{M.~Ahmadi}, \bibinfo{author}{D.~Biswas},
  \bibinfo{author}{M.~Lin}, \bibinfo{author}{F.~D. Vrionis},
  \bibinfo{author}{J.~Hashemi}, \bibinfo{author}{Y.~Tang},
\newblock \bibinfo{title}{Physics-informed machine learning for advancing
  computational medical imaging: integrating data-driven approaches with
  fundamental physical principles},
\newblock \bibinfo{journal}{Artificial Intelligence Review}
  \bibinfo{volume}{58} (\bibinfo{year}{2025}) \bibinfo{pages}{1--49}.
\bibitem[{Safari et~al.(2025)Safari, Eidex, Pan, Qiu, and
  Yang}]{https://doi.org/10.1002/mp.17675}
\bibinfo{author}{M.~Safari}, \bibinfo{author}{Z.~Eidex},
  \bibinfo{author}{S.~Pan}, \bibinfo{author}{R.~L.~J. Qiu},
  \bibinfo{author}{X.~Yang},
\newblock \bibinfo{title}{Self-supervised adversarial diffusion models for fast
  mri reconstruction},
\newblock \bibinfo{journal}{Medical Physics} \bibinfo{volume}{52}
  (\bibinfo{year}{2025}) \bibinfo{pages}{3888--3899}. \URLprefix
  \url{https://aapm.onlinelibrary.wiley.com/doi/abs/10.1002/mp.17675}.
  \DOIprefix\doi{https://doi.org/10.1002/mp.17675}.
  \href{http://arxiv.org/abs/https://aapm.onlinelibrary.wiley.com/doi/pdf/10.1002/mp.17675}{{\tt
  arXiv:https://aapm.onlinelibrary.wiley.com/doi/pdf/10.1002/mp.17675}}.
\bibitem[{Chen et~al.(2021)Chen, Schaeffter, Kolbitsch, and
  Kofler}]{chen2021ground}
\bibinfo{author}{D.~Chen}, \bibinfo{author}{T.~Schaeffter},
  \bibinfo{author}{C.~Kolbitsch}, \bibinfo{author}{A.~Kofler},
\newblock \bibinfo{title}{Ground-truth-free deep learning for artefacts
  reduction in 2d radial cardiac cine mri using a synthetically generated
  dataset},
\newblock \bibinfo{journal}{Physics in Medicine \& Biology}
  \bibinfo{volume}{66} (\bibinfo{year}{2021}) \bibinfo{pages}{095005}.
\bibitem[{Dabrowski et~al.(2022)Dabrowski, Courvoisier, Falcone, Klauser,
  Songeon, Kocher, Chopard, and Lazeyras}]{Dabrowski2022}
\bibinfo{author}{O.~Dabrowski}, \bibinfo{author}{S.~Courvoisier},
  \bibinfo{author}{J.-L. Falcone}, \bibinfo{author}{A.~Klauser},
  \bibinfo{author}{J.~Songeon}, \bibinfo{author}{M.~Kocher},
  \bibinfo{author}{B.~Chopard}, \bibinfo{author}{F.~Lazeyras},
\newblock \bibinfo{title}{Choreography controlled (choco) brain mri artifact
  generation for labeled motion-corrupted datasets},
\newblock \bibinfo{journal}{Physica Medica: European Journal of Medical
  Physics} \bibinfo{volume}{102} (\bibinfo{year}{2022})
  \bibinfo{pages}{79--87}. \URLprefix
  \url{https://doi.org/10.1016/j.ejmp.2022.09.005}.
  \DOIprefix\doi{10.1016/j.ejmp.2022.09.005}.
\bibitem[{Shaw et~al.(2020)Shaw, Sudre, Varsavsky, Ourselin, and
  Cardoso}]{shaw2020k}
\bibinfo{author}{R.~Shaw}, \bibinfo{author}{C.~H. Sudre},
  \bibinfo{author}{T.~Varsavsky}, \bibinfo{author}{S.~Ourselin},
  \bibinfo{author}{M.~J. Cardoso},
\newblock \bibinfo{title}{A k-space model of movement artefacts: application to
  segmentation augmentation and artefact removal},
\newblock \bibinfo{journal}{IEEE transactions on medical imaging}
  \bibinfo{volume}{39} (\bibinfo{year}{2020}) \bibinfo{pages}{2881--2892}.
\bibitem[{P{\'e}rez-Garc{\'\i}a et~al.(2021)P{\'e}rez-Garc{\'\i}a, Sparks, and
  Ourselin}]{perez2021torchio}
\bibinfo{author}{F.~P{\'e}rez-Garc{\'\i}a}, \bibinfo{author}{R.~Sparks},
  \bibinfo{author}{S.~Ourselin},
\newblock \bibinfo{title}{Torchio: a python library for efficient loading,
  preprocessing, augmentation and patch-based sampling of medical images in
  deep learning},
\newblock \bibinfo{journal}{Computer methods and programs in biomedicine}
  \bibinfo{volume}{208} (\bibinfo{year}{2021}) \bibinfo{pages}{106236}.
\bibitem[{Tamada et~al.(2020)Tamada, Kromrey, Ichikawa, Onishi, and
  Motosugi}]{tamada2020motion}
\bibinfo{author}{D.~Tamada}, \bibinfo{author}{M.-L. Kromrey},
  \bibinfo{author}{S.~Ichikawa}, \bibinfo{author}{H.~Onishi},
  \bibinfo{author}{U.~Motosugi},
\newblock \bibinfo{title}{Motion artifact reduction using a convolutional
  neural network for dynamic contrast enhanced mr imaging of the liver},
\newblock \bibinfo{journal}{Magnetic resonance in medical sciences}
  \bibinfo{volume}{19} (\bibinfo{year}{2020}) \bibinfo{pages}{64--76}.
\bibitem[{Reddy et~al.(2024)Reddy, Yogananda, Truong, Madhuranthakam, Maldjian,
  and Fei}]{reddy2024gan}
\bibinfo{author}{V.~V. R.~K. Reddy}, \bibinfo{author}{C.~G.~B. Yogananda},
  \bibinfo{author}{N.~C. Truong}, \bibinfo{author}{A.~J. Madhuranthakam},
  \bibinfo{author}{J.~A. Maldjian}, \bibinfo{author}{B.~Fei},
\newblock \bibinfo{title}{Gan-based motion artifact correction of 3d mr volumes
  using an image-to-image translation algorithm},
\newblock in: \bibinfo{booktitle}{Medical Imaging 2024: Clinical and Biomedical
  Imaging}, volume \bibinfo{volume}{12930}, \bibinfo{organization}{SPIE},
  \bibinfo{year}{2024}, pp. \bibinfo{pages}{529--536}.
\bibitem[{Eichhorn et~al.(2023)Eichhorn, Hammernik, Spieker, Epp, Rueckert,
  Preibisch, and Schnabel}]{eichhorn2023deep}
\bibinfo{author}{H.~Eichhorn}, \bibinfo{author}{K.~Hammernik},
  \bibinfo{author}{V.~Spieker}, \bibinfo{author}{S.~M. Epp},
  \bibinfo{author}{D.~Rueckert}, \bibinfo{author}{C.~Preibisch},
  \bibinfo{author}{J.~A. Schnabel},
\newblock \bibinfo{title}{Deep learning-based detection of motion-affected
  k-space lines for t2*-weighted mri},
\newblock \bibinfo{journal}{arXiv preprint arXiv:2303.10987}
  (\bibinfo{year}{2023}).
\bibitem[{Weng et~al.(2024)Weng, Bhupathiraju, Samant, Dresner, Wu, and
  Samant}]{weng2024convolutional}
\bibinfo{author}{J.~Weng}, \bibinfo{author}{S.~Bhupathiraju},
  \bibinfo{author}{T.~Samant}, \bibinfo{author}{A.~Dresner},
  \bibinfo{author}{J.~Wu}, \bibinfo{author}{S.~Samant},
\newblock \bibinfo{title}{Convolutional lstm model for cine image prediction of
  abdominal motion},
\newblock \bibinfo{journal}{Physics in Medicine \& Biology}
  \bibinfo{volume}{69} (\bibinfo{year}{2024}) \bibinfo{pages}{085024}.
\bibitem[{Pawar et~al.(2020)Pawar, Chen, Seah, Law, Close, and
  Egan}]{pawar2020clinical}
\bibinfo{author}{K.~Pawar}, \bibinfo{author}{Z.~Chen},
  \bibinfo{author}{J.~Seah}, \bibinfo{author}{M.~Law},
  \bibinfo{author}{T.~Close}, \bibinfo{author}{G.~Egan},
\newblock \bibinfo{title}{Clinical utility of deep learning motion correction
  for t1 weighted mprage mr images},
\newblock \bibinfo{journal}{European Journal of Radiology}
  \bibinfo{volume}{133} (\bibinfo{year}{2020}) \bibinfo{pages}{109384}.
\bibitem[{Al-Masni et~al.(2022)Al-Masni, Lee, Yi, Kim, Gho, Choi, and
  Kim}]{al2022stacked}
\bibinfo{author}{M.~A. Al-Masni}, \bibinfo{author}{S.~Lee},
  \bibinfo{author}{J.~Yi}, \bibinfo{author}{S.~Kim}, \bibinfo{author}{S.-M.
  Gho}, \bibinfo{author}{Y.~H. Choi}, \bibinfo{author}{D.-H. Kim},
\newblock \bibinfo{title}{Stacked u-nets with self-assisted priors towards
  robust correction of rigid motion artifact in brain mri},
\newblock \bibinfo{journal}{NeuroImage} \bibinfo{volume}{259}
  (\bibinfo{year}{2022}) \bibinfo{pages}{119411}.
\bibitem[{Pawar et~al.(2022)Pawar, Chen, Shah, and Egan}]{pawar2022suppressing}
\bibinfo{author}{K.~Pawar}, \bibinfo{author}{Z.~Chen}, \bibinfo{author}{N.~J.
  Shah}, \bibinfo{author}{G.~F. Egan},
\newblock \bibinfo{title}{Suppressing motion artefacts in mri using an
  inception-resnet network with motion simulation augmentation},
\newblock \bibinfo{journal}{NMR in Biomedicine} \bibinfo{volume}{35}
  (\bibinfo{year}{2022}) \bibinfo{pages}{e4225}.
\bibitem[{Safari et~al.(2025)Safari, Wang, Li, Eidex, Qiu, Chang, Mao, and
  Yang}]{safari2025mrimotioncorrectionefficient}
\bibinfo{author}{M.~Safari}, \bibinfo{author}{S.~Wang},
  \bibinfo{author}{Q.~Li}, \bibinfo{author}{Z.~Eidex},
  \bibinfo{author}{R.~L.~J. Qiu}, \bibinfo{author}{C.-W. Chang},
  \bibinfo{author}{H.~Mao}, \bibinfo{author}{X.~Yang}, \bibinfo{title}{Mri
  motion correction via efficient residual-guided denoising diffusion
  probabilistic models}, \bibinfo{year}{2025}. \URLprefix
  \url{https://arxiv.org/abs/2505.03498}.
  \href{http://arxiv.org/abs/2505.03498}{{\tt arXiv:2505.03498}}.
\bibitem[{Al-Masni et~al.(2023)Al-Masni, Lee, Al-Shamiri, Gho, Choi, and
  Kim}]{al2023knowledge}
\bibinfo{author}{M.~A. Al-Masni}, \bibinfo{author}{S.~Lee},
  \bibinfo{author}{A.~K. Al-Shamiri}, \bibinfo{author}{S.-M. Gho},
  \bibinfo{author}{Y.~H. Choi}, \bibinfo{author}{D.-H. Kim},
\newblock \bibinfo{title}{A knowledge interaction learning for multi-echo mri
  motion artifact correction towards better enhancement of swi},
\newblock \bibinfo{journal}{Computers in biology and medicine}
  \bibinfo{volume}{153} (\bibinfo{year}{2023}) \bibinfo{pages}{106553}.
\bibitem[{Zhang et~al.(2024)Zhang, Wang, Ye, Li, Zhuang, Yang, Fu, Chen, Gao,
  Ren et~al.}]{zhang2024anti}
\bibinfo{author}{Y.~Zhang}, \bibinfo{author}{X.~Wang}, \bibinfo{author}{M.~Ye},
  \bibinfo{author}{Z.~Li}, \bibinfo{author}{Y.~Zhuang},
  \bibinfo{author}{Q.~Yang}, \bibinfo{author}{Q.~Fu},
  \bibinfo{author}{R.~Chen}, \bibinfo{author}{E.~Gao},
  \bibinfo{author}{Y.~Ren}, et~al.,
\newblock \bibinfo{title}{Anti-motion ultrafast t2 mapping technique for
  quantitative detection of the normal-appearing corticospinal tract changes in
  subacute-chronic stroke patients with distal lesions},
\newblock \bibinfo{journal}{Academic Radiology} \bibinfo{volume}{31}
  (\bibinfo{year}{2024}) \bibinfo{pages}{2488--2500}.
\bibitem[{Lyu et~al.(2021)Lyu, Shan, Xie, Kwan, Otaki, Kuronuma, Li, and
  Wang}]{lyu2021cine}
\bibinfo{author}{Q.~Lyu}, \bibinfo{author}{H.~Shan}, \bibinfo{author}{Y.~Xie},
  \bibinfo{author}{A.~C. Kwan}, \bibinfo{author}{Y.~Otaki},
  \bibinfo{author}{K.~Kuronuma}, \bibinfo{author}{D.~Li},
  \bibinfo{author}{G.~Wang},
\newblock \bibinfo{title}{Cine cardiac mri motion artifact reduction using a
  recurrent neural network},
\newblock \bibinfo{journal}{IEEE Transactions on Medical Imaging}
  \bibinfo{volume}{40} (\bibinfo{year}{2021}) \bibinfo{pages}{2170--2181}.
\bibitem[{Lee et~al.(2021)Lee, Kim, and Park}]{lee2021mc2}
\bibinfo{author}{J.~Lee}, \bibinfo{author}{B.~Kim}, \bibinfo{author}{H.~Park},
\newblock \bibinfo{title}{Mc2-net: motion correction network for multi-contrast
  brain mri},
\newblock \bibinfo{journal}{Magnetic Resonance in Medicine}
  \bibinfo{volume}{86} (\bibinfo{year}{2021}) \bibinfo{pages}{1077--1092}.
\bibitem[{Liu et~al.(2020)Liu, Kocak, Supanich, and Deng}]{liu2020motion}
\bibinfo{author}{J.~Liu}, \bibinfo{author}{M.~Kocak},
  \bibinfo{author}{M.~Supanich}, \bibinfo{author}{J.~Deng},
\newblock \bibinfo{title}{Motion artifacts reduction in brain mri by means of a
  deep residual network with densely connected multi-resolution blocks
  (drn-dcmb)},
\newblock \bibinfo{journal}{Magnetic resonance imaging} \bibinfo{volume}{71}
  (\bibinfo{year}{2020}) \bibinfo{pages}{69--79}.
\bibitem[{Pirkl et~al.(2022)Pirkl, Cencini, Kurzawski, Waldmannstetter, Li,
  Sekuboyina, Endt, Peretti, Donatelli, Pasquariello
  et~al.}]{pirkl2022learning}
\bibinfo{author}{C.~M. Pirkl}, \bibinfo{author}{M.~Cencini},
  \bibinfo{author}{J.~W. Kurzawski}, \bibinfo{author}{D.~Waldmannstetter},
  \bibinfo{author}{H.~Li}, \bibinfo{author}{A.~Sekuboyina},
  \bibinfo{author}{S.~Endt}, \bibinfo{author}{L.~Peretti},
  \bibinfo{author}{G.~Donatelli}, \bibinfo{author}{R.~Pasquariello}, et~al.,
\newblock \bibinfo{title}{Learning residual motion correction for fast and
  robust 3d multiparametric mri},
\newblock \bibinfo{journal}{Medical Image Analysis} \bibinfo{volume}{77}
  (\bibinfo{year}{2022}) \bibinfo{pages}{102387}.
\bibitem[{Batchelor et~al.(2005)Batchelor, Atkinson, Irarrazaval, Hill, Hajnal,
  and Larkman}]{https://doi.org/10.1002/mrm.20656}
\bibinfo{author}{P.~G. Batchelor}, \bibinfo{author}{D.~Atkinson},
  \bibinfo{author}{P.~Irarrazaval}, \bibinfo{author}{D.~L.~G. Hill},
  \bibinfo{author}{J.~Hajnal}, \bibinfo{author}{D.~Larkman},
\newblock \bibinfo{title}{Matrix description of general motion correction
  applied to multishot images},
\newblock \bibinfo{journal}{Magnetic Resonance in Medicine}
  \bibinfo{volume}{54} (\bibinfo{year}{2005}) \bibinfo{pages}{1273--1280}.
  \URLprefix \url{https://onlinelibrary.wiley.com/doi/abs/10.1002/mrm.20656}.
  \DOIprefix\doi{https://doi.org/10.1002/mrm.20656}.
  \href{http://arxiv.org/abs/https://onlinelibrary.wiley.com/doi/pdf/10.1002/mrm.20656}{{\tt
  arXiv:https://onlinelibrary.wiley.com/doi/pdf/10.1002/mrm.20656}}.
\bibitem[{Hossbach et~al.(2023)Hossbach, Splitthoff, Cauley, Clifford, Polak,
  Lo, Meyer, and Maier}]{hossbach2023deep}
\bibinfo{author}{J.~Hossbach}, \bibinfo{author}{D.~N. Splitthoff},
  \bibinfo{author}{S.~Cauley}, \bibinfo{author}{B.~Clifford},
  \bibinfo{author}{D.~Polak}, \bibinfo{author}{W.-C. Lo},
  \bibinfo{author}{H.~Meyer}, \bibinfo{author}{A.~Maier},
\newblock \bibinfo{title}{Deep learning-based motion quantification from
  k-space for fast model-based magnetic resonance imaging motion correction},
\newblock \bibinfo{journal}{Medical physics} \bibinfo{volume}{50}
  (\bibinfo{year}{2023}) \bibinfo{pages}{2148--2161}.
\bibitem[{Dabrowski et~al.(2024)Dabrowski, Falcone, Klauser, Songeon, Kocher,
  Chopard, Lazeyras, and Courvoisier}]{dabrowski_SISMIK_2024}
\bibinfo{author}{O.~Dabrowski}, \bibinfo{author}{J.-L. Falcone},
  \bibinfo{author}{A.~Klauser}, \bibinfo{author}{J.~Songeon},
  \bibinfo{author}{M.~Kocher}, \bibinfo{author}{B.~Chopard},
  \bibinfo{author}{F.~Lazeyras}, \bibinfo{author}{S.~Courvoisier},
\newblock \bibinfo{title}{Sismik for brain mri: Deep-learning-based motion
  estimation and model-based motion correction in k-space},
\newblock \bibinfo{journal}{IEEE Transactions on Medical Imaging}
  (\bibinfo{year}{2024}) \bibinfo{pages}{1--1}.
  \DOIprefix\doi{10.1109/TMI.2024.3446450}.
\bibitem[{Haskell et~al.(2019)Haskell, Cauley, Bilgic, Hossbach, Splitthoff,
  Pfeuffer, Setsompop, and Wald}]{haskell2019network}
\bibinfo{author}{M.~W. Haskell}, \bibinfo{author}{S.~F. Cauley},
  \bibinfo{author}{B.~Bilgic}, \bibinfo{author}{J.~Hossbach},
  \bibinfo{author}{D.~N. Splitthoff}, \bibinfo{author}{J.~Pfeuffer},
  \bibinfo{author}{K.~Setsompop}, \bibinfo{author}{L.~L. Wald},
\newblock \bibinfo{title}{Network accelerated motion estimation and reduction
  (namer): convolutional neural network guided retrospective motion correction
  using a separable motion model},
\newblock \bibinfo{journal}{Magnetic resonance in medicine}
  \bibinfo{volume}{82} (\bibinfo{year}{2019}) \bibinfo{pages}{1452--1461}.
\bibitem[{Rizzuti et~al.(2022)Rizzuti, Sbrizzi, and
  Van~Leeuwen}]{rizzuti2022joint}
\bibinfo{author}{G.~Rizzuti}, \bibinfo{author}{A.~Sbrizzi},
  \bibinfo{author}{T.~Van~Leeuwen},
\newblock \bibinfo{title}{Joint retrospective motion correction and
  reconstruction for brain mri with a reference contrast},
\newblock \bibinfo{journal}{IEEE Transactions on Computational Imaging}
  \bibinfo{volume}{8} (\bibinfo{year}{2022}) \bibinfo{pages}{490--504}.
\bibitem[{Rizzuti et~al.(2024)Rizzuti, Schakel, Huttinga, Dankbaar, van
  Leeuwen, and Sbrizzi}]{rizzuti2024towards}
\bibinfo{author}{G.~Rizzuti}, \bibinfo{author}{T.~Schakel},
  \bibinfo{author}{N.~R. Huttinga}, \bibinfo{author}{J.~W. Dankbaar},
  \bibinfo{author}{T.~van Leeuwen}, \bibinfo{author}{A.~Sbrizzi},
\newblock \bibinfo{title}{Towards retrospective motion correction and
  reconstruction for clinical 3d brain mri protocols with a reference
  contrast},
\newblock \bibinfo{journal}{Magnetic Resonance Materials in Physics, Biology
  and Medicine}  (\bibinfo{year}{2024}) \bibinfo{pages}{1--17}.
\bibitem[{Morales et~al.(2019)Morales, Izquierdo-Garcia, Aganj,
  Kalpathy-Cramer, Rosen, and Catana}]{morales2019implementation}
\bibinfo{author}{M.~A. Morales}, \bibinfo{author}{D.~Izquierdo-Garcia},
  \bibinfo{author}{I.~Aganj}, \bibinfo{author}{J.~Kalpathy-Cramer},
  \bibinfo{author}{B.~R. Rosen}, \bibinfo{author}{C.~Catana},
\newblock \bibinfo{title}{Implementation and validation of a three-dimensional
  cardiac motion estimation network},
\newblock \bibinfo{journal}{Radiology: Artificial Intelligence}
  \bibinfo{volume}{1} (\bibinfo{year}{2019}) \bibinfo{pages}{e180080}.
\bibitem[{Gonzales et~al.(2021)Gonzales, Zhang, Papie{\.z}, Werys, Lukaschuk,
  Popescu, Burrage, Shanmuganathan, Ferreira, and
  Piechnik}]{gonzales2021moconet}
\bibinfo{author}{R.~A. Gonzales}, \bibinfo{author}{Q.~Zhang},
  \bibinfo{author}{B.~W. Papie{\.z}}, \bibinfo{author}{K.~Werys},
  \bibinfo{author}{E.~Lukaschuk}, \bibinfo{author}{I.~A. Popescu},
  \bibinfo{author}{M.~K. Burrage}, \bibinfo{author}{M.~Shanmuganathan},
  \bibinfo{author}{V.~M. Ferreira}, \bibinfo{author}{S.~K. Piechnik},
\newblock \bibinfo{title}{Moconet: robust motion correction of cardiovascular
  magnetic resonance t1 mapping using convolutional neural networks},
\newblock \bibinfo{journal}{Frontiers in Cardiovascular Medicine}
  \bibinfo{volume}{8} (\bibinfo{year}{2021}) \bibinfo{pages}{768245}.
\bibitem[{Terpstra et~al.(2020)Terpstra, Maspero, d’Agata, Stemkens, Intven,
  Lagendijk, Van~den Berg, and Tijssen}]{terpstra2020deep}
\bibinfo{author}{M.~L. Terpstra}, \bibinfo{author}{M.~Maspero},
  \bibinfo{author}{F.~d’Agata}, \bibinfo{author}{B.~Stemkens},
  \bibinfo{author}{M.~P. Intven}, \bibinfo{author}{J.~J. Lagendijk},
  \bibinfo{author}{C.~A. Van~den Berg}, \bibinfo{author}{R.~H. Tijssen},
\newblock \bibinfo{title}{Deep learning-based image reconstruction and motion
  estimation from undersampled radial k-space for real-time mri-guided
  radiotherapy},
\newblock \bibinfo{journal}{Physics in Medicine \& Biology}
  \bibinfo{volume}{65} (\bibinfo{year}{2020}) \bibinfo{pages}{155015}.
\bibitem[{Terpstra et~al.(2021)Terpstra, Maspero, Bruijnen, Verhoeff,
  Lagendijk, and van~den Berg}]{terpstra2021real}
\bibinfo{author}{M.~L. Terpstra}, \bibinfo{author}{M.~Maspero},
  \bibinfo{author}{T.~Bruijnen}, \bibinfo{author}{J.~J. Verhoeff},
  \bibinfo{author}{J.~J. Lagendijk}, \bibinfo{author}{C.~A. van~den Berg},
\newblock \bibinfo{title}{Real-time 3d motion estimation from undersampled mri
  using multi-resolution neural networks},
\newblock \bibinfo{journal}{Medical physics} \bibinfo{volume}{48}
  (\bibinfo{year}{2021}) \bibinfo{pages}{6597--6613}.
\bibitem[{Zhi et~al.(2023)Zhi, Wang, Xiao, Bai, Li, Tang, Liu, Li, Li, Ge
  et~al.}]{zhi2023coarse}
\bibinfo{author}{S.~Zhi}, \bibinfo{author}{Y.~Wang}, \bibinfo{author}{H.~Xiao},
  \bibinfo{author}{T.~Bai}, \bibinfo{author}{B.~Li}, \bibinfo{author}{Y.~Tang},
  \bibinfo{author}{C.~Liu}, \bibinfo{author}{W.~Li}, \bibinfo{author}{T.~Li},
  \bibinfo{author}{H.~Ge}, et~al.,
\newblock \bibinfo{title}{Coarse--super-resolution--fine network (cosf-net): A
  unified end-to-end neural network for 4d-mri with simultaneous motion
  estimation and super-resolution},
\newblock \bibinfo{journal}{IEEE Transactions on Medical Imaging}
  (\bibinfo{year}{2023}).
\bibitem[{Chen et~al.(2024)Chen, Ren, Li, and Li}]{chen2024motion}
\bibinfo{author}{Z.~Chen}, \bibinfo{author}{H.~Ren}, \bibinfo{author}{Q.~Li},
  \bibinfo{author}{X.~Li},
\newblock \bibinfo{title}{Motion correction and super-resolution for
  multi-slice cardiac magnetic resonance imaging via an end-to-end deep
  learning approach},
\newblock \bibinfo{journal}{Computerized Medical Imaging and Graphics}
  \bibinfo{volume}{115} (\bibinfo{year}{2024}) \bibinfo{pages}{102389}.
\bibitem[{Eichhorn et~al.(2023)Eichhorn, Hammernik, Spieker, Epp, Rueckert,
  Preibisch, and Schnabel}]{eichhorn2023physics}
\bibinfo{author}{H.~Eichhorn}, \bibinfo{author}{K.~Hammernik},
  \bibinfo{author}{V.~Spieker}, \bibinfo{author}{S.~M. Epp},
  \bibinfo{author}{D.~Rueckert}, \bibinfo{author}{C.~Preibisch},
  \bibinfo{author}{J.~A. Schnabel},
\newblock \bibinfo{title}{Physics-aware motion simulation for t2*-weighted
  brain mri},
\newblock in: \bibinfo{booktitle}{International Workshop on Simulation and
  Synthesis in Medical Imaging}, \bibinfo{organization}{Springer},
  \bibinfo{year}{2023}, pp. \bibinfo{pages}{42--52}.
\bibitem[{Yasaka et~al.(2024)Yasaka, Akai, Kato, Tajima, Yoshioka, Furuta,
  Kageyama, Toda, Akahane, Ohtomo et~al.}]{yasaka2024iterative}
\bibinfo{author}{K.~Yasaka}, \bibinfo{author}{H.~Akai},
  \bibinfo{author}{S.~Kato}, \bibinfo{author}{T.~Tajima},
  \bibinfo{author}{N.~Yoshioka}, \bibinfo{author}{T.~Furuta},
  \bibinfo{author}{H.~Kageyama}, \bibinfo{author}{Y.~Toda},
  \bibinfo{author}{M.~Akahane}, \bibinfo{author}{K.~Ohtomo}, et~al.,
\newblock \bibinfo{title}{Iterative motion correction technique with deep
  learning reconstruction for brain mri: A volunteer and patient study},
\newblock \bibinfo{journal}{Journal of Imaging Informatics in Medicine}
  (\bibinfo{year}{2024}) \bibinfo{pages}{1--7}.
\bibitem[{Cui et~al.(2023)Cui, Song, Wang, Wang, Wu, Xie, Li, and
  Yang}]{cui2023motion}
\bibinfo{author}{L.~Cui}, \bibinfo{author}{Y.~Song}, \bibinfo{author}{Y.~Wang},
  \bibinfo{author}{R.~Wang}, \bibinfo{author}{D.~Wu}, \bibinfo{author}{H.~Xie},
  \bibinfo{author}{J.~Li}, \bibinfo{author}{G.~Yang},
\newblock \bibinfo{title}{Motion artifact reduction for magnetic resonance
  imaging with deep learning and k-space analysis},
\newblock \bibinfo{journal}{PloS one} \bibinfo{volume}{18}
  (\bibinfo{year}{2023}) \bibinfo{pages}{e0278668}.
\bibitem[{Safari et~al.(2025)Safari, Wang, Eidex, Qiu, Chang, Yu, and
  Yang}]{safari2025physicsinformeddeeplearningmodel}
\bibinfo{author}{M.~Safari}, \bibinfo{author}{S.~Wang},
  \bibinfo{author}{Z.~Eidex}, \bibinfo{author}{R.~Qiu}, \bibinfo{author}{C.-W.
  Chang}, \bibinfo{author}{D.~S. Yu}, \bibinfo{author}{X.~Yang},
  \bibinfo{title}{A physics-informed deep learning model for mri brain motion
  correction}, \bibinfo{year}{2025}. \URLprefix
  \url{https://arxiv.org/abs/2502.09296}.
  \href{http://arxiv.org/abs/2502.09296}{{\tt arXiv:2502.09296}}.
\bibitem[{Wang and Salerno(2024)}]{wang2024deep}
\bibinfo{author}{J.~Wang}, \bibinfo{author}{M.~Salerno},
\newblock \bibinfo{title}{Deep learning-based rapid image reconstruction and
  motion correction for high-resolution cartesian first-pass myocardial
  perfusion imaging at 3t},
\newblock \bibinfo{journal}{Magnetic Resonance in Medicine}
  (\bibinfo{year}{2024}).
\bibitem[{Eichhorn et~al.(2024)Eichhorn, Spieker, Hammernik, Saks, Weiss,
  Preibisch, and Schnabel}]{eichhorn2024physics}
\bibinfo{author}{H.~Eichhorn}, \bibinfo{author}{V.~Spieker},
  \bibinfo{author}{K.~Hammernik}, \bibinfo{author}{E.~Saks},
  \bibinfo{author}{K.~Weiss}, \bibinfo{author}{C.~Preibisch},
  \bibinfo{author}{J.~A. Schnabel},
\newblock \bibinfo{title}{Physics-informed deep learning for motion-corrected
  reconstruction of quantitative brain mri},
\newblock \bibinfo{journal}{arXiv preprint arXiv:2403.08298}
  (\bibinfo{year}{2024}).
\bibitem[{Zhen et~al.(2016)Zhen, Yin, Bhaduri, Nachum, Laidley, and
  Li}]{zhen2016multi}
\bibinfo{author}{X.~Zhen}, \bibinfo{author}{Y.~Yin},
  \bibinfo{author}{M.~Bhaduri}, \bibinfo{author}{I.~B. Nachum},
  \bibinfo{author}{D.~Laidley}, \bibinfo{author}{S.~Li},
\newblock \bibinfo{title}{Multi-task shape regression for medical image
  segmentation},
\newblock in: \bibinfo{booktitle}{Medical Image Computing and Computer-Assisted
  Intervention-MICCAI 2016: 19th International Conference, Athens, Greece,
  October 17-21, 2016, Proceedings, Part III 19},
  \bibinfo{organization}{Springer}, \bibinfo{year}{2016}, pp.
  \bibinfo{pages}{210--218}.
\bibitem[{Pei et~al.(2020)Pei, Wang, Zhao, Zhong, Liao, Shen, and
  Li}]{pei2020anatomy}
\bibinfo{author}{Y.~Pei}, \bibinfo{author}{L.~Wang}, \bibinfo{author}{F.~Zhao},
  \bibinfo{author}{T.~Zhong}, \bibinfo{author}{L.~Liao},
  \bibinfo{author}{D.~Shen}, \bibinfo{author}{G.~Li},
\newblock \bibinfo{title}{Anatomy-guided convolutional neural network for
  motion correction in fetal brain mri},
\newblock in: \bibinfo{booktitle}{Machine Learning in Medical Imaging: 11th
  International Workshop, MLMI 2020, Held in Conjunction with MICCAI 2020,
  Lima, Peru, October 4, 2020, Proceedings 11},
  \bibinfo{organization}{Springer}, \bibinfo{year}{2020}, pp.
  \bibinfo{pages}{384--393}.
\bibitem[{Oksuz et~al.(2020)Oksuz, Clough, Ruijsink, Anton, Bustin, Cruz,
  Prieto, King, and Schnabel}]{oksuz2020deep}
\bibinfo{author}{I.~Oksuz}, \bibinfo{author}{J.~R. Clough},
  \bibinfo{author}{B.~Ruijsink}, \bibinfo{author}{E.~P. Anton},
  \bibinfo{author}{A.~Bustin}, \bibinfo{author}{G.~Cruz},
  \bibinfo{author}{C.~Prieto}, \bibinfo{author}{A.~P. King},
  \bibinfo{author}{J.~A. Schnabel},
\newblock \bibinfo{title}{Deep learning-based detection and correction of
  cardiac mr motion artefacts during reconstruction for high-quality
  segmentation},
\newblock \bibinfo{journal}{IEEE Transactions on Medical Imaging}
  \bibinfo{volume}{39} (\bibinfo{year}{2020}) \bibinfo{pages}{4001--4010}.
\bibitem[{Xu et~al.(2022)Xu, Kothapalli, Liu, Kahali, Gan, Yablonskiy, and
  Kamilov}]{xu2022learning}
\bibinfo{author}{X.~Xu}, \bibinfo{author}{S.~V. Kothapalli},
  \bibinfo{author}{J.~Liu}, \bibinfo{author}{S.~Kahali},
  \bibinfo{author}{W.~Gan}, \bibinfo{author}{D.~A. Yablonskiy},
  \bibinfo{author}{U.~S. Kamilov},
\newblock \bibinfo{title}{Learning-based motion artifact removal networks for
  quantitative r2s mapping},
\newblock \bibinfo{journal}{Magnetic resonance in medicine}
  \bibinfo{volume}{88} (\bibinfo{year}{2022}) \bibinfo{pages}{106--119}.
\bibitem[{Li et~al.(2022)Li, Wu, Qi, Si, Ding, and Chen}]{li2022motion}
\bibinfo{author}{Y.~Li}, \bibinfo{author}{C.~Wu}, \bibinfo{author}{H.~Qi},
  \bibinfo{author}{D.~Si}, \bibinfo{author}{H.~Ding},
  \bibinfo{author}{H.~Chen},
\newblock \bibinfo{title}{Motion correction for native myocardial t1 mapping
  using self-supervised deep learning registration with contrast separation},
\newblock \bibinfo{journal}{NMR in Biomedicine} \bibinfo{volume}{35}
  (\bibinfo{year}{2022}) \bibinfo{pages}{e4775}.
\bibitem[{Tran et~al.(2015)Tran, Bourdev, Fergus, Torresani, and
  Paluri}]{tran2015learning}
\bibinfo{author}{D.~Tran}, \bibinfo{author}{L.~Bourdev},
  \bibinfo{author}{R.~Fergus}, \bibinfo{author}{L.~Torresani},
  \bibinfo{author}{M.~Paluri},
\newblock \bibinfo{title}{Learning spatiotemporal features with 3d
  convolutional networks},
\newblock in: \bibinfo{booktitle}{Proceedings of the IEEE international
  conference on computer vision}, \bibinfo{year}{2015}, pp.
  \bibinfo{pages}{4489--4497}.
\bibitem[{Simonyan(2014)}]{simonyan2014very}
\bibinfo{author}{K.~Simonyan},
\newblock \bibinfo{title}{Very deep convolutional networks for large-scale
  image recognition},
\newblock \bibinfo{journal}{arXiv preprint arXiv:1409.1556}
  (\bibinfo{year}{2014}).
\bibitem[{Tan(2019)}]{tan2019efficientnet}
\bibinfo{author}{M.~Tan},
\newblock \bibinfo{title}{Efficientnet: Rethinking model scaling for
  convolutional neural networks},
\newblock \bibinfo{journal}{arXiv preprint arXiv:1905.11946}
  (\bibinfo{year}{2019}).
\bibitem[{Krizhevsky et~al.(2012)Krizhevsky, Sutskever, and
  Hinton}]{krizhevsky2012imagenet}
\bibinfo{author}{A.~Krizhevsky}, \bibinfo{author}{I.~Sutskever},
  \bibinfo{author}{G.~E. Hinton},
\newblock \bibinfo{title}{Imagenet classification with deep convolutional
  neural networks},
\newblock \bibinfo{journal}{Advances in neural information processing systems}
  \bibinfo{volume}{25} (\bibinfo{year}{2012}).
\bibitem[{He et~al.(2016)He, Zhang, Ren, and Sun}]{he2016deep}
\bibinfo{author}{K.~He}, \bibinfo{author}{X.~Zhang}, \bibinfo{author}{S.~Ren},
  \bibinfo{author}{J.~Sun},
\newblock \bibinfo{title}{Deep residual learning for image recognition},
\newblock in: \bibinfo{booktitle}{Proceedings of the IEEE conference on
  computer vision and pattern recognition}, \bibinfo{year}{2016}, pp.
  \bibinfo{pages}{770--778}.
\bibitem[{Vakli et~al.(2023)Vakli, Weiss, Szalma, Barsi, Gyuricza, Kemenczky,
  Somogyi, N{\'a}rai, G{\'a}l, Hermann et~al.}]{vakli2023automatic}
\bibinfo{author}{P.~Vakli}, \bibinfo{author}{B.~Weiss},
  \bibinfo{author}{J.~Szalma}, \bibinfo{author}{P.~Barsi},
  \bibinfo{author}{I.~Gyuricza}, \bibinfo{author}{P.~Kemenczky},
  \bibinfo{author}{E.~Somogyi}, \bibinfo{author}{{\'A}.~N{\'a}rai},
  \bibinfo{author}{V.~G{\'a}l}, \bibinfo{author}{P.~Hermann}, et~al.,
\newblock \bibinfo{title}{Automatic brain mri motion artifact detection based
  on end-to-end deep learning is similarly effective as traditional machine
  learning trained on image quality metrics},
\newblock \bibinfo{journal}{Medical Image Analysis} \bibinfo{volume}{88}
  (\bibinfo{year}{2023}) \bibinfo{pages}{102850}.
\bibitem[{Oksuz et~al.(2019)Oksuz, Ruijsink, Puyol-Ant{\'o}n, Clough, Cruz,
  Bustin, Prieto, Botnar, Rueckert, Schnabel et~al.}]{oksuz2019automatic}
\bibinfo{author}{I.~Oksuz}, \bibinfo{author}{B.~Ruijsink},
  \bibinfo{author}{E.~Puyol-Ant{\'o}n}, \bibinfo{author}{J.~R. Clough},
  \bibinfo{author}{G.~Cruz}, \bibinfo{author}{A.~Bustin},
  \bibinfo{author}{C.~Prieto}, \bibinfo{author}{R.~Botnar},
  \bibinfo{author}{D.~Rueckert}, \bibinfo{author}{J.~A. Schnabel}, et~al.,
\newblock \bibinfo{title}{Automatic cnn-based detection of cardiac mr motion
  artefacts using k-space data augmentation and curriculum learning},
\newblock \bibinfo{journal}{Medical image analysis} \bibinfo{volume}{55}
  (\bibinfo{year}{2019}) \bibinfo{pages}{136--147}.
\bibitem[{Usman et~al.(2013)Usman, Atkinson, Odille, Kolbitsch, Vaillant,
  Schaeffter, Batchelor, and Prieto}]{usman2013motion}
\bibinfo{author}{M.~Usman}, \bibinfo{author}{D.~Atkinson},
  \bibinfo{author}{F.~Odille}, \bibinfo{author}{C.~Kolbitsch},
  \bibinfo{author}{G.~Vaillant}, \bibinfo{author}{T.~Schaeffter},
  \bibinfo{author}{P.~G. Batchelor}, \bibinfo{author}{C.~Prieto},
\newblock \bibinfo{title}{Motion corrected compressed sensing for
  free-breathing dynamic cardiac mri},
\newblock \bibinfo{journal}{Magnetic resonance in medicine}
  \bibinfo{volume}{70} (\bibinfo{year}{2013}) \bibinfo{pages}{504--516}.
\bibitem[{Arshad et~al.(2024)Arshad, Potter, Chen, Liu, Chandrasekaran,
  Crabtree, Tong, Simonetti, Han, and Ahmad}]{arshad2024motion}
\bibinfo{author}{S.~M. Arshad}, \bibinfo{author}{L.~C. Potter},
  \bibinfo{author}{C.~Chen}, \bibinfo{author}{Y.~Liu},
  \bibinfo{author}{P.~Chandrasekaran}, \bibinfo{author}{C.~Crabtree},
  \bibinfo{author}{M.~S. Tong}, \bibinfo{author}{O.~P. Simonetti},
  \bibinfo{author}{Y.~Han}, \bibinfo{author}{R.~Ahmad},
\newblock \bibinfo{title}{Motion-robust free-running volumetric cardiovascular
  mri},
\newblock \bibinfo{journal}{Magnetic Resonance in Medicine}
  (\bibinfo{year}{2024}).
\bibitem[{Safari et~al.(2024)Safari, Eidex, Chang, Qiu, and
  Yang}]{safari2024fast}
\bibinfo{author}{M.~Safari}, \bibinfo{author}{Z.~Eidex}, \bibinfo{author}{C.-W.
  Chang}, \bibinfo{author}{R.~L. Qiu}, \bibinfo{author}{X.~Yang},
\newblock \bibinfo{title}{Fast mri reconstruction using deep learning-based
  compressed sensing: A systematic review},
\newblock \bibinfo{journal}{ArXiv}  (\bibinfo{year}{2024}).
\bibitem[{Gong et~al.(2021)Gong, Tong, Li, He, Zhang, and Zhong}]{gong2021deep}
\bibinfo{author}{T.~Gong}, \bibinfo{author}{Q.~Tong}, \bibinfo{author}{Z.~Li},
  \bibinfo{author}{H.~He}, \bibinfo{author}{H.~Zhang},
  \bibinfo{author}{J.~Zhong},
\newblock \bibinfo{title}{Deep learning-based method for reducing residual
  motion effects in diffusion parameter estimation},
\newblock \bibinfo{journal}{Magnetic Resonance in Medicine}
  \bibinfo{volume}{85} (\bibinfo{year}{2021}) \bibinfo{pages}{2278--2293}.
\bibitem[{Mongan et~al.(2020)Mongan, Moy, and Kahn~Jr}]{mongan2020checklist}
\bibinfo{author}{J.~Mongan}, \bibinfo{author}{L.~Moy}, \bibinfo{author}{C.~E.
  Kahn~Jr}, \bibinfo{title}{Checklist for artificial intelligence in medical
  imaging (claim): a guide for authors and reviewers}, \bibinfo{year}{2020}.
\bibitem[{Liberati et~al.(2009)Liberati, Altman, Tetzlaff, Mulrow, G{\o}tzsche,
  Ioannidis, Clarke, Devereaux, Kleijnen, and Moher}]{liberati2009prisma}
\bibinfo{author}{A.~Liberati}, \bibinfo{author}{D.~G. Altman},
  \bibinfo{author}{J.~Tetzlaff}, \bibinfo{author}{C.~Mulrow},
  \bibinfo{author}{P.~C. G{\o}tzsche}, \bibinfo{author}{J.~P. Ioannidis},
  \bibinfo{author}{M.~Clarke}, \bibinfo{author}{P.~J. Devereaux},
  \bibinfo{author}{J.~Kleijnen}, \bibinfo{author}{D.~Moher},
\newblock \bibinfo{title}{The prisma statement for reporting systematic reviews
  and meta-analyses of studies that evaluate health care interventions:
  explanation and elaboration},
\newblock \bibinfo{journal}{Annals of internal medicine} \bibinfo{volume}{151}
  (\bibinfo{year}{2009}) \bibinfo{pages}{W--65}.
\bibitem[{Virtanen et~al.(2020)Virtanen, Gommers, Oliphant, Haberland, Reddy,
  Cournapeau, Burovski, Peterson, Weckesser, Bright et~al.}]{virtanen2020scipy}
\bibinfo{author}{P.~Virtanen}, \bibinfo{author}{R.~Gommers},
  \bibinfo{author}{T.~E. Oliphant}, \bibinfo{author}{M.~Haberland},
  \bibinfo{author}{T.~Reddy}, \bibinfo{author}{D.~Cournapeau},
  \bibinfo{author}{E.~Burovski}, \bibinfo{author}{P.~Peterson},
  \bibinfo{author}{W.~Weckesser}, \bibinfo{author}{J.~Bright}, et~al.,
\newblock \bibinfo{title}{Scipy 1.0: fundamental algorithms for scientific
  computing in python},
\newblock \bibinfo{journal}{Nature methods} \bibinfo{volume}{17}
  (\bibinfo{year}{2020}) \bibinfo{pages}{261--272}.
\bibitem[{Zbontar et~al.(2018)Zbontar, Knoll, Sriram, Murrell, Huang, Muckley,
  Defazio, Stern, Johnson, Bruno et~al.}]{zbontar2018fastmri}
\bibinfo{author}{J.~Zbontar}, \bibinfo{author}{F.~Knoll},
  \bibinfo{author}{A.~Sriram}, \bibinfo{author}{T.~Murrell},
  \bibinfo{author}{Z.~Huang}, \bibinfo{author}{M.~J. Muckley},
  \bibinfo{author}{A.~Defazio}, \bibinfo{author}{R.~Stern},
  \bibinfo{author}{P.~Johnson}, \bibinfo{author}{M.~Bruno}, et~al.,
\newblock \bibinfo{title}{fastmri: An open dataset and benchmarks for
  accelerated mri},
\newblock \bibinfo{journal}{arXiv preprint arXiv:1811.08839}
  (\bibinfo{year}{2018}).
\bibitem[{N{\'a}rai et~al.(2022)N{\'a}rai, Hermann, Auer, Kemenczky, Szalma,
  Homolya, Somogyi, Vakli, Weiss, and Vidny{\'a}nszky}]{narai2022movement}
\bibinfo{author}{{\'A}.~N{\'a}rai}, \bibinfo{author}{P.~Hermann},
  \bibinfo{author}{T.~Auer}, \bibinfo{author}{P.~Kemenczky},
  \bibinfo{author}{J.~Szalma}, \bibinfo{author}{I.~Homolya},
  \bibinfo{author}{E.~Somogyi}, \bibinfo{author}{P.~Vakli},
  \bibinfo{author}{B.~Weiss}, \bibinfo{author}{Z.~Vidny{\'a}nszky},
\newblock \bibinfo{title}{Movement-related artefacts (mr-art) dataset of
  matched motion-corrupted and clean structural mri brain scans},
\newblock \bibinfo{journal}{Scientific data} \bibinfo{volume}{9}
  (\bibinfo{year}{2022}) \bibinfo{pages}{630}.
\bibitem[{Bycroft et~al.(2018)Bycroft, Freeman, Petkova, Band, Elliott, Sharp,
  Motyer, Vukcevic, Delaneau, O’Connell et~al.}]{bycroft2018uk}
\bibinfo{author}{C.~Bycroft}, \bibinfo{author}{C.~Freeman},
  \bibinfo{author}{D.~Petkova}, \bibinfo{author}{G.~Band},
  \bibinfo{author}{L.~T. Elliott}, \bibinfo{author}{K.~Sharp},
  \bibinfo{author}{A.~Motyer}, \bibinfo{author}{D.~Vukcevic},
  \bibinfo{author}{O.~Delaneau}, \bibinfo{author}{J.~O’Connell}, et~al.,
\newblock \bibinfo{title}{The uk biobank resource with deep phenotyping and
  genomic data},
\newblock \bibinfo{journal}{Nature} \bibinfo{volume}{562}
  (\bibinfo{year}{2018}) \bibinfo{pages}{203--209}.
\bibitem[{Simk{\'o} et~al.(2023)Simk{\'o}, Ruiter, L{\"o}fstedt, Garpebring,
  Nyholm, Bylund, and Jonsson}]{simko2023improving}
\bibinfo{author}{A.~Simk{\'o}}, \bibinfo{author}{S.~Ruiter},
  \bibinfo{author}{T.~L{\"o}fstedt}, \bibinfo{author}{A.~Garpebring},
  \bibinfo{author}{T.~Nyholm}, \bibinfo{author}{M.~Bylund},
  \bibinfo{author}{J.~Jonsson},
\newblock \bibinfo{title}{Improving mr image quality with a multi-task model,
  using convolutional losses},
\newblock \bibinfo{journal}{BMC Medical Imaging} \bibinfo{volume}{23}
  (\bibinfo{year}{2023}) \bibinfo{pages}{1--14}.
\bibitem[{Wang et~al.(2023)Wang, Yang, Yang, and Sun}]{wang2023dual}
\bibinfo{author}{J.~Wang}, \bibinfo{author}{Y.~Yang},
  \bibinfo{author}{Y.~Yang}, \bibinfo{author}{J.~Sun},
\newblock \bibinfo{title}{Dual domain motion artifacts correction for mr
  imaging under guidance of k-space uncertainty},
\newblock in: \bibinfo{booktitle}{International Conference on Medical Image
  Computing and Computer-Assisted Intervention},
  \bibinfo{organization}{Springer}, \bibinfo{year}{2023}, pp.
  \bibinfo{pages}{293--302}.
\bibitem[{Olsson et~al.(2024)Olsson, Millward, Starke, Gladytz, Klein, Fehr,
  Lai, Lippert, Niendorf, and Waiczies}]{olsson2024simulating}
\bibinfo{author}{H.~Olsson}, \bibinfo{author}{J.~M. Millward},
  \bibinfo{author}{L.~Starke}, \bibinfo{author}{T.~Gladytz},
  \bibinfo{author}{T.~Klein}, \bibinfo{author}{J.~Fehr}, \bibinfo{author}{W.-C.
  Lai}, \bibinfo{author}{C.~Lippert}, \bibinfo{author}{T.~Niendorf},
  \bibinfo{author}{S.~Waiczies},
\newblock \bibinfo{title}{Simulating rigid head motion artifacts on brain
  magnitude mri data--outcome on image quality and segmentation of the cerebral
  cortex},
\newblock \bibinfo{journal}{Plos one} \bibinfo{volume}{19}
  (\bibinfo{year}{2024}) \bibinfo{pages}{e0301132}.
\bibitem[{Belton et~al.(2024)Belton, Hagos, Lawlor, and
  Curran}]{belton2024towards}
\bibinfo{author}{N.~Belton}, \bibinfo{author}{M.~T. Hagos},
  \bibinfo{author}{A.~Lawlor}, \bibinfo{author}{K.~M. Curran},
\newblock \bibinfo{title}{Towards a unified approach for unsupervised brain mri
  motion artefact detection with few shot anomaly detection},
\newblock \bibinfo{journal}{Computerized Medical Imaging and Graphics}
  \bibinfo{volume}{115} (\bibinfo{year}{2024}) \bibinfo{pages}{102391}.
\bibitem[{Van~Essen et~al.(2012)Van~Essen, Ugurbil, Auerbach, Barch, Behrens,
  Bucholz, Chang, Chen, Corbetta, Curtiss et~al.}]{van2012human}
\bibinfo{author}{D.~C. Van~Essen}, \bibinfo{author}{K.~Ugurbil},
  \bibinfo{author}{E.~Auerbach}, \bibinfo{author}{D.~Barch},
  \bibinfo{author}{T.~E. Behrens}, \bibinfo{author}{R.~Bucholz},
  \bibinfo{author}{A.~Chang}, \bibinfo{author}{L.~Chen},
  \bibinfo{author}{M.~Corbetta}, \bibinfo{author}{S.~W. Curtiss}, et~al.,
\newblock \bibinfo{title}{The human connectome project: a data acquisition
  perspective},
\newblock \bibinfo{journal}{Neuroimage} \bibinfo{volume}{62}
  (\bibinfo{year}{2012}) \bibinfo{pages}{2222--2231}.
\bibitem[{Ettehadi et~al.(2022)Ettehadi, Kashyap, Zhang, Wang, Semanek, Desai,
  Guo, Posner, and Laine}]{ettehadi2022automated}
\bibinfo{author}{N.~Ettehadi}, \bibinfo{author}{P.~Kashyap},
  \bibinfo{author}{X.~Zhang}, \bibinfo{author}{Y.~Wang},
  \bibinfo{author}{D.~Semanek}, \bibinfo{author}{K.~Desai},
  \bibinfo{author}{J.~Guo}, \bibinfo{author}{J.~Posner}, \bibinfo{author}{A.~F.
  Laine},
\newblock \bibinfo{title}{Automated multiclass artifact detection in diffusion
  mri volumes via 3d residual squeeze-and-excitation convolutional neural
  networks},
\newblock \bibinfo{journal}{Frontiers in Human Neuroscience}
  \bibinfo{volume}{16} (\bibinfo{year}{2022}) \bibinfo{pages}{877326}.
\bibitem[{Sudlow et~al.(2015)Sudlow, Gallacher, Allen, Beral, Burton, Danesh,
  Downey, Elliott, Green, Landray et~al.}]{sudlow2015uk}
\bibinfo{author}{C.~Sudlow}, \bibinfo{author}{J.~Gallacher},
  \bibinfo{author}{N.~Allen}, \bibinfo{author}{V.~Beral},
  \bibinfo{author}{P.~Burton}, \bibinfo{author}{J.~Danesh},
  \bibinfo{author}{P.~Downey}, \bibinfo{author}{P.~Elliott},
  \bibinfo{author}{J.~Green}, \bibinfo{author}{M.~Landray}, et~al.,
\newblock \bibinfo{title}{Uk biobank: an open access resource for identifying
  the causes of a wide range of complex diseases of middle and old age},
\newblock \bibinfo{journal}{PLoS medicine} \bibinfo{volume}{12}
  (\bibinfo{year}{2015}) \bibinfo{pages}{e1001779}.
\bibitem[{Mohebbian et~al.(2021)Mohebbian, Walia, Habibullah, Stapleton, and
  Wahid}]{mohebbian2021classifying}
\bibinfo{author}{M.~Mohebbian}, \bibinfo{author}{E.~Walia},
  \bibinfo{author}{M.~Habibullah}, \bibinfo{author}{S.~Stapleton},
  \bibinfo{author}{K.~A. Wahid},
\newblock \bibinfo{title}{Classifying mri motion severity using a stacked
  ensemble approach},
\newblock \bibinfo{journal}{Magnetic Resonance Imaging} \bibinfo{volume}{75}
  (\bibinfo{year}{2021}) \bibinfo{pages}{107--115}.
\bibitem[{Chatterjee et~al.(2020)Chatterjee, Sciarra, D{\"u}nnwald,
  Oeltze-Jafra, N{\"u}rnberger, and Speck}]{chatterjee2020retrospective}
\bibinfo{author}{S.~Chatterjee}, \bibinfo{author}{A.~Sciarra},
  \bibinfo{author}{M.~D{\"u}nnwald}, \bibinfo{author}{S.~Oeltze-Jafra},
  \bibinfo{author}{A.~N{\"u}rnberger}, \bibinfo{author}{O.~Speck},
\newblock \bibinfo{title}{Retrospective motion correction of mr images using
  prior-assisted deep learning},
\newblock \bibinfo{journal}{arXiv preprint arXiv:2011.14134}
  (\bibinfo{year}{2020}).
\bibitem[{Di~Martino et~al.(2014)Di~Martino, Yan, Li, Denio, Castellanos,
  Alaerts, Anderson, Assaf, Bookheimer, Dapretto et~al.}]{di2014autism}
\bibinfo{author}{A.~Di~Martino}, \bibinfo{author}{C.-G. Yan},
  \bibinfo{author}{Q.~Li}, \bibinfo{author}{E.~Denio}, \bibinfo{author}{F.~X.
  Castellanos}, \bibinfo{author}{K.~Alaerts}, \bibinfo{author}{J.~S. Anderson},
  \bibinfo{author}{M.~Assaf}, \bibinfo{author}{S.~Y. Bookheimer},
  \bibinfo{author}{M.~Dapretto}, et~al.,
\newblock \bibinfo{title}{The autism brain imaging data exchange: towards a
  large-scale evaluation of the intrinsic brain architecture in autism},
\newblock \bibinfo{journal}{Molecular psychiatry} \bibinfo{volume}{19}
  (\bibinfo{year}{2014}) \bibinfo{pages}{659--667}.
\bibitem[{Di~Martino et~al.(2017)Di~Martino, O’connor, Chen, Alaerts,
  Anderson, Assaf, Balsters, Baxter, Beggiato, Bernaerts
  et~al.}]{di2017enhancing}
\bibinfo{author}{A.~Di~Martino}, \bibinfo{author}{D.~O’connor},
  \bibinfo{author}{B.~Chen}, \bibinfo{author}{K.~Alaerts},
  \bibinfo{author}{J.~S. Anderson}, \bibinfo{author}{M.~Assaf},
  \bibinfo{author}{J.~H. Balsters}, \bibinfo{author}{L.~Baxter},
  \bibinfo{author}{A.~Beggiato}, \bibinfo{author}{S.~Bernaerts}, et~al.,
\newblock \bibinfo{title}{Enhancing studies of the connectome in autism using
  the autism brain imaging data exchange ii},
\newblock \bibinfo{journal}{Scientific data} \bibinfo{volume}{4}
  (\bibinfo{year}{2017}) \bibinfo{pages}{1--15}.
\bibitem[{Li and Zhao(2022)}]{li2022addressing}
\bibinfo{author}{S.~Li}, \bibinfo{author}{Y.~Zhao},
\newblock \bibinfo{title}{Addressing motion blurs in brain mri scans using
  conditional adversarial networks and simulated curvilinear motions},
\newblock \bibinfo{journal}{Journal of Imaging} \bibinfo{volume}{8}
  (\bibinfo{year}{2022}) \bibinfo{pages}{84}.
\bibitem[{Zhao et~al.(2021)Zhao, Ossowski, Wang, Li, Devinsky, Martin, and
  Pardoe}]{zhao2021localized}
\bibinfo{author}{Y.~Zhao}, \bibinfo{author}{J.~Ossowski},
  \bibinfo{author}{X.~Wang}, \bibinfo{author}{S.~Li},
  \bibinfo{author}{O.~Devinsky}, \bibinfo{author}{S.~P. Martin},
  \bibinfo{author}{H.~R. Pardoe},
\newblock \bibinfo{title}{Localized motion artifact reduction on brain mri
  using deep learning with effective data augmentation techniques},
\newblock in: \bibinfo{booktitle}{2021 International Joint Conference on Neural
  Networks (IJCNN)}, \bibinfo{organization}{IEEE}, \bibinfo{year}{2021}, pp.
  \bibinfo{pages}{1--9}.
\bibitem[{Marcus et~al.(2007)Marcus, Wang, Parker, Csernansky, Morris, and
  Buckner}]{marcus2007open}
\bibinfo{author}{D.~S. Marcus}, \bibinfo{author}{T.~H. Wang},
  \bibinfo{author}{J.~Parker}, \bibinfo{author}{J.~G. Csernansky},
  \bibinfo{author}{J.~C. Morris}, \bibinfo{author}{R.~L. Buckner},
\newblock \bibinfo{title}{Open access series of imaging studies (oasis):
  cross-sectional mri data in young, middle aged, nondemented, and demented
  older adults},
\newblock \bibinfo{journal}{Journal of cognitive neuroscience}
  \bibinfo{volume}{19} (\bibinfo{year}{2007}) \bibinfo{pages}{1498--1507}.
\bibitem[{Bernard et~al.(2018)Bernard, Lalande, Zotti, Cervenansky, Yang, Heng,
  Cetin, Lekadir, Camara, Ballester et~al.}]{bernard2018deep}
\bibinfo{author}{O.~Bernard}, \bibinfo{author}{A.~Lalande},
  \bibinfo{author}{C.~Zotti}, \bibinfo{author}{F.~Cervenansky},
  \bibinfo{author}{X.~Yang}, \bibinfo{author}{P.-A. Heng},
  \bibinfo{author}{I.~Cetin}, \bibinfo{author}{K.~Lekadir},
  \bibinfo{author}{O.~Camara}, \bibinfo{author}{M.~A.~G. Ballester}, et~al.,
\newblock \bibinfo{title}{Deep learning techniques for automatic mri cardiac
  multi-structures segmentation and diagnosis: is the problem solved?},
\newblock \bibinfo{journal}{IEEE transactions on medical imaging}
  \bibinfo{volume}{37} (\bibinfo{year}{2018}) \bibinfo{pages}{2514--2525}.
\bibitem[{Petersen et~al.(2010)Petersen, Aisen, Beckett, Donohue, Gamst,
  Harvey, Jack~Jr, Jagust, Shaw, Toga et~al.}]{petersen2010alzheimer}
\bibinfo{author}{R.~C. Petersen}, \bibinfo{author}{P.~S. Aisen},
  \bibinfo{author}{L.~A. Beckett}, \bibinfo{author}{M.~C. Donohue},
  \bibinfo{author}{A.~C. Gamst}, \bibinfo{author}{D.~J. Harvey},
  \bibinfo{author}{C.~Jack~Jr}, \bibinfo{author}{W.~J. Jagust},
  \bibinfo{author}{L.~M. Shaw}, \bibinfo{author}{A.~W. Toga}, et~al.,
\newblock \bibinfo{title}{Alzheimer's disease neuroimaging initiative (adni)
  clinical characterization},
\newblock \bibinfo{journal}{Neurology} \bibinfo{volume}{74}
  (\bibinfo{year}{2010}) \bibinfo{pages}{201--209}.
\bibitem[{Loizillon et~al.(2024)Loizillon, Bottani, Maire, Str{\"o}er, Dormont,
  Colliot, Burgos, Initiative, Group et~al.}]{loizillon2024automatic}
\bibinfo{author}{S.~Loizillon}, \bibinfo{author}{S.~Bottani},
  \bibinfo{author}{A.~Maire}, \bibinfo{author}{S.~Str{\"o}er},
  \bibinfo{author}{D.~Dormont}, \bibinfo{author}{O.~Colliot},
  \bibinfo{author}{N.~Burgos}, \bibinfo{author}{A.~D.~N. Initiative},
  \bibinfo{author}{A.~S. Group}, et~al.,
\newblock \bibinfo{title}{Automatic motion artefact detection in brain
  t1-weighted magnetic resonance images from a clinical data warehouse using
  synthetic data},
\newblock \bibinfo{journal}{Medical Image Analysis} \bibinfo{volume}{93}
  (\bibinfo{year}{2024}) \bibinfo{pages}{103073}.
\bibitem[{Shusharina and Bortfeld(2021)}]{shusharina2021glioma}
\bibinfo{author}{N.~Shusharina}, \bibinfo{author}{T.~Bortfeld},
\newblock \bibinfo{title}{Glioma image segmentation for radiotherapy: Rt
  targets, barriers to cancer spread, and organs at risk [data set]},
\newblock \bibinfo{journal}{The Cancer Imaging Archive}
  (\bibinfo{year}{2021}).
\bibitem[{Commowick et~al.(2018)Commowick, Istace, Kain, Laurent, Leray, Simon,
  Pop, Girard, Ameli, Ferr{\'e} et~al.}]{commowick2018objective}
\bibinfo{author}{O.~Commowick}, \bibinfo{author}{A.~Istace},
  \bibinfo{author}{M.~Kain}, \bibinfo{author}{B.~Laurent},
  \bibinfo{author}{F.~Leray}, \bibinfo{author}{M.~Simon},
  \bibinfo{author}{S.~C. Pop}, \bibinfo{author}{P.~Girard},
  \bibinfo{author}{R.~Ameli}, \bibinfo{author}{J.-C. Ferr{\'e}}, et~al.,
\newblock \bibinfo{title}{Objective evaluation of multiple sclerosis lesion
  segmentation using a data management and processing infrastructure},
\newblock \bibinfo{journal}{Scientific reports} \bibinfo{volume}{8}
  (\bibinfo{year}{2018}) \bibinfo{pages}{13650}.
\bibitem[{Mercier et~al.(2012)Mercier, Del~Maestro, Petrecca, Araujo, Haegelen,
  and Collins}]{mercier2012online}
\bibinfo{author}{L.~Mercier}, \bibinfo{author}{R.~F. Del~Maestro},
  \bibinfo{author}{K.~Petrecca}, \bibinfo{author}{D.~Araujo},
  \bibinfo{author}{C.~Haegelen}, \bibinfo{author}{D.~L. Collins},
\newblock \bibinfo{title}{Online database of clinical mr and ultrasound images
  of brain tumors},
\newblock \bibinfo{journal}{Medical physics} \bibinfo{volume}{39}
  (\bibinfo{year}{2012}) \bibinfo{pages}{3253--3261}.
\bibitem[{Cocosco et~al.(1997)Cocosco, Kollokian, Kwan, and
  Evans}]{cocosco5online}
\bibinfo{author}{C.~Cocosco}, \bibinfo{author}{V.~Kollokian},
  \bibinfo{author}{R.~Kwan}, \bibinfo{author}{A.~Evans},
\newblock \bibinfo{title}{Online interface to a 3d mri simulated brain
  database},
\newblock \bibinfo{journal}{NeuroImage} \bibinfo{volume}{5}
  (\bibinfo{year}{1997}) \bibinfo{pages}{S425}.
\bibitem[{Forstmann et~al.(2014)Forstmann, Keuken, Schafer, Bazin, Alkemade,
  and Turner}]{forstmann2014multi}
\bibinfo{author}{B.~Forstmann}, \bibinfo{author}{M.~Keuken},
  \bibinfo{author}{A.~Schafer}, \bibinfo{author}{P.~Bazin},
  \bibinfo{author}{A.~Alkemade}, \bibinfo{author}{R.~Turner},
  \bibinfo{title}{Multi-modal ultra-high resolution structural 7-tesla mri data
  repository. sci. data 1, 140050}, \bibinfo{year}{2014}.
\bibitem[{Caspers et~al.(2014)Caspers, Moebus, Lux, Pundt, Sch{\"u}tz,
  M{\"u}hleisen, Gras, Eickhoff, Romanzetti, St{\"o}cker
  et~al.}]{caspers2014studying}
\bibinfo{author}{S.~Caspers}, \bibinfo{author}{S.~Moebus},
  \bibinfo{author}{S.~Lux}, \bibinfo{author}{N.~Pundt},
  \bibinfo{author}{H.~Sch{\"u}tz}, \bibinfo{author}{T.~W. M{\"u}hleisen},
  \bibinfo{author}{V.~Gras}, \bibinfo{author}{S.~B. Eickhoff},
  \bibinfo{author}{S.~Romanzetti}, \bibinfo{author}{T.~St{\"o}cker}, et~al.,
\newblock \bibinfo{title}{Studying variability in human brain aging in a
  population-based german cohort—rationale and design of 1000brains},
\newblock \bibinfo{journal}{Frontiers in aging neuroscience}
  \bibinfo{volume}{6} (\bibinfo{year}{2014}) \bibinfo{pages}{149}.
\bibitem[{Roecher et~al.(2024)Roecher, M{\"o}sch, Zweerings, Thiele, Caspers,
  Gaebler, Eisner, Sarkheil, and Mathiak}]{roecher2024motion}
\bibinfo{author}{E.~Roecher}, \bibinfo{author}{L.~M{\"o}sch},
  \bibinfo{author}{J.~Zweerings}, \bibinfo{author}{F.~O. Thiele},
  \bibinfo{author}{S.~Caspers}, \bibinfo{author}{A.~J. Gaebler},
  \bibinfo{author}{P.~Eisner}, \bibinfo{author}{P.~Sarkheil},
  \bibinfo{author}{K.~Mathiak},
\newblock \bibinfo{title}{Motion artifact detection for t1-weighted brain mr
  images using convolutional neural networks},
\newblock \bibinfo{journal}{International journal of neural systems}
  (\bibinfo{year}{2024}) \bibinfo{pages}{2450052}.
\bibitem[{Abadi et~al.(2015)Abadi, Agarwal, Barham, Brevdo, Chen, Citro,
  Corrado, Davis, Dean, Devin, Ghemawat, Goodfellow, Harp, Irving, Isard, Jia,
  Jozefowicz, Kaiser, Kudlur, Levenberg, Mané, Monga, Moore, Murray, Olah,
  Schuster, Shlens, Steiner, Sutskever, Talwar, Tucker, Vanhoucke, Vasudevan,
  Viégas, Vinyals, Warden, Wattenberg, Wicke, Yu, and
  Zheng}]{tensorflow2015whitepaper}
\bibinfo{author}{M.~Abadi}, \bibinfo{author}{A.~Agarwal},
  \bibinfo{author}{P.~Barham}, \bibinfo{author}{E.~Brevdo},
  \bibinfo{author}{Z.~Chen}, \bibinfo{author}{C.~Citro}, \bibinfo{author}{G.~S.
  Corrado}, \bibinfo{author}{A.~Davis}, \bibinfo{author}{J.~Dean},
  \bibinfo{author}{M.~Devin}, \bibinfo{author}{S.~Ghemawat},
  \bibinfo{author}{I.~Goodfellow}, \bibinfo{author}{A.~Harp},
  \bibinfo{author}{G.~Irving}, \bibinfo{author}{M.~Isard},
  \bibinfo{author}{Y.~Jia}, \bibinfo{author}{R.~Jozefowicz},
  \bibinfo{author}{L.~Kaiser}, \bibinfo{author}{M.~Kudlur},
  \bibinfo{author}{J.~Levenberg}, \bibinfo{author}{D.~Mané},
  \bibinfo{author}{R.~Monga}, \bibinfo{author}{S.~Moore},
  \bibinfo{author}{D.~Murray}, \bibinfo{author}{C.~Olah},
  \bibinfo{author}{M.~Schuster}, \bibinfo{author}{J.~Shlens},
  \bibinfo{author}{B.~Steiner}, \bibinfo{author}{I.~Sutskever},
  \bibinfo{author}{K.~Talwar}, \bibinfo{author}{P.~Tucker},
  \bibinfo{author}{V.~Vanhoucke}, \bibinfo{author}{V.~Vasudevan},
  \bibinfo{author}{F.~Viégas}, \bibinfo{author}{O.~Vinyals},
  \bibinfo{author}{P.~Warden}, \bibinfo{author}{M.~Wattenberg},
  \bibinfo{author}{M.~Wicke}, \bibinfo{author}{Y.~Yu},
  \bibinfo{author}{X.~Zheng}, \bibinfo{title}{Tensorflow: Large-scale machine
  learning on heterogeneous systems}, \bibinfo{year}{2015}. \URLprefix
  \url{https://www.tensorflow.org/}.
\bibitem[{Paszke et~al.(2019)Paszke, Gross, Massa, Lerer, Bradbury, Chanan,
  Killeen, Lin, Gimelshein, Antiga, Desmaison, Kopf, Yang, DeVito, Raison,
  Tejani, Chilamkurthy, Steiner, Fang, Bai, and Chintala}]{paszke2019pytorch}
\bibinfo{author}{A.~Paszke}, \bibinfo{author}{S.~Gross},
  \bibinfo{author}{F.~Massa}, \bibinfo{author}{A.~Lerer},
  \bibinfo{author}{J.~Bradbury}, \bibinfo{author}{G.~Chanan},
  \bibinfo{author}{T.~Killeen}, \bibinfo{author}{Z.~Lin},
  \bibinfo{author}{N.~Gimelshein}, \bibinfo{author}{L.~Antiga},
  \bibinfo{author}{A.~Desmaison}, \bibinfo{author}{A.~Kopf},
  \bibinfo{author}{E.~Yang}, \bibinfo{author}{Z.~DeVito},
  \bibinfo{author}{M.~Raison}, \bibinfo{author}{A.~Tejani},
  \bibinfo{author}{S.~Chilamkurthy}, \bibinfo{author}{B.~Steiner},
  \bibinfo{author}{L.~Fang}, \bibinfo{author}{J.~Bai},
  \bibinfo{author}{S.~Chintala},
\newblock \bibinfo{title}{Pytorch: An imperative style, high-performance deep
  learning library},
\newblock in: \bibinfo{editor}{H.~Wallach}, \bibinfo{editor}{H.~Larochelle},
  \bibinfo{editor}{A.~Beygelzimer}, \bibinfo{editor}{F.~d'Alché Buc},
  \bibinfo{editor}{E.~Fox}, \bibinfo{editor}{R.~Garnett} (Eds.),
  \bibinfo{booktitle}{Advances in Neural Information Processing Systems 32},
  \bibinfo{year}{2019}, pp. \bibinfo{pages}{8024--8035}. \URLprefix
  \url{https://pytorch.org/}.
\bibitem[{Gao et~al.(2023)Gao, Luo, Ding, Yang, and Sun}]{gao2023lightweight}
\bibinfo{author}{R.~Gao}, \bibinfo{author}{G.~Luo}, \bibinfo{author}{R.~Ding},
  \bibinfo{author}{B.~Yang}, \bibinfo{author}{H.~Sun},
\newblock \bibinfo{title}{A lightweight deep learning framework for automatic
  mri data sorting and artifacts detection},
\newblock \bibinfo{journal}{Journal of Medical Systems} \bibinfo{volume}{47}
  (\bibinfo{year}{2023}) \bibinfo{pages}{124}.
\bibitem[{Weaver et~al.(2023)Weaver, DiPiero, Rodrigues, Cordash, Davidson,
  Planalp, and Dean~III}]{weaver2023automated}
\bibinfo{author}{J.~M. Weaver}, \bibinfo{author}{M.~DiPiero},
  \bibinfo{author}{P.~G. Rodrigues}, \bibinfo{author}{H.~Cordash},
  \bibinfo{author}{R.~J. Davidson}, \bibinfo{author}{E.~M. Planalp},
  \bibinfo{author}{D.~C. Dean~III},
\newblock \bibinfo{title}{Automated motion artifact detection in early
  pediatric diffusion mri using a convolutional neural network},
\newblock \bibinfo{journal}{Imaging Neuroscience} \bibinfo{volume}{1}
  (\bibinfo{year}{2023}) \bibinfo{pages}{1--16}.
\bibitem[{Wang et~al.(2004)Wang, Bovik, Sheikh, and Simoncelli}]{wang2004image}
\bibinfo{author}{Z.~Wang}, \bibinfo{author}{A.~C. Bovik},
  \bibinfo{author}{H.~R. Sheikh}, \bibinfo{author}{E.~P. Simoncelli},
\newblock \bibinfo{title}{Image quality assessment: from error visibility to
  structural similarity},
\newblock \bibinfo{journal}{IEEE transactions on image processing}
  \bibinfo{volume}{13} (\bibinfo{year}{2004}) \bibinfo{pages}{600--612}.
\bibitem[{Kingma(2014)}]{kingma2014adam}
\bibinfo{author}{D.~P. Kingma},
\newblock \bibinfo{title}{Adam: A method for stochastic optimization},
\newblock \bibinfo{journal}{arXiv preprint arXiv:1412.6980}
  (\bibinfo{year}{2014}).
\bibitem[{Kromrey et~al.(2020)Kromrey, Tamada, Johno, Funayama, Nagata,
  Ichikawa, K{\"u}hn, Onishi, and Motosugi}]{kromrey2020reduction}
\bibinfo{author}{M.-L. Kromrey}, \bibinfo{author}{D.~Tamada},
  \bibinfo{author}{H.~Johno}, \bibinfo{author}{S.~Funayama},
  \bibinfo{author}{N.~Nagata}, \bibinfo{author}{S.~Ichikawa},
  \bibinfo{author}{J.-P. K{\"u}hn}, \bibinfo{author}{H.~Onishi},
  \bibinfo{author}{U.~Motosugi},
\newblock \bibinfo{title}{Reduction of respiratory motion artifacts in
  gadoxetate-enhanced mr with a deep learning--based filter using convolutional
  neural network},
\newblock \bibinfo{journal}{European radiology} \bibinfo{volume}{30}
  (\bibinfo{year}{2020}) \bibinfo{pages}{5923--5932}.
\bibitem[{Zhang et~al.(2024{\natexlab{a}})Zhang, Huang, Jin, and
  Lu}]{zhang2024vision}
\bibinfo{author}{J.~Zhang}, \bibinfo{author}{J.~Huang},
  \bibinfo{author}{S.~Jin}, \bibinfo{author}{S.~Lu},
\newblock \bibinfo{title}{Vision-language models for vision tasks: A survey},
\newblock \bibinfo{journal}{IEEE Transactions on Pattern Analysis and Machine
  Intelligence}  (\bibinfo{year}{2024}{\natexlab{a}}).
\bibitem[{Zhang et~al.(2024{\natexlab{b}})Zhang, Wang, Herskovits, Melhem,
  Chang, Wang, and Ernst}]{10647626}
\bibinfo{author}{L.~Zhang}, \bibinfo{author}{X.~Wang}, \bibinfo{author}{E.~H.
  Herskovits}, \bibinfo{author}{E.~R. Melhem}, \bibinfo{author}{L.~Chang},
  \bibinfo{author}{Z.~Wang}, \bibinfo{author}{T.~Ernst},
\newblock \bibinfo{title}{Reducing motion artifacts in brain mri using vision
  transformers and self-supervised learning},
\newblock in: \bibinfo{booktitle}{2024 IEEE International Conference on Image
  Processing (ICIP)}, \bibinfo{year}{2024}{\natexlab{b}}, pp.
  \bibinfo{pages}{3165--3171}. \DOIprefix\doi{10.1109/ICIP51287.2024.10647626}.

\end{thebibliography}

\end{document}